\documentclass{wiley-article}

\usepackage{natbib}

\usepackage{graphicx}
\usepackage{colortbl}
\usepackage{comment}

\usepackage{setspace}
\usepackage{verbatim} 

\usepackage{latexsym}
\usepackage{amsmath}
\usepackage{amsfonts}
\usepackage{amssymb}
\usepackage{amsbsy}
\usepackage{wrapfig}

\usepackage{chemformula}
\usepackage{siunitx}

\usepackage[ruled,noend]{algorithm2e}
\usepackage{multirow}
\allowdisplaybreaks

\papertype{Research Article}

\title{Interpretable Biomanufacturing Process Risk and Sensitivity Analyses for Quality-by-Design and Stability Control}

\author[1]{Wei Xie}
\author[1]{Bo Wang}
\author[2]{Cheng Li}
\author[3]{Dongming Xie}
\author[4]{Jared Auclair}

\affil[1]{Department of Mechanical and Industrial Engineering, Northeastern University, Boston, MA 02115, USA}
\affil[2]{Department of Statistics and Applied Probability, National University of Singapore, Singapore}
\affil[3]{Department of Chemical Engineering, University of Massachusetts Lowell, Lowell, MA 01854, USA}
\affil[4]{Department of Chemistry and Chemical Biology, Northeastern University, Boston, MA 02115, USA}

\corraddress{Wei Xie, Department of Mechanical and Industrial Engineering, Northeastern University, Boston, MA 02115, USA}
\corremail{w.xie@northeastern.edu}

\runningauthor{Bioprocess Risk and Sensitivity Analyses}

\begin{document}

\begin{frontmatter}
\maketitle

\begin{abstract}
While biomanufacturing plays a significant role in supporting the economy and ensuring public health, it faces critical challenges, including complexity, high variability, lengthy lead time, and very limited process data, especially for personalized new cell and gene biotherapeutics. Driven by these challenges, we propose an interpretable semantic bioprocess probabilistic knowledge graph and develop a game theory based risk and sensitivity analyses for production process to facilitate quality-by-design and stability control. Specifically, by exploring the causal relationships and interactions of critical process parameters and quality attributes (CPPs/CQAs), we create a Bayesian network based probabilistic knowledge graph characterizing the complex causal interdependencies of all factors. Then, we introduce a Shapley value based sensitivity analysis, which can correctly quantify the variation contribution from each input factor on the outputs (i.e., productivity, product quality). Since the bioprocess model coefficients are learned from limited process observations, we derive the Bayesian posterior distribution to quantify model uncertainty and further develop the Shapley value based sensitivity analysis to evaluate the impact of estimation uncertainty from each set of model coefficients. Therefore, the proposed bioprocess risk and sensitivity analyses can identify the bottlenecks, guide the reliable process specifications and the most informative data collection, and improve production stability.

\end{abstract}

\end{frontmatter}

\section{Introduction}
\label{sec:Intro}

{Biomanufacturing is growing rapidly and playing an increasingly significant role in supporting the economy and ensuring public health.} For example, the biopharmaceutical industry generated more than \$300 billion in revenue in 2019 and more than 40\% of the drug products in the development pipeline were biopharmaceuticals \citep{rader2019single}. 
However, drug shortages have occurred at unprecedented rates over the past decade. The current systems are unable to rapidly produce new drugs to meet urgent needs in the presence of a major public health emergency. 
The COVID-19 pandemic is having a profound impact globally and caused over 111 millions confirmed cases by February, 2021. Even COVID-19 
vaccines are discovered,
developing the production process 
and manufacturing the billions of doses needed to immunize the world's population will be extremely time-consuming using existing technologies, thus lengthening the time period of human and economic distress. 
\textit{It is critically important to speed up the bioprocess development and ensure product quality consistency.
}

\textit{However, biomanufacturing faces several critical challenges, including high complexity and variability, and lengthy lead time \citep{Kaminsky_Wang_2015}.}
Biomanufacturing is based on living cells whose biological processes are very complex and have highly variable outputs. 
The productivity and product critical quality attributes (CQAs) 
are determined by the interactions of hundreds of critical process parameters (CPPs), including raw materials, media compositions, feeding strategy, and process operational conditions, such as pH and dissolved oxygen in the bioreactor. As new biotherapeutics (e.g., cell and gene therapies) become more and more ``personalized", the production, regulation procedure, and analytical testing time required by biopharmaceuticals of complex molecular structure is lengthy, and the historical observations are relatively limited in particular for drugs in early stages of production process development. 

\textit{Therefore, it is crucial to integrate all sources of data and mechanism information, provide the risk- and science-based understanding of the complex bioprocess CPPs/CQAs causal interdependencies, and identify and control the key factors contributing the most to the output variation.} This study can accelerate the development of productive and reliable biomanufacturing, facilitate building the quality into the production process or quality-by-design (QbD), support real-time monitoring and release, and reduce the time to market. 

Various \textit{Process Analytical Technologies (PAT)} and methodologies have been proposed to improve the bioprocess understanding and guide the process development, decision making, and risk control; see the review in \cite{steinwandter2019data}. 
Most PATs are based on multivariate data analysis; see Section~\ref{sec:background}.
Ordinary or partial 
differential equations (ODEs/PDEs) based mechanistic models are developed for simulating individual biomanufacturing unit operations; see for example \cite{kyriakopoulos2018kinetic}. 
On the other hand, 
various operations research/management (OR/OM) methods are also proposed for biomanufacturing system analytics and decision-making; see the review \citep{Kaminsky_Wang_2015}.  
Overall, existing methodologies
have the \textit{key limitations}: (1) the multivariate statistics based PAT and OR/OM approaches focus on developing general methodologies without incorporating the bioprocess causal relationship and structural mechanism information, which limits their performance, interpretability, and adoption, especially with limited data; and (2) the mechanistic models are usually deterministic and focus on individual unit operations without providing an reliable integrated bioprocess learning and risk management framework.

Driven by the critical challenges in the biomanufacturing industry, in this paper, we propose a bioprocess semantic probabilistic knowledge graph, characterizing the risk- and science-based understanding of integrated production process, which can integrate all sources of heterogeneous data and leverage the information from existing mechanism models and historical data. Then, we introduce comprehensive and rigorous bioprocess risk and sensitivity analyses, accounting for model risk, which can guide the process specifications and most informative data collection to facilitate the learning and improve the production reliability and stability (e.g., product quality consistency). 

The \textit{key contributions} of this paper are three fold. First, by exploring the causal relationships and interactions  of many factors within and between operation units (i.e., CPPs/CQAs), such as raw materials, production process parameters, and product quality, we consider a bioprocess ontology based data integration and develop a Bayesian network (BN) based bioprocess probabilistic knowledge graph, 
characterizing the process inherent stochastic uncertainty and causal interdependencies of all input and output factors.
Second, building on the process knowledge graph, we introduce a game theory -- Shapley value (SV) -- based sensitivity analysis (SA), considering the complex bioprocess interdependencies, which can correctly quantify the contribution and criticality of each random input factor on the variance of outputs (i.e., productivity and product CQAs), identify the bottlenecks, and accelerate the reliable bioprocess specifications. Third, since the coefficients of interpretable bioprocess model or probabilistic knowledge graph 
are estimated from limited real-world process data, which induces model uncertainty (MU) or model risk (MR), we further propose Bayesian uncertainty quantification and Shapley value based model uncertainty sensitivity analysis to support process learning and faithfully assess the impact of estimation uncertainty from each set of model coefficients. \textit{Thus, our study can: (1) identify the bottlenecks of bioprocess; (2) accelerate the reliable process specifications and development to improve the production process stability and facilitate QbD; and (3) support the most ``informative" data collection to reduce the model risk of process probabilistic knowledge graph and improve the bioprocess understanding.} 


This paper is organized as follows. 
In Section~\ref{sec:background}, we review the related literature on biomanufacturing process modeling and PATs, Bayesian network, and process sensitivity analysis. 
In Section~\ref{sec:problem_Description}, we present the problem description and summarize the proposed framework.  In Section~\ref{sec:ProcessModeling}, we develop 
the Bayesian network (BN) based bioprocess probabilistic knowledge graph to characterize the risk- and science-based process understanding. We derive the Shapley value (SV) based bioprocess sensitivity analysis in Section~\ref{sec:RiskAnalysis} to support the process specifications, improve the production stability, and ensure product quality consistency. We further introduce the process model coefficient uncertainty quantification and Shapley value based sensitivity analysis studying the impact of each model coefficient estimation uncertainty on process risk analysis and CPPs/CQAs criticality assessment in Section~\ref{sec:sensitivityModelRisk}. We conduct the empirical study on the performance of our proposed framework in both simulation and real data analysis in Section~\ref{sec:empiricalStudy}, and then conclude with some discussion in Section~\ref{sec:conclusion}.


\section{Background}
\label{sec:background}


The Process Analytical Technologies (PAT) are defined as ``a system for designing, analyzing and controlling manufacturing through timely measurements of critical quality and performance attributes of raw and in-process materials and processes, with the goal of ensuring final product quality"; see \cite{fda2004guidance}. 
With the established process sensors and analyzers, such as near infrared spectroscopy, Raman spectrocopy, and multiwavelength fluorescence, various multivariate data analysis approaches 
have been used for bioprocess PATs, including principal component analysis (PCA) \citep{ayech2012new},
 partial least squares (PLS) \citep{de2010prediction},
clustering \citep{prinsloo2008acetone}, multilinear regression \citep{wechselberger2012efficient},
artificial neural network (ANN) \citep{li2018neural}, 
genetic algorithm \citep{sokolov2018sequential}, elastic net \citep{severson2015elastic}, support vector machines \citep{li2006prediction}, and root cause analysis \citep{borchert2019comparison}; 
see an overview in \cite{rathore2010process}. 
However, existing PAT approaches are usually based on generalized multivariate ``black-box" approaches quantifying the input-output relationship without incorporating the bioprocess mechanism 
information. 

On the other hand, OR/OM methodology development for biomanufacturing analysis and decision making is still in its infancy \citep{Kaminsky_Wang_2015}. Mixed integer linear programming \citep{lakhdar2007multiobjective,leachman2014automated}, 
dynamic lot size model \citep{fleischhacker2011planning}, and queueing network and simulation models \citep{lim2004decisional,kulkarni2015modular} 
have been developed to study resource planning, scheduling and material consumption in biomanufacturing. Those approaches focus on developing general methodologies without fully exploring the bio-technology domain knowledge (e.g., causal relationship, structural information of the bioprocess). Some recent works,  e.g., \cite{martagan2016optimal,martagan2017performance, martagan2018managing}, account for physical-chemical characteristics and biology-induced randomness in either fermentation or chromatography stage, and develop Markov decision models to optimize the corresponding operational policies.

For complex systems,
Bayesian network (BN) can be used to combine the expert knowledge with data and facilitate data integration and process analysis in various applications. 
For example, \cite{wang2018knowledge} proposed a BN based knowledge management system for additive manufacturing. \cite{troyanskaya2003bayesian} introduced a BN that combines evidence from gene co-expression and experimental data to predict whether two genes are functionally related. \cite{moullec2013toward} provided a BN approach for system architecture generation and evaluation, and 
\cite{telenko2014probabilistic} applied
probabilistic graphical model to study how the usage context factors, including human factors, situational factors, and product design factors, impact on the energy consumption of the lightweight vehicle to guide usage scenarios and vehicle designs.
{Furthermore, Bayesian posterior and belief propagation based risk assessment has been studied in information system security \citep{feng2014security}, water mains failure \citep{kabir2015evaluating}, and supply chain \citep{ojha2018bayesian}.
Motivated by these studies, we propose a Bayesian network for modeling the complex interdependence of production process parameters and bio-drug properties, which can fully utilize the structural knowledge and causal relationship, and integrate the data from end-to-end bioprocess.} 


Finally, we briefly discuss the related literature on sensitivity analysis; see the review \citep{borgonovo2016sensitivity}. 
The existing sensitivity analysis studies associated with Bayesian network 
tend to systematically vary one of network's parameter at a time while fixing the other parameters and then obtain analytic expressions for the sensitivity functions \citep{van2007sensitivity,castillo1997sensitivity}. 
In our case, we are interested in stochastic uncertainty contributed by each factor, which is closely related to global probabilistic sensitivity analysis. 
Existing literature on global sensitivity analysis can be divided into several categories, including: (1) regression based methods, e.g., \cite{helton1993uncertainty}, which use the standardized regression coefficients as sensitivity measure; (2) variance based methods \citep{wagner1995global,sobol1993sensitivity} 
which assess the contribution of each random input based on expected reduction in model output variance; (3) functional ANOVA decomposition \citep{rabitz1999general} which provides variance decomposition under independence through high dimensional model representation theory; (4) density-based methods \citep{zhai2014generalized} 
that directly quantify the output density without reference to a particular moment.
Since the commonly used variance-based sensitivity measures (i.e.,
first-order effects and total effects) fail to 
adequately account for probabilistic dependence of inputs and process structural interactions or interdependencies, \cite{owen2014sobol} introduced a new sensitivity measure based on the game theory, called the Shapley Value (SV).
\cite{song2016shapley} further analyzed this measure and proposed a Monte Carlo algorithm for the estimation of Shapley values. 
\cite{lundberg2017unified} proposed Shapley value based unified framework for interpreting predictions.
Inspired by these studies, building on the proposed BN-based bioprocess knowledge graph characterizing the process causal interdependencies, we introduce SV-based probabilistic sensitivity analysis to assess the contribution or criticality of each random input (e.g., CPP and CQA ) on the output variance, while accounting for the impact of model estimation uncertainty associated with each set of model coefficients.

\section{Problem Description and Proposed Framework}
\label{sec:problem_Description}

\textit{We create a probabilistic graph model characterizing the risk- and science-based understanding of causal interdependencies between bioprocess CPPs/CQAs, and then propose risk and sensitivity analyses for integrated biomanufacturing process, accounting for model uncertainty.} This study can: (1) provide a reliable guidance on process specification, CPPs/CQAs monitoring, and most informative data collection; (2) facilitate production stability control and quality-by-design (QbD); and (3) accelerate real-time release, speed up the time to market, and reduce the drug shortage.

\begin{figure}[h!]
	\centering
	\includegraphics[width=1\textwidth]{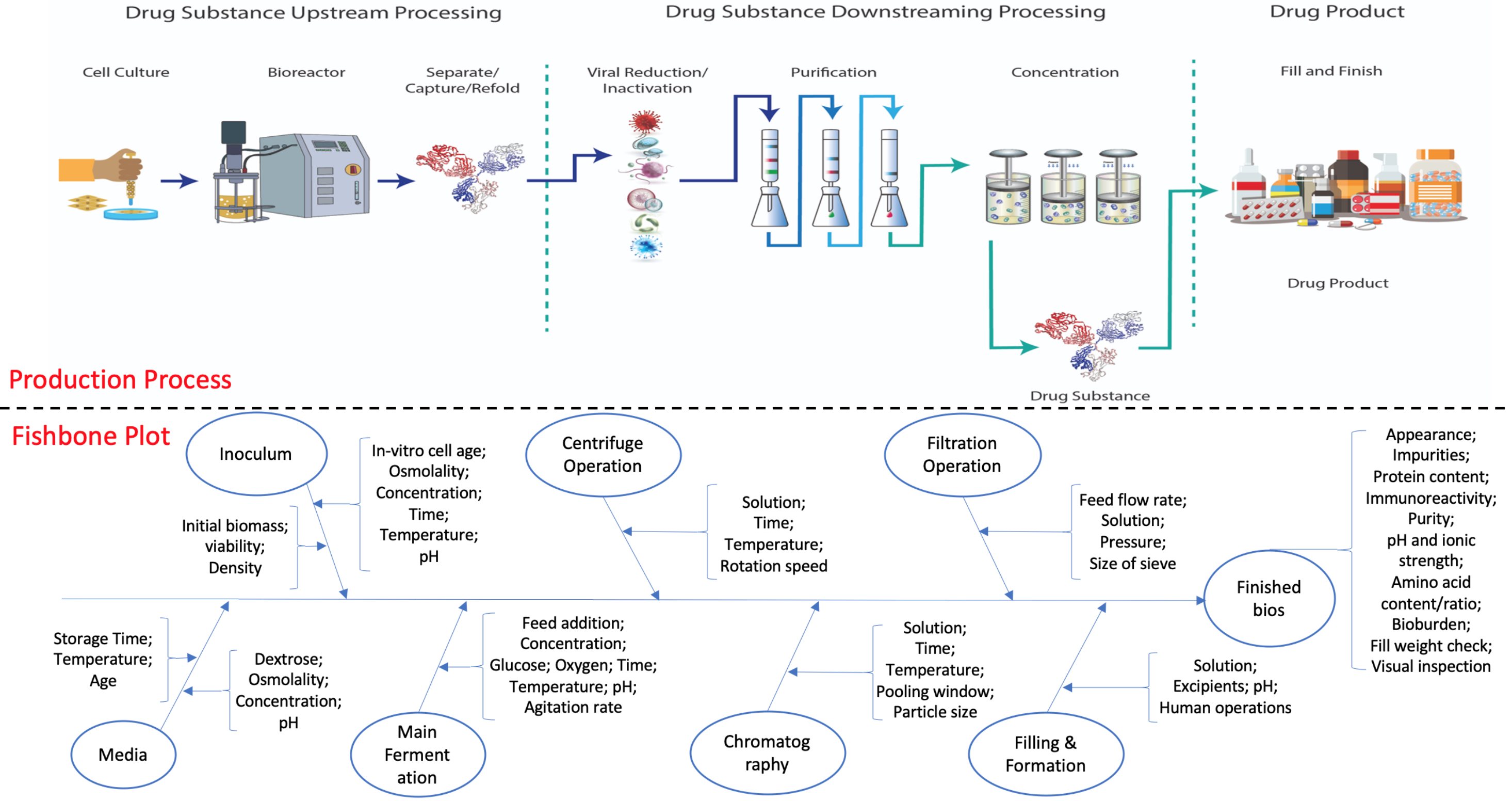}
	\caption{An illustration of general biomanufacturing process and fish-bone representation \citep{walsh2013pharmaceutical}.} \label{fig_CurrentAPI_production}
\end{figure}

An illustration of biomanufacturing process is provided in Fig.~\ref{fig_CurrentAPI_production} with a fish bone representation of bioprocess input factors introduced in each unit operation impacting on the outputs. The biomanufacturing process typically has several main unit operations, including: (1)~media preparation, (2) inoculum fermentation, (3) main fermentation, (4) centrifugation(s), (5) chromatography/purification, (6) filtration, (7) fill and finish, and (8) quality control. Steps (1)--(3) belong to upstream cell culture, Steps~(4)--(6) belong to downstream target protein purification, and Steps~(7)--(8) are for finished drug filling/formulation and final product quality control testing.

The \textit{interactions} of many factors impact the variability of outputs (e.g., drug quality, productivity). They can be divided into CPPs and CQAs in general; see the definitions of CPPs/CQAs in ICH Q8(R2) \cite{guideline2009pharmaceutical}.
\begin{itemize}
	\item[CPP:] At each process unit operation, CPPs are defined as critical process parameters whose \textit{variability} impacts on product CQAs, and therefore should be monitored and controlled to ensure the process produces the desired quality.
	
	\item[CQA:] A physical, chemical, biological, or microbiological property that should be within an appropriate limit, range, or distribution to ensure the desired {product} quality.
\end{itemize}
Since the raw material attributes are outputs of release materials, they should be considered along with CPPs as impacting process variability.

We represent the system output (e.g., product CQAs, productivity) with a random variable, denoted by $Y$, which depends on 
CPPs/CQAs inputs, denoted by $\mathbf{X}$, and other uncontrolled/uncontrollable input variables (e.g., contamination), modeled by residuals $\mathbf{e}$. We represent the impact of complex interactions of input factors $(\mathbf{X},\mathbf{e})$ throughout the production process on the response by ${Y} = g(\mathbf{X},\mathbf{e}|\pmb{\theta})$, where the unknown function $g(\mathbf{X},\mathbf{e}|\pmb{\theta})$, specified by model coefficients $\pmb{\theta}$, models the complex interactions of integrated bioprocess and characterizes the impact of random inputs $(\mathbf{X},\mathbf{e})$ on the output $Y$. For notation simplification, we consider the unit-variate response/output in the paper, and the proposed framework can be naturally extended to a vector of responses.

To provide the risk- and science-based production process understanding and guide the reliable process development, we need to 
correctly quantify all sources of uncertainties. There are {two types of uncertainty}: 
(1) bioprocess inherent \textit{stochastic uncertainty} from CPPs/CQAs and other uncontrolled variables (i.e., randomness of $\mathbf{X}$ and $\mathbf{e}$), which can be reduced by the identification of missed CPPs and tighter specification of selected CPPs; and 
(2) \textit{model uncertainty (MU)} 
(i.e., the estimation uncertainty of bioprocess model coefficients $\pmb{\theta}$), which can be reduced by collecting ``most informative" process observations.
\textit{Correctly quantifying all sources of uncertainty can facilitate learning, guide risk elimination/control, and improve robust, automatic, and reliable bioprocess decision making.} 


\subsection{Review of Game Theory based Sensitivity Measure - Shapley Value}
\label{subsec:ShapleyValue}

In game theory, the Shapley value (SV) was originally introduced to evaluate the contribution of a player in a cooperative game \cite{shapley1953value}. A cooperative game is defined as a set of players $\mathcal{K}=\{1, 2, \ldots, K\}$, with a function $c(\cdot)$ that maps a subset of players to its corresponding payoff, $c: 2^{\mathcal{K}} \to \mathbb{R}$ with $c(\emptyset) = 0$, where $2^{\mathcal{K}}$ denotes the power set of $\mathcal{K}$ (i.e., the set of all subsets of $\mathcal{K}$). Thus, $c(\mathcal{J})$ characterizes the total gain that the players in subset $\mathcal{J} \subset \mathcal{K}$ can
obtain by cooperation. The SV of player $k \in \mathcal{K}$ with respect to $c(\cdot)$ is defined by
\begin{equation}
	\mbox{Sh}_k 
	=\sum_{\mathcal{J}\subset \mathcal{K}/\{k\}}\dfrac{(K -|\mathcal{J}|-1)! |\mathcal{J}|!} {K!}\left[ c(\mathcal{J}\cup\{k\}) - c(\mathcal{J}) \right],
	\label{eq.Shapley_ori}
\end{equation}
where $K=|\mathcal{K}|$ is the total number of players and $|\mathcal{J}|$ is the size of subset $\mathcal{J}$ from $\mathcal{K}/\{k\}$. \textit{This SV can be interpreted as the average incremental payoff by including player $k$ over all possible cooperation group formations, i.e., $\mathcal{J}\subset \mathcal{K}/\{k\}$, and $\mbox{Sh}_k$ can be used to measure the contribution of the player $k$.} This assessment approach satisfies the ``efficiency property" that the sum of the SVs of all players equals the gain of the grand coalition, i.e., $c(\mathcal{K}) = \sum_{k=1}^K \mbox{Sh}_k$.

The Shapley value was recently introduced 
for global sensitivity analysis to measure the variance of output contributed by each random input 
\citep{owen2014sobol}. 
Denote the set of inputs as $\mathbf{U}_{\mathcal{K}} = \{U_1, U_2, \ldots, U_K\}$, and model the output $V = \eta(\mathbf{U}_{\mathcal{K}})$ as a function $\eta(\cdot)$ of the inputs, accounting for their interactions. Two most commonly used variance-based sensitivity measures are: (1) the first-order effect $O_k\equiv\mbox{Var}(V)-\mbox{E}[\mbox{Var}(V|U_k)]$ that considers the variance reduction when we fix $U_k$; and (2) the total effect $T_k\equiv\mbox{E}[\mbox{Var}(V|\mathbf{U}_{-k})]$ that considers the expected remaining variance when all other factors, denoted by $\mathbf{U}_{-k}$, are fixed.
However, both measures fail to appropriately quantify the sensitivity or variance contribution when there exist probabilistic interdependence among inputs and process structural interaction \citep{song2016shapley}.

Built on the SV from game theory, given a cooperative game with inputs $\mathbf{U}_{\mathcal{K}}$ as the players and the payoff as the incremental variance in output $V$ induced by any index subset $\mathcal{J} \subset \mathcal{K}$, one can define the payoff function as
\begin{equation}
	{ c(\mathcal{J}) = \mbox{Var}(V) -  \mbox{E}[\mbox{Var}[V|\pmb U_{\mathcal{J}}]]~~\mbox{or}~~} 
	c(\mathcal{J}) = \mbox{E}[\mbox{Var}[V|\pmb U_{-\mathcal{J}}]].
	\label{eq.costfun}
\end{equation}
Thus, \cite{owen2014sobol} introduced a new SV-based sensitivity measure, with $\mbox{Sh}_{U_k,V}$ computed by Equations~\eqref{eq.Shapley_ori} and  \eqref{eq.costfun}. In this paper, we use $c(\mathcal{J}) = \mbox{E}[\mbox{Var}[V|\pmb U_{-\mathcal{J}}]]$ in Equation~\eqref{eq.Shapley_ori}, which can simplify the computation of the contribution from any random input $U_k$ on the output variance $\mbox{Var}(V)$, $\mbox{Sh}_{U_k,V}=\mbox{Sh}_k$.
The SV-based sensitivity analysis overcomes the limitations of first-order effect and total effect measures by accounting for the interdependence of inputs and process interactions. 
The variance of output $V$ can be decomposed into the contribution from each random input $U_k$ and we can define the \textit{criticality} as the proportion of $\mbox{Var}(V)$ contributed from $U_k$, denoted by $p_{U_k,V}$, 
\begin{equation}
	\mbox{Var}(V) = \sum_{k=1}^K \mbox{Sh}_{U_k,V} ~~
	\mbox{     and    } ~~ p_{U_k,V} = \frac{\mbox{Sh}_{U_k,V}}{ \mbox{Var}(V)}.
	\nonumber 
\end{equation}

The main benefits of SV over first-order and total effect sensitivity measures include: (1) the uncertainty contributions sum up to total variance of output; and (2) SV can automatically account for probabilistic dependence and structural interactions occurring in the complex production process.
\subsection{Summary of Proposed Interpretable Bioprocess Model, Risk and Sensitivity Analyses for Integrated Bioprocess Stability Control}
\label{subsec:summaryProposedFramework}

\begin{figure}[h!]
	\centering
	\includegraphics[width=0.9\textwidth]{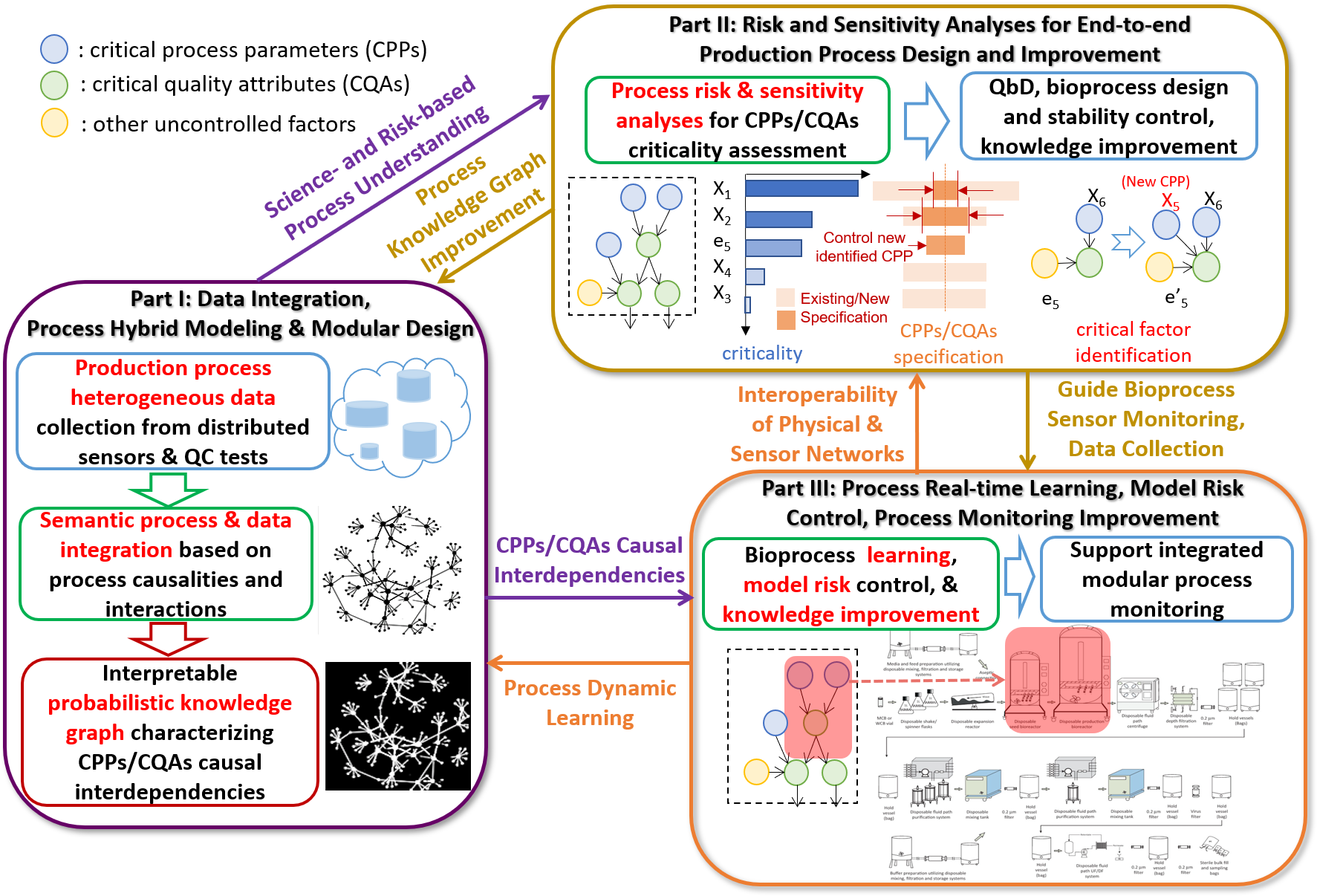}
	\vspace{-0.1in}
	\caption{The flowchart of proposed biomanufacturing process risk and sensitivity analyses framework.} \label{fig:flowchart}
\end{figure}

Fig.~\ref{fig:flowchart} provides the flowchart of proposed risk and sensitivity analyses framework, which can accelerate learning of the end-to-end production process and guide the development of stable biomanufacturing. 
Parts I and II focus on modeling and reducing of process stochastic uncertainty.  Part III focuses on analyzing and controlling the model risk. 
By exploring the causal relationships and interactions of CPPs/CQAs of raw materials/in-process materials/product within and between different process modules, in Section~\ref{sec:ProcessModeling}, we develop ontology based data integration and create an interpretable Bayesian network (BN) based bioprocess semantic probabilistic knowledge graph, specified by the model coefficients $\pmb{\theta}$. This knowledge graph can characterize the risk- and science-based understanding of integrated bioprocess and quantify the \textit{causal interdependencies} of inputs $(\mathbf{X},\mathbf{e})$ and output $Y$. 
	\textit{It is interpretable and extendable, which can support flexible process modular design, incorporate the  existing mechanisms from different modules and operation units, quantify the bioprocess causal interdependencies, and greatly reduce the dimensionality of bioprocess design space to guide the decision making.} %

\vspace{-.0in}

\begin{sloppypar}

	Building on this interpretable probabilistic knowledge graph, in Section~\ref{sec:RiskAnalysis}, we develop the SV-based process risk and sensitivity analyses studying \textit{stochastic uncertainty}, and derive variance decomposition to quantify the contribution from each random input,
	\[
	\mbox{Var}(Y|\pmb{\theta}) = \sum_{X_k } \mbox{Sh}_{X_k,Y}(\pmb{\theta} ) + \sum_{e_k} \mbox{Sh}_{e_k,Y}( \pmb{\theta} ), 
	\nonumber 
	\]
	where Shapley values, $\mbox{Sh}_{X_k,Y}(\pmb{\theta})$ and $\mbox{Sh}_{e_k,Y}(\pmb{\theta})$, 
	measure the contributions from any CPP/CQA, $X_k\in \mathbf{X}$, and residual factor, $e_k\in \mathbf{e}$ (representing the impact of remaining uncontrolled factors on the CQA $X_k$), to the output variance $\mbox{Var}(Y|\pmb{\theta})$.
	For any input factor $W_k$ (i.e., either $X_k$ or $e_k$), the \textit{criticality},  $p_{W_k,Y}(\pmb{\theta})\equiv \mbox{Sh}_{W_k,Y}(\pmb{\theta})/\mbox{Var}(Y|\pmb{\theta})$, can be used to identify the bottlenecks that contribute the most to $\mbox{Var}(Y)$, and guide the process specifications to efficiently improve production process stability. The CPPs/CQAs $X_k$ with high criticality requires more restrict stability control, while the residual $e_k$, with high impact on the output variance, can guide us to identify the ignored CPPs. 
	\textit{Since the Shapley value (SV) based sensitivity analysis is developed based on game theory, 
	its combination with bioprocess probabilistic knowledge graph, accounting for the complex causal interdependencies, can correctly assess the risk effect from each set of random input factors on the output variation.} 
	


	The ``correct" process model coefficients, denoted by $\pmb{\theta}^c$, characterizing the bioprocess underlying probabilistic interdependence, is unknown and estimated by using the real-world process data, denoted by $\mathcal{X}$. 
	Given limited historical process data, the model risk or estimation uncertainty can have a large impact on the bioprocess risk and sensitivity analyses. 
	Since the estimation uncertainty of model coefficients at different parts of bioprocess can be different and interdependent, the \textit{model uncertainty (MU)} 
	is quantified with the joint posterior distribution $p(\pmb{\theta}|\mathcal{X})$.
	\textit{We further develop the SV-based sensitivity analysis to study the impact of model uncertainty from each part of process in Section~\ref{sec:sensitivityModelRisk}}, 
	which can guide the ``most informative" data collection to reduce the impact from model risk and  efficiently improve the accuracy of bioprocess risk analysis and critiality assessment, especially for those factors contributing the most to the output variance.
	For any random input $W_k$, the Shapley value $\mbox{Sh}_{W_k,Y}$ is estimated with error, which can be contributed by the model coefficients located along the paths propagating the uncertainty of $W_k$ to the output $Y$, denoted by $\pmb{\theta}(W_k,Y)$. 
	We introduce the \textit{BN-SV-MU sensitivity analysis} to provide the comprehensive study over the impact of model uncertainty, 
	\begin{equation}
			\mbox{Var}^*[\mbox{Sh}_{W_k,Y}|\mathcal{X}] = \sum_{\theta_\ell \in \pmb{\theta}(W_k,Y)}
			\mbox{Sh}^*_{\theta_\ell}\left[
			\left. \mbox{Sh}_{W_k,Y}\left( \widetilde{\pmb{\theta}}(W_k,Y) \right) \right| \mathcal{X} \right]=
			\sum_{\theta_\ell \in \pmb{\theta}(W_k,Y)}
			\mbox{Sh}^*_{\theta_\ell}\left[ \left. \mbox{Sh}_{W_k,Y} \right| \mathcal{X}\right],
			\label{eq.VarStarDecomp2}
	\end{equation}
	where the subscript ``$*$" represents any measure calculated based on the posterior $p(\pmb{\theta}|\mathcal{X})$ and $\mbox{Sh}^*_{\theta_\ell}\left[ \left. \cdot \right| \mathcal{X}\right]$ measures the contribution from coefficient estimation uncertainty of $\theta_\ell \in \pmb{\theta}(W_k,Y)$. 
	In the proposed interpretable bioprocess model, $\theta_\ell$  
	can be interpreted as certain mechanistic coefficients (e.g., cell growth rate in the cell culture). 
	Thus, the decomposition in (\ref{eq.VarStarDecomp2}) provides the detailed information on how the model uncertainty of each part of integrated production process influences the estimation uncertainty of $\mbox{Sh}_{W_k,Y}$. 
\end{sloppypar}



\begin{figure}[h!]
	\centering
	\includegraphics[width=0.7\textwidth]{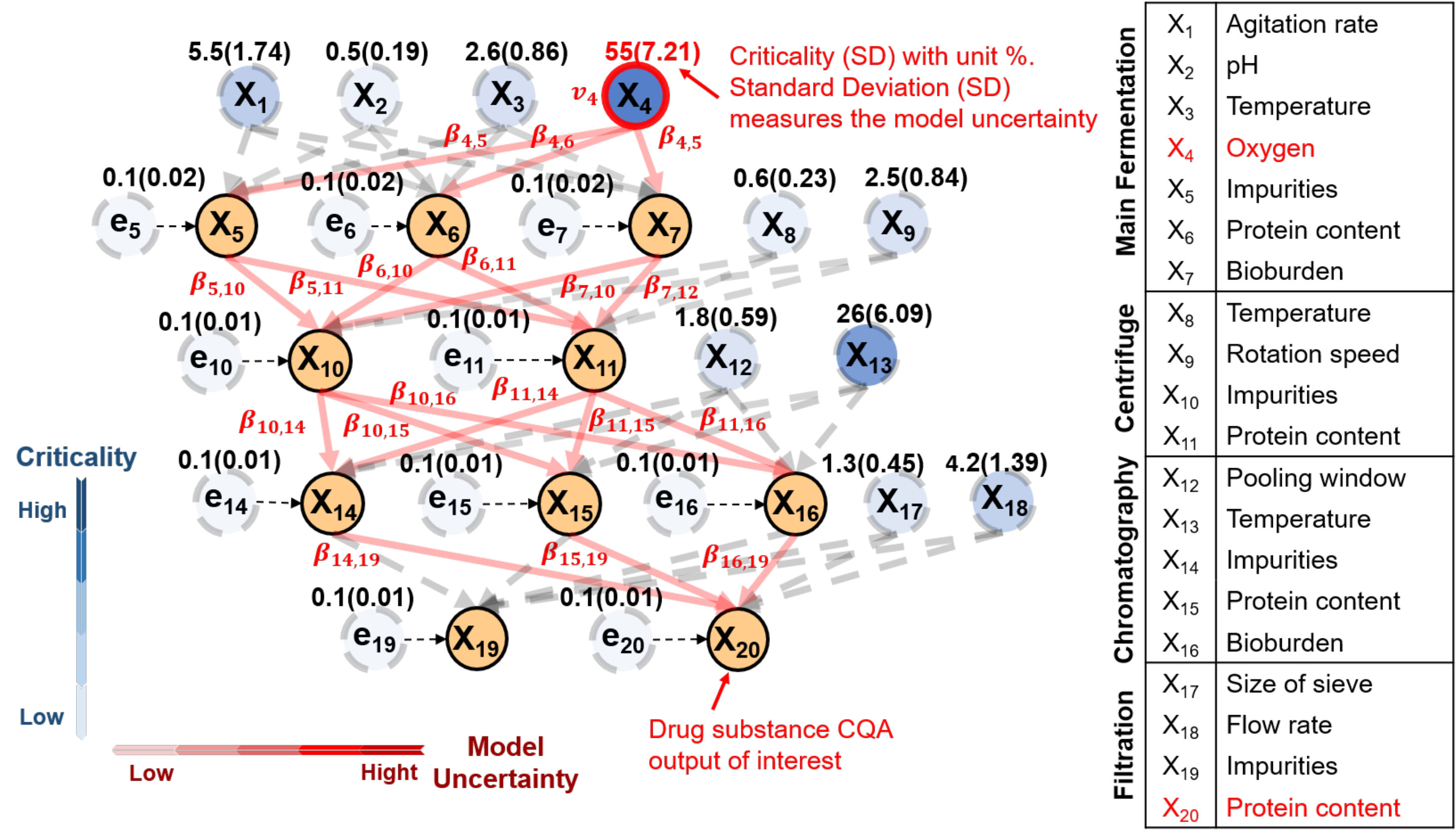}
	\vspace{-0.15in}
	\caption{An example illustration visualization of the integrated bioprocess sensitivity framework for criticality assessment and model uncertainty.} \label{fig:MR_example}
\end{figure}

To illustrate the key ideas of the proposed bioprocess risk and sensitivity analyses, we use a simplified monoclonal antibody (mAbs) drug substance production example, including main fermentation, centrifuge, chromatograph, and filtration; see the interactions in Fig.~\ref{fig:MR_example}. We consider the dominant CPPs/CQAs in each step, while the impacts of remaining factors are included in $\mathbf{e}$.
We are interested in the variance contribution (or criticality) from each CPP to drug substance protein content $Y=X_{20}$, and also account for the impact of model uncertainty on criticality assessment.

\vspace{-0in}

The results of risk and sensitivity analyses can be visualized along the graph model for this bioprocess example; see Fig.~\ref{fig:MR_example}. The process knowledge graph model is specified by coefficients,
$\pmb{\theta}= (\pmb{\mu}, \pmb{v}^2, \pmb{\beta})$, where $\pmb{\mu}$ and $\pmb{v}^2$ are the mean and variance vectors of all factors (listed in the table), and $\pmb{\beta}$ quantifies the effects from parents' nodes on each child node.
The darkness of nodes indicates the criticality level of process factors, which can guide the better bioprocess specification. The results show that $X_4$ 
contributes to $55\%$ variance of $X_{20}$. In addition, the darkness of directed edges and circle boundaries indicates the distribution of model uncertainty in the process. For instance, the variance of dissolved oxygen in bioreactor, $v_4^2$, has dominant model uncertainty impact on the estimation uncertainty of criticality $p_{X_4,Y}$, which suggests additional data  should be collected to improve the estimation of $v_4^2$. We will revisit this example and use it to study the performance of proposed framework in Section~\ref{subsec:simulationEmpiricalStudy}.

\section{Interpretable Bioprocess Probabilistic Model Development}
\label{sec:ProcessModeling}

We develop an interpretable bioprocess probabilistic model, which can be extendable to end-to-end biomanfacturing supply chain and support evidence-based biopharmaceutical production process development. The model can incorporate the existing mechanism models from each module and unit operation, and facilitate the learning from distributed and heterogeneous process data.
In this section, 
we develop a Bayesian network (BN) based interpretable bioprocess probabistic knowledge graph, which can characterize the complex CPPs/CQAs causal interdependencies and support flexible modular biomanfuacturing development.  




By exploring the causal relationships and interactions, we consider \textit{bioprocess ontology-based data integration}, which can connect all distributed and heterogeneous data collected from bioprocess; see more description in Online Appendix~\ref{subsec:relationalGraph}.
This relational graph can enable the connectivity of  end-to-end process.
Nodes represent factors (i.e., CPPs/CQAs, media feed, bioreactor operating conditions, other uncontrolled factors) impacting the process outputs, and the directed edges model the causal relationships. 
Within each module, which can be each phase of cell culture process (such as cell growth and production phases) or each unit operation, we can model complex interactions, e.g., biological/physical/chemical interactions.
In the relational graph, the shaded nodes represent the variables with real-world observations, including the testing and sensor monitoring data of CPPs/CQAs for raw materials, operation conditions, and intermediate/final drug products. The unshaded 
nodes represent variables without observations and residuals, including 
quality status of intermediate and final drug products, and other uncontrollable factors (e.g., contamination) introduced during the process unit operations.
Since bio-products have very complex structures, we cannot observe the underlying complete quality status and the monitoring of CQAs can carry partial information. 
Building on the bioprocess relational graph, we develop a BN based probabilistic graphical model composing of random CPPs/CQAs/residuals factors and their conditional dependencies via directed edges. 
\textit{It can characterize the probabilistic causal interdependencies among all factors of integrated bioprocess.} 

\begin{figure}[h!]
	\centering
	\includegraphics[width=0.88\textwidth]{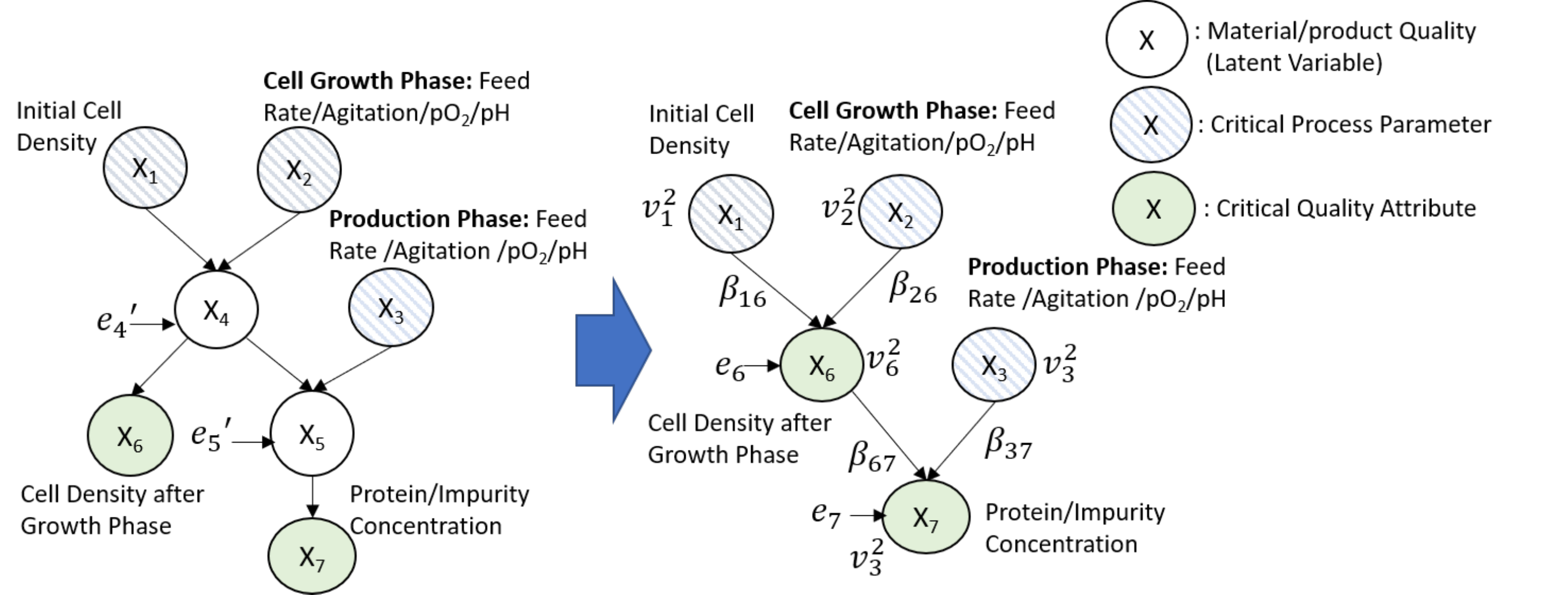}
	\vspace{-0.1in}
	\caption{Left: knowledge relational graph; Right: simplified knowledge graph.} \label{fig_2}
\end{figure}

To make it easy to follow, we first provide a simple illustration example of cell culture process, including two phases, to present the key ideas of \textit{modular bioprocess modeling}, 
and then develop the complete probabilistic knowledge graph model for general integrated bioprocess. 
Specifically, we use a simple bioreactor fermentation example with two phases (i.e., cell growth and production phases; see Fig.~\ref{fig_2}) to illustrate the probabilistic graphical model development. 
It is based on the causal relationships and  interactions between CPPs and CQAs. 
Each node represents a CPP/CQA with a random variable $X$ modeling its variability. Each directed edge 
represents the causal impact of parent node $X_i$ on child node $X_j$. 
The pattern-fill nodes ($X_1,X_2,X_3$) represent the CPPs. 
The solid fill nodes ($X_6,X_7$) represent the monitored CQAs of intermediate materials and drug products. 
The nodes $X_4$ and $X_5$ represent the underlying status of working cells after cell growth phase and the protein/impurity structure after cell production phase. 
The CQAs $X_6$ and $X_7$ represent the partial information of quality variables $X_4$ and $X_5$. Except the CPPs $X_1,X_2,X_3$, the impacts from other uncontrolled factors introduced during two phases of cell culture are modeled through $e_4^\prime$ and $e_5^\prime$.

Since it is hard to uniquely specify the underlying cells/proteins with very complex structures, $X_4$ and $X_5$ are hidden, which can lead to an identification issue. Typically, the non-identifiable BN with hidden nodes is transformed to the equivalent BN by structural simplification to avoid analytical issues (see Chapter 19 in \cite{koller2009probabilistic}).
Thus, we simplify and transform the relational graphical model to a graph without hidden nodes, depicted in the right panel of Fig.~\ref{fig_2}.
The new residual $e_6$ in updated graph accounts for both original residual $e^\prime_4$ and also the uncertainty of underlying cell health status, $X_4$, impacting on CQA $X_6$, similar for new residual $e_7$. 
According to the right plot in Fig.~\ref{fig_2}, the sources of bioprocess \textit{stochastic uncertainty} impacting on the variability of $X_7$ include CPPs, $(X_1,X_2,X_3)$, and other factors with the impact represented by residuals $(e_6,e_7)$. 
Thus, we have CPPs $X_1,X_2$ as inputs and CQA $X_6$ as output for the first cell growth phase, and have CQA $X_6$ and CPP $X_3$ as inputs and $X_7$ as output for the second protein production phase.
To study the impact of each CPP on the CQA of interest (i.e., $X_6$ and $X_7$), we can decompose the variance of $X_6$ and $X_7$ into the contributions from $X_1$, $X_2$ and $X_3$, and remaining parts coming from $e_6$ and $e_7$; see the process risk and sensitivity analyses in Section~\ref{sec:RiskAnalysis}. In this way, we can identify the main sources of uncertainty and quantify their impacts, which can guide the CPPs/CQAs specifications and the quality control to improve the product quality stability.



Now we describe the BN-based bioprocess model for general situations. Suppose that the integrated bioprocess can be represented by a probabilistic graphical model with $m+1$ nodes: $m$ process factors (denoted by $\mathbf{X}$) and a single response, denoted by $Y$, such as the impurity concentration or protein content.
Let the first $m^p$ nodes representing CPPs $\mathbf{X}^p=\{X_1, X_2, \ldots, X_{m^p}\}$, the next $m^a$ nodes representing CQAs $\mathbf{X}^a=\{X_{m^p+1}, X_{m^p+2}, \ldots, X_{m}\}$, and the last node representing the response $Y\triangleq X_{m+1}$ with $m=m^p+m^a$.
\textit{The modular bioprocess probabilistic knowledge graph can be modeled by 
marginal and conditional distributions of each node} as follows:
\begin{eqnarray}
	\small
	X_k&\sim&  \mathcal{N}(\mu_k, v_k^2)
	\mbox{ for CPP $X_k$ with $k=1,2,\ldots,m^p$,}
	\label{eq.CPP}  \\
	X_k &=& f(Pa({X}_k); \pmb{\theta}_{k}) + e_k 
	\mbox{ for CQA $X_k$ with $k=m^p+1, \ldots, m+1$}
	\label{eq.CQA5}
\end{eqnarray}
where $\mathcal{N}(u,v^2)$ denotes the normal distribution with mean $u$ and variance $v^2$, and $Pa(X_k)$ denotes the parent nodes of $X_k$. 
By applying central limit theory (CLT),
we assume that the residual $e_k \sim \mathcal{N}(0, v_k^2)$ with the conditional variance $v_k^2\equiv \mbox{Var}[X_k|Pa(X_k)]$.  Since the amount of real-world bioprocess batch data is often very limited, Gaussian distribution is used to model the variability of each variable or node, which is often used in the existing biopharmaceutical studies (see for example \cite{coleman2006retrospective}). It also makes the process risk and sensitivity analyses tractable.

\textit{The proposed probabilistic knowledge graph is a hybrid model of integrated bioprocess, which can leverage the existing mechanisms and learn from real-world process data.}
Basically, the prior of the function $f(\cdot)$ in a generalized regression model (\ref{eq.CQA5}) can be specified based on the existing knowledge on underlying bioprocess mechanisms (e.g., biophysicochemical kinetics) within each module of bioprocess; see for example \cite{kyriakopoulos2018kinetic,lu2018sparse,doran1995bioprocess}.
The unknown {model coefficients} $\pmb{\theta}_k$ (e.g., cell growth rate, media consumption rate) 
need be estimated from process data. 
In this paper, 
we consider linear function accounting for the main effects, i.e.,
\begin{equation}
	X_k = \mu_k + \sum_{X_j\in Pa(X_k)}\beta_{jk}(X_j - \mu_j) + e_k 
	\mbox{ for CQA $X_k$ with $k=m^p+1, \ldots, m+1$}
	\label{eq.CQA}
\end{equation}
where the coefficient $\beta_{jk}$ can be used to measure the effect from the parent node $X_j$ to child node $X_k$.

Here, we use some illustrative examples to briefly show how
the proposed Bayesian network based process probabilistic model allows us to incorporate the existing bioprocess mechanisms.
We first consider the cell exponential growth mechanism for the fermentation step, $x=x_0e^{\mu t}$, where $x_0$ and $x$ denote the starting and ending cell densities, and $\mu$ is the unknown growth rate. 
This is a commonly used mechanism model in biomanufacturing industry; see more information in \cite{Pauline_2013}.
Suppose that there is a fixed cell culture duration $t$. 
By doing the log transformation and setting $X_k=\log(x_0)$, $X_{k+1} = \log(x)$ and $\beta_0= \mu t$, we can take the exponential growth mechanism as prior and get the hybrid probabilistic model for the exponential growth phase in fermentation or cell culture process, $X_{k+1} = \beta_0 +X_k + e_k$, where $e_k$ represents the residual term characterizing the integrated effect from many other factors and it follows a Gaussian distribution by following CLT.
Notice that it is a special case of BN-based process model~(\ref{eq.CQA}).
The similar idea can be applied to the situations where we have PDE/ODE-based bioprocess kinetics mechanism models, 
\begin{equation}
 \frac{d}{dt} x(t) = f(x(t);\theta_t) 
\approx  \frac{x(t_{k+1})-x(t_k)}{t_{k+1}-t_k} = f(x(t_k);\theta_{t_k}), 
\nonumber 
\end{equation}
where $x(t)$ can represent the concentrations of protein and metabolite waste at time $t$ and $\theta_t$ can denote the nonstationary growth rate. 
We can take the existing mechanism model as the prior knowledge of production process. By applying the finite difference on the gradient $dx(t)/dt$ and first-order Taylor approximation on function $f(\cdot)$, we can construct a probabilistic hybrid model matching with the formula in Equation~(\ref{eq.CQA}), which can leverage the information from existing PDE/ODE-based bioprocess kinetics mechanism models. This approximation can be very accurate if the data are collected from real-time production process sensor monitoring with high sampling frequency.

\begin{sloppypar}
	The complex bioprocess CPPs/CQAs causal interdependencies are characterized by the BN-based probabilistic knowledge graph.
	Given the model parameters $\pmb \theta = (\pmb \mu, \pmb v^2, \pmb \beta)$ with mean $\pmb \mu=(\mu_1, \ldots, \mu_{m+1})^\top$, conditional variance $\pmb v^2=(v_1^2, \ldots, v_{m+1}^2)^\top$, and linear coefficients $\pmb \beta = \{\beta_{jk}; k = m^p+1,\ldots,m+1 \mbox{ and } X_j\in Pa(X_k)\}$, 
	the conditional distribution for each CQA node $X_k$ becomes,
	\begin{equation}
		p(X_k|{Pa(X_k)}) = \mathcal{N}\left(\mu_k + \sum_{X_j\in Pa(X_k)}\beta_{jk}(X_j - \mu_j), v_k^2\right) \mbox{ for }k = m^p+1 \ldots,m+1. \nonumber 
	\end{equation}
	For any CPP node $X_k$ without parent nodes, $Pa(X_k)$ is an empty set and $P(X_k|Pa(X_k))$ is just the marginal distribution $P(X_k)$ in (\ref{eq.CPP}).
	\textit{Therefore, the joint distribution characterizing the interdependencies of CPPs and CQAs involved in the production process can be written as
	$p(X_1, X_2, \ldots, X_{m+1}) = \prod_{k=1}^{m+1}p(X_k|{Pa(X_k)}) 
	$.}
\end{sloppypar}


\section{Process Risk and Sensitivity Analyses} 
\label{sec:RiskAnalysis}

Given the bioprocess probabilistic knowledge graph specified by the coefficients $\pmb \theta = (\pmb \mu, \pmb v^2, \pmb \beta)$, \textit{we develop the BN-SV based sensitivity analysis for integrated production process and quantify the criticality of each random input factor measuring its contribution to the output variance $\mbox{Var}(X_{m+1})$}. This study can guide the CPPs/CQAs specification and improve the production process stability. To make it easy to follow, we start with a simple illustration example, provide the general process risk and sensitivity analyses, and then present the algorithm at the end of this section. 

\begin{figure}[h!]
	\centering
	\includegraphics[width=0.6\textwidth]{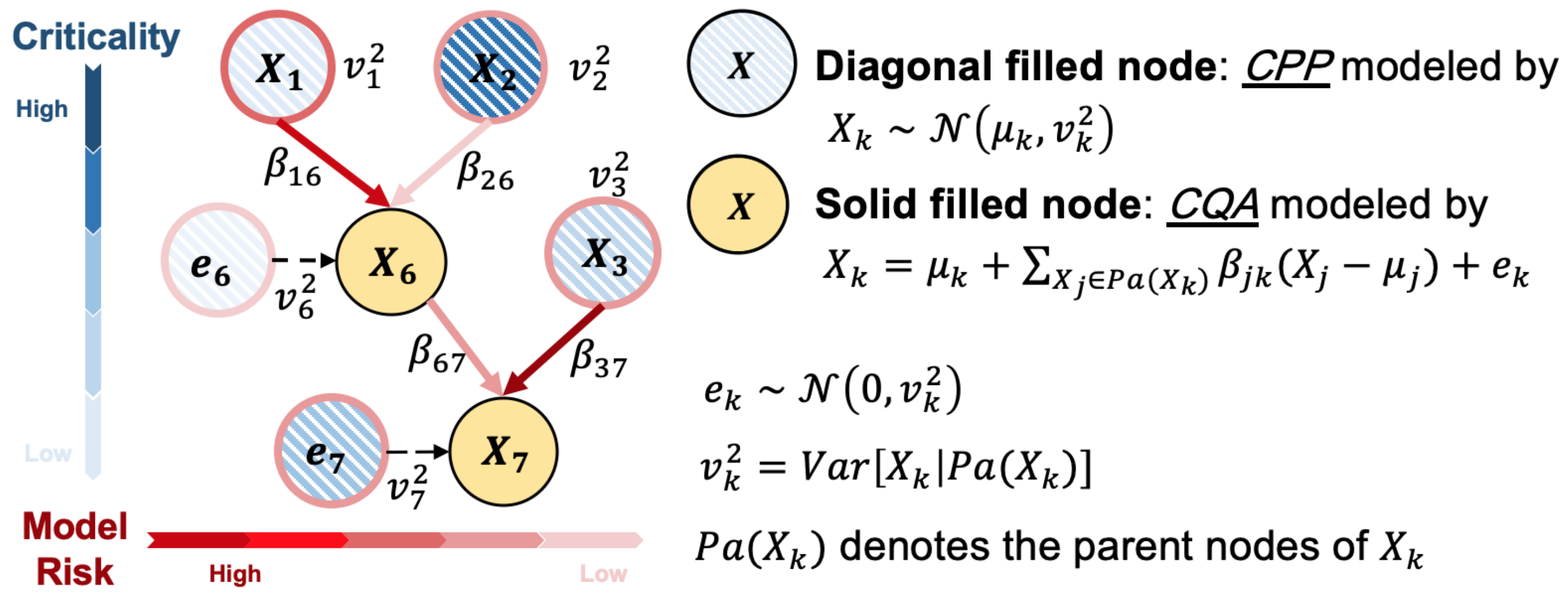}
	\caption{A simple example to illustrate BN-SV based process risk and sensitivity analysis.}\label{fig:sensitivityEx}
\end{figure}

We again use the simple example in Fig.~\ref{fig_2}
to illustrate the results of proposed BN-SV based bioprocess risk and sensitivity analyses, which can decompose the output variance of 
protein/impurity concentration $X_7$ after fermentation process
to each random input -- including $X_1$, $X_2$, $X_3$, $e_6$, $e_7$ -- as
\begin{equation}
	\resizebox{0.92\hsize}{!}{%
	$Var(X_7|\pmb \theta) = \underbrace{(\beta_{16}\beta_{67})^2 v_{1}^2}_{\text{contribution from $X_1$}}+ \underbrace{(\beta_{26}\beta_{67})^2 v_{2}^2}_{\text{contribution from $X_2$}} + \underbrace{\beta_{37}^2 v_{3}^2}_{\text{contribution from $X_3$}} + \underbrace{\beta_{67}^2 v_{6}^2}_{\text{contribution from $e_6$}} + \underbrace{v_{7}^2}_{\text{contribution from $e_7$}}.$%
} \label{eq.VD_example}
\end{equation}
The variance contribution from each random input, denoted by $W_k$ (i.e., $X_1,X_2,X_3,e_6,e_7$), depends on its variance $v^2_k$ and the product of coefficients $\pmb{\beta}$ located along the paths propagating the uncertainty from $W_k$ to the output $X_7$; see 
Fig.~\ref{fig:sensitivityEx}.
The darker blue 
filled node (i.e., cell growth phase CPP $X_2$, feed rate) contributes more to the output variance and has higher criticality. Thus, to efficiently reduce the output variance, it requires more restrictive stability control. 
The high impact of $e_7$ (with darker color) can guide us to identify unrecognized or missed CPPs.

This simple example illustrates that the proposed production process BN-SV risk and sensitivity analyses and the CPPs/CQAs criticality assessment are based on the bioprocess probabilistic knowledge graph, characterizing the complex CPPs/CQAs causal interdependencies and accounting for all sources of process inherent uncertainty, which can (1) guide the process specifications; (2) improve the product quality consistency and bioprocess stability; and (3) advance the risk- and science-based understanding on bioprocess.


\begin{sloppypar}
	Now we present the general process risk and sensitivity analyses.
	We first derive the Shapley value (SV) quantifying the contribution of each random input factor from CPPs $\mathbf{X}^p$ and other factors $\mathbf{e}$ to $\mbox{Var}(X_{m+1})$, which accounts for cases with dependent input factors. 
	According to the Gaussian BN model presented in (\ref{eq.CPP}) and (\ref{eq.CQA}), 
	we can write
	\begin{equation}
		X_{m+1} = \mu_{m+1} + \sum_{k=1}^{m^p}\gamma_{k,m+1}(X_k - \mu_k) + \sum_{k=m^p+1}^{m+1}\gamma_{k,m+1} e_k, \label{eq.represent}
	\end{equation}
	where the weight coefficient of any CPP $X_k$ to CQA $X_n$ with $k\le m^p<n\leq m+1$,
	\begin{equation}
		\gamma_{kn} = \beta_{kn} + \sum_{m^p<\ell<n} \beta_{k\ell}\beta_{\ell n} + \sum_{m^p<\ell_1<\ell_2<n} \beta_{k\ell_1}\beta_{\ell_1\ell_2}\beta_{\ell_2 n} + \ldots + \beta_{k,m^p+1}\beta_{m^p+1,m^p+2}\ldots \beta_{n-1,n}, 
		\label{eq.weightY}   
	\end{equation}
	the weight coefficient of any $e_k$ to a CQA node $X_n$ with $m^p<k<n\leq m+1$,
	\begin{equation}
		\gamma_{kn} = \beta_{kn} + \sum_{k<\ell<n} \beta_{k\ell}\beta_{\ell n} + \sum_{k<\ell_1<\ell_2<n} \beta_{k\ell_1}\beta_{\ell_1\ell_2}\beta_{\ell_2 n} + \ldots + \beta_{k,k+1}\beta_{k+1,k+2}\ldots \beta_{n-1,n}; 
		\label{eq.weightY2}   
	\end{equation}
	and $\gamma_{nn} = 1$ for any $n$; see the derivation for (\ref{eq.represent}) in Appendix \ref{sec:appendix1}.
	The weight coefficient $\gamma_{kn}$ is the product sum of $\pmb{\beta}$ located along the paths from node $X_k$ to node $X_n$ in the graph model. 
	Let $\pmb W = \{X_1,\ldots,X_{m^p}, e_{m^p+1}, \ldots, e_{m+1}\} \triangleq \{W_1, W_2, \ldots, W_{m+1}\}$ represent all random input factors, 
	with the index set $\mathcal{K} = \{1, 2, \ldots, m+1 \}$.
	Then, the SV for the $k$-th factor $W_k$ is,
	\begin{equation}
		\mbox{Sh}_{W_k,X_{m+1}} = \sum_{\mathcal{J}\subset \mathcal{K}/\{k\}}\dfrac{(m -|\mathcal{J}|)! |\mathcal{J}|!} {(m+1)!}\left[ c(\mathcal{J}\cup\{k\}) - c(\mathcal{J}) \right].
		\nonumber 
	\end{equation}
	Based on \eqref{eq.represent}, we compute the cost function,
	\[
	c(\mathcal{J}) = \mbox{E}[\mbox{Var}[X_{i}|\pmb W_{-\mathcal{J}}]]  = \sum_{k\in\mathcal{J}}\gamma_{k,m+1}^2 \mbox{Var}(W_k) + 2\sum_{k_1<k_2 \in \mathcal{J} }\gamma_{k_1,m+1}\gamma_{k_2,m+1}\mbox{Cov}(W_{k_1},W_{k_2}).
	\]
The random input factors, $\pmb{W}=(X_1,\ldots,X_{m^p},e_{m^p+1},\ldots,e_{m+1})$, including CPPs and residual terms introduced at each CQA nodes, are often independent as the real biomanufacturing process specification is often based on each CPP or CQA. To make the proposed framework general, we consider the potential interdependence between some inputs $W_{k_1}$ and $W_{k_2}$ with $k_1\neq k_2$, and the covariance $\mbox{Cov}(W_{k_1},W_{k_2})$ can be estimated by using the process data.

	Then, for each $W_k$ and $\mathcal{J} \subset \mathcal{K}/\{k\}$, we can obtain
	\begin{equation}
		c(\mathcal{J}\cup\{k\}) - c(\mathcal{J}) = \gamma_{k,m+1}^2\mbox{Var}(W_k) + 2\sum_{\ell \in \mathcal{J}}\gamma_{k ,m+1}\gamma_{\ell,m+1}\mbox{Cov}(W_k, W_\ell).
		\nonumber 
	\end{equation}
	Given the BN-based bioprocess knowledge graph model parameters $\pmb{\theta}$, by applying (\ref{eq.Shapley_ori}), we can derive the Shapley value, $\mbox{Sh}_{W_k,X_{m+1}}(\pmb{\theta})$, characterizing the contribution from any input factor $W_k$ to the output variance,
	\begin{eqnarray}
		\mbox{Sh}_{W_k,X_{m+1}}(\pmb{\theta}) = 
		\gamma_{k,m+1}^2\mbox{Var}(W_k) + \sum_{\ell \neq k}\gamma_{k, m+1}\gamma_{\ell, m+1}\mbox{Cov}(W_k, W_\ell).  \label{eq.shapley_3}
	\end{eqnarray}
	The derivation of (\ref{eq.shapley_3}) is provided in Appendix \ref{sec:appendix2}. Therefore, we can decompose the variance of output $X_{m+1}$ and estimate the contribution from each random input from $\mathbf{X}^p$ and $\mathbf{e}$,
	\begin{equation}
		\mbox{Var}(X_{m+1}|\pmb{\theta}) = \sum \mbox{Sh}_{W_k,X_{m+1}}(\pmb{\theta})
		= \sum_{k=1}^{m^p}\mbox{Sh}_{X_k,X_{m+1}}(\pmb{\theta}) + \sum_{k=m^p+1}^{m+1}\mbox{Sh}_{e_k,X_{m+1}}(\pmb{\theta})
		. \label{eq.VarDecompCPP}
	\end{equation}
	Equation~(\ref{eq.VarDecompCPP}) can be used to identify the dominant factors in $\mathbf{X}^p$ and $\mathbf{e}$ contributing the most to the output variance, which can guide the CPPs identification and process specification to improve the process stability and quality consistency. 
	As a result, 
	the \textit{criticality} of any input factor $W_k$ can be calculated as $
	p_{W_k,X_{m+1}} (\pmb{\theta})
	\equiv {\mbox{Sh}_{W_k,X_{m+1}}(\pmb{\theta})}/
	{\mbox{Var}(X_{m+1}|\pmb{\theta})}$. Notice that for any independent input factor $W_k$, the SV in Equation~\eqref{eq.shapley_3} is reduced to $\mbox{Sh}_{W_k,X_{m+1}}(\pmb{\theta})= \gamma_{k,m+1}^2 v_k^2$. Under the case that all input factors $\pmb W$ are mutually independent, the variance decomposition Equation~(\ref{eq.VarDecompCPP}) can be written as $\mbox{Var}(X_{m+1}|\pmb{\theta}) = \sum \gamma_{k,m+1}^2 v_k^2$, which gives the example results in Equation~\eqref{eq.VD_example}.

	This risk and sensitivity analyses can be  applied to any part of production process including one or multiple modules. Under this situation, the input factors $\pmb W$ include those nodes without parent node within the considered range of production (i.e., CPPs, CQAs or uncontrolled factors), and output of interest $X_i$ is certain CQA at the end of the procedure. For example, in Fig.~\ref{fig:sensitivityEx}, we consider the subgraph, including $\{X_3, X_6, X_7\}$, for cell production phase with the starting CQA $X_6$ carrying the information from previous cell growth phase. We can study the impacts of $X_6$ and CPP $X_3$ on the variability of CQA $X_7$. The SV of any input $W_k$ and the variance decomposition of $X_i$, still follow Equations~\eqref{eq.shapley_3} and (\ref{eq.VarDecompCPP}) by replacing the output $X_{m+1}$ with $X_i$. The criticality of $W_k$ on $X_i$ can be measured by proportion $p_{W_k,X_i}(\pmb{\theta})=\mbox{Sh}_{W_k,X_{i}}(\pmb{\theta})/\mbox{Var}(X_i|\pmb{\theta})$.
	
\end{sloppypar}

Given the BN parameters $\pmb{\theta}$, we summarize the procedure for production process BN-SV based sensitivity analysis in Algorithm~\ref{alg:procedure1}, in which we consider several consecutive operation steps, and our objective is to quantify the contribution of each random factor in $\pmb{W}$ to $\mbox{Var}(X_i|\pmb{\theta})$. 
\begin{algorithm}[hbt!]
	\small
	\caption{Procedure for Production Process BN-SV based Sensitivity Analysis} \label{alg:procedure1}
	\KwIn{BN parameters $\pmb{\theta}$, group of input factors $\pmb{W}$, response node $X_{m+1}$.}
	\KwOut{Variance decomposition of $X_i$ in terms of all random inputs within $\pmb{W}$.}
	
	
	
	(1) Calculate the Shapley value $\mbox{Sh}_{W_k,X_{m+1}}(\pmb{\theta})$ 
	by using Equation~\eqref{eq.shapley_3}, which measures the contribution from $W_k$ to the variance of response CQA $X_{m+1}$\;
	
	(2) Provide the variance decomposition of $\mbox{Var}(X_{m+1}|\pmb{\theta})$ by using Equation~\eqref{eq.VarDecompCPP}, and obtain the criticality of $W_k$ on the variance of $X_{m+1}$:  $p_{W_k,X_{m+1}}(\pmb{\theta})=\mbox{Sh}_{W_k,X_{m+1}}(\pmb{\theta})/\mbox{Var}(X_{m+1}|\pmb{\theta})$.
	
\end{algorithm}
\vspace{-0.2in}

\vspace{-0.in}
\section{Sensitivity Analysis for Model Risk Reduction}
\label{sec:sensitivityModelRisk}

Since the underlying true process model coefficients $\pmb{\theta}^c$ are unknown, given finite real-world data $\mathcal{X}$, there exists the model uncertainty (MU) characterizing our limited knowledge on 
the probabilistic interdependence of integrated bioprocess.
To study the impact of MU on the production process risk and sensitivity analyses for stochastic uncertainty and further assess CPPs/CQAs criticality,
we propose the \textit{BN-SV-MU based uncertainty quantification and sensitivity analysis}, which can guide the process monitoring and ``most informative" data collection. In Section~\ref{subsec:BayesianLearning}, we develop the posterior $p(\pmb{\theta}|\mathcal{X})$ and a Gibbs sampler to generate posterior samples, $\widetilde{\pmb{\theta}}^{(b)}\sim p(\pmb{\theta}|\mathcal{X})$ with $b=1,2,\ldots,B$, quantifying the model uncertainty, and then we quantify the overall impact of model uncertainty on the process risk analysis and CPPs/CQAs criticality assessment. 
In Section~\ref{subsec:SA}, we propose the BN-SV-MU based sensitivity analysis, which can study the impact of each model coefficient(s) estimation uncertainty on the process risk analysis and criticality assessment; 
see the result visualization in Fig.~\ref{fig:MR_example}. 


\subsection{Bayesian Learning and Model Uncertainty Quantification}
\label{subsec:BayesianLearning}

\begin{sloppypar}
	We consider the case with $R$ batches of complete production process data, denoted as $\mathcal{X} = \{ (x_1^{(r)}, x_2^{(r)}, \ldots, x_{m+1}^{(r)}), r=1,2,\ldots,R\}$.
	Without strong prior information, we consider the following conjugate (vague) prior (with initial hyperparameters giving relatively flat density), 
	\begin{equation}
		p(\pmb \mu, \pmb v^2, \pmb \beta) = \prod_{i=1}^{m+1}p(\mu_i)p(v_i^2)\cdot \prod_{i\neq j} p(\beta_{ij}), 
		\label{eq.prior}
	\end{equation}
	with
	$
	p(\mu_i) = \mathcal{N}(\mu_i^{(0)}, \sigma_i^{(0)2}),
	p(v_i^2) = \mbox{Inv-}\Gamma\left(\dfrac{\kappa_i^{(0)}}{2}, \dfrac{\lambda_i^{(0)}}{2}\right)$ and $
	p(\beta_{ij}) = \mathcal{N}(\theta_{ij}^{(0)}, \tau_{ij}^{(0)2})$, where $\mbox{Inv-}\Gamma$ denotes the inverse-gamma distribution.
	Given the data $\mathcal{X}$, by applying the Bayes' rule, we can obtain the posterior distribution 
	\begin{equation}
		p(\pmb \mu, \pmb v^2, \pmb \beta|\mathcal{X}) \propto \prod_{r=1}^{R}\left[\prod_{i=1}^{m+1}p(x_i^{(r)} | x_{Pa(X_i)}^{(r)})\right] p(\pmb \mu, \pmb v^2, \pmb \beta), \label{eq.posterior1}
	\end{equation}
	quantifying the model uncertainty. 
	
	Then, we develop a Gibbs sampler to generate the posterior samples from \eqref{eq.posterior1} quantifying the model uncertainty. We derive the conditional posterior for each  parameter in $(\pmb \mu, \pmb v^2, \pmb \beta)$. Let $\pmb \mu_{-i}$, $\pmb v_{-i}^2$ and $\pmb \beta_{-ij}$ denote the collection of parameters $\pmb{\mu}, \pmb{v}^2, \pmb{\beta}$ excluding the $i$-th or ${(i,j)}$-th element. Let ${S}(X_i)$ denote the set of direct succeeding or child nodes of node $X_i$.
	We first derive the conditional posterior for the coefficient 
	$\beta_{ij}$,
	\begin{eqnarray}
		p(\beta_{ij}|\mathcal{X}, \pmb \mu, \pmb v^2, \pmb \beta_{-ij})= \mathcal{N}(\theta_{ij}^{(R)}, \tau_{ij}^{(R)2}), 
		\label{eq.post_b1}
	\end{eqnarray}
	where
	$
	\theta_{ij}^{(R)} = \dfrac{\tau_{ij}^{(0)2}\sum_{r=1}^{R} \alpha_i^{(r)} m_{ij}^{(r)} + v_j^2\theta_{ij}^{(0)}}{\tau_{ij}^{(0)2}\sum_{r=1}^{R} \alpha_i^{(r)2} + v_j^2} \quad \textrm{and} \quad
	\tau_{ij}^{(R)2} = \dfrac{\tau_{ij}^{(0)2}v_j^2} {\tau_{ij}^{(0)2}\sum_{r=1}^{R} \alpha_i^{(r)2} + v_j^2}
	\nonumber 
	$
	with 
	$
	\alpha_{i}^{(r)} = x_i^{(r)} - \mu_i$ and $
	m_{ij}^{(r)} = (x_j^{(r)} - \mu_j) - \sum_{X_k\in Pa(X_j)/\{X_i\}}\beta_{kj}(x_k^{(r)} - \mu_k).$
	Then, we derive the conditional posterior for $v_i^2=\mbox{Var}[X_i|Pa(X_i)]$ with $i=1,2,\ldots,m+1$,
	\begin{eqnarray}
		p(v_i^2|\mathcal{X}, \pmb \mu, \pmb v_{-i}^2, \pmb \beta) = \mbox{Inv-}\Gamma\left(\dfrac{\kappa_i^{(R)}}{2}, \dfrac{\lambda_i^{(R)}}{2}\right), 
		\label{eq.post_v1}
	\end{eqnarray}
	where $\kappa_i^{(R)} = \kappa_i^{(0)} + R$, 
	$\lambda_i^{(R)} = \lambda_i^{(0)} + \sum_{r=1}^{R}u_i^{(r)2}$ and 
	$u_i^{(r)} = (x_i^{(r)} - \mu_i) - \sum_{X_k\in Pa(X_i)}\beta_{ki}(x_k^{(r)} - \mu_k).
	$
	After that, we derive the conditional posterior for the mean parameter $\mu_i$ with $i=1,2,\ldots,m+1$ for any CPP/CQA,
	\begin{eqnarray}
		p(\mu_i|\mathcal{X}, \pmb \mu_{-i}, \pmb v^2, \pmb \beta) \propto p(\mu_i) \prod_{r=1}^{R}\left[p(x_i^{(r)}|x_{Pa(X_i)}^{(r)}) \prod_{j\in \mathcal{S}(X_i)} p(x_j^{(r)}|x_{Pa(X_j)}^{(r)}) \right] 
			= \mathcal{N}(\mu_i^{(R)}, \sigma_i^{(R)2}), 
			\label{eq.post_mu1}
	\end{eqnarray}
	where $
	\mu_i^{(R)} = \sigma_i^{(R)2} \left[ \dfrac{\mu_i^{(0)}}{\sigma_i^{(0)2}} + \sum_{r=1}^{R} \dfrac{a_i^{(r)}}{v_i^2} + \sum_{r=1}^{R}\sum_{X_j\in S(X_i)} \dfrac{\beta_{ij}c_{ij}^{(r)}}{v_j^2} \right] $ 
	and 
	$\dfrac{1}{\sigma_i^{(R)2}} = \dfrac{1}{\sigma_i^{(0)2}} + \dfrac{R}{v_i^2} + \sum_{X_j\in S(X_i)}\dfrac{R\beta_{ij}^2}{v_j^2}
	$
	with 
	$a_{i}^{(r)} = x_i^{(r)} - \sum_{X_k\in Pa(X_i)}\beta_{kj}(x_k^{(r)} - \mu_k)$ and $
	c_{ij}^{(r)} = \beta_{ij}x_i^{(r)} - (x_j^{(r)} - \mu_j) + \sum_{X_k\in Pa(X_j)/\{X_i\}}\beta_{kj}(x_k^{(r)} - \mu_k).$
	The Gibbs sampler iteratively draws the posterior samples of $(\pmb \mu, \pmb v^2, \pmb \beta)$ by applying the conditional posterior distributions given in \eqref{eq.post_b1}, \eqref{eq.post_v1}, and \eqref{eq.post_mu1} until convergence \citep{Gelman_2004}.

	Besides the case with complete production data, we often have additional incomplete batch data. Since the lead time for biopharmaceutical production is lengthy \citep{otto2014science}, we can have some batches in the middle of production. In addition, the bio-drug quality requirements are restricted, especially for human drugs. Following the quality control, we could discard some batches after main fermentation or even in the middle of downstream purification. 
	Thus, we provide the 
	Gibbs sampler (see Algorithm~\ref{alg:procedure2}) for both cases with complete or mixing data in Appendix \ref{subsubsec:completeIncompleteData}.
	
\end{sloppypar}







\begin{sloppypar}
	Next, we study the impact model uncertainty on the bioprocess risk and sensitivity analyses and CPPs/CQAs criticality assessment. 
	Based on Section~\ref{sec:RiskAnalysis}, the contribution from any random input factor $W_k$ 
	to the output variance $\mbox{Var}(X_{m+1})$ is measured by the Shapley value, $\mbox{Sh}_{W_k,X_{m+1}}(\pmb{\theta}^c)$.
	The unknown parameters $\pmb{\theta}^c$ specifying the underlying process probabilistic model are estimated by using limited real-world data $\mathcal{X}$. 
	Thus, the estimation uncertainty of the contribution from factor $W_k$ can be quantified by the posterior distribution, $\mbox{Sh}_{W_k,X_{m+1}}(\widetilde{\pmb{\theta}})$ with $\widetilde{\pmb{\theta}} \sim p(\pmb{\theta}|\mathcal{X})$. 
	We can use the posterior mean to estimate the expected variance contribution and criticality,
	$
	\mbox{E}^*[\mbox{Sh}_{W_k,X_{m+1}}|\mathcal{X}]
	\equiv \mbox{E}^*_{p(\pmb{\theta}|\mathcal{X})}
	[ \mbox{Sh}_{W_k,X_{m+1}}(\widetilde{\pmb{\theta}}) |\mathcal{X} ]
	$ and $
	\mbox{E}^*[p_{W_k,X_{m+1}}|\mathcal{X}]
	\equiv
	\mbox{E}^*_{p(\pmb{\theta}|\mathcal{X})}
	[  p_{W_k,X_{m+1}}(\widetilde{\pmb{\theta}})
	|\mathcal{X} ]$,
	where $p_{W_k,X_{m+1}}(\widetilde{\pmb{\theta}}) = {\mbox{Sh}_{W_k,X_{m+1}}(\widetilde{\pmb{\theta}})}/{\mbox{Var}(X_{m+1}|\widetilde{\pmb{\theta}})} $.
	The posterior variance is used to quantify the overall estimation uncertainty induced by model uncertainty, $
	\mbox{Var}^*[\mbox{Sh}_{W_k,X_{m+1}}|\mathcal{X}]
	\equiv 
	\mbox{Var}^*_{p(\pmb{\theta}|\mathcal{X})}
	[  \mbox{Sh}_{W_k,X_{m+1}}(\widetilde{\pmb{\theta}}) |\mathcal{X} ]$ and $
	\mbox{Var}^*[p_{W_k,X_{m+1}}|\mathcal{X}]
	\equiv  \mbox{Var}^*_{p(\pmb{\theta}|\mathcal{X})}
	[  p_{W_k,X_{m+1}}(\widetilde{\pmb{\theta}})
	|\mathcal{X} ]. 
	$
\end{sloppypar}

\begin{sloppypar}
	Since we do not have the closed form solutions, we can estimate the posterior mean and variance of $\mbox{Sh}(W_k)$ and $p_{W_k,X_{m+1}}$ through the sampling approach.
	By applying the Gibbs sampler in \textcolor{red}{}Appendix~\ref{sec:derGibbs}, we can generate posterior samples $\widetilde{\pmb{\theta}}^{(b)}  \sim p(\pmb{\theta}|\mathcal{X})$ with $b=1,2,\ldots,B$. 
	At any $\widetilde{\pmb{\theta}}^{(b)}$, we can compute $
	\mbox{Sh}_{W_k,X_{m+1}}(\widetilde{\pmb{\theta}}^{(b)})$ following the description in Section~\ref{sec:RiskAnalysis}.
	The expected contribution from $W_k$ to the variance of $X_{m+1}$ is estimated by $
	\widehat{\mbox{E}}^*[\mbox{Sh}_{W_k,X_{m+1}}|\mathcal{X}]
	=\bar{\mbox{Sh}}_{W_k,X_{m+1}}(\mathcal{X}) = \dfrac{1}{B}\sum_{b=1}^{B}\mbox{Sh}_{W_k,X_{m+1}}(\widetilde{\pmb{\theta}}^{(b)}). \label{eq.shapley_exp_h}
	$
	And the overall estimation uncertainty can be estimated by sample variance,
	\begin{equation}
		\widehat{\mbox{Var}}^*[\mbox{Sh}_{W_k,X_{m+1}}|\mathcal{X}] = \dfrac{1}{B-1}\sum_{b=1}^{B} \left[ \mbox{Sh}_{W_k,X_{m+1}}(\widetilde{\pmb{\theta}}^{(b)}) - \bar{\mbox{Sh}}_{W_k,X_{m+1}}(\mathcal{X}) \right]^2. \label{eq.shapley_var_h}
	\end{equation}
	Similarly, we can estimate the expected criticality by $\widehat{\mbox{E}}^*[p_{W_k,X_{m+1}}|\mathcal{X}]
	= \bar{p}_{W_k,X_{m+1}} = \dfrac{1}{B}\sum_{b=1}^{B}
	p_{W_k,X_{m+1}}(\widetilde{\pmb{\theta}}^{(b)})$ and estimate the overall estimation uncertainty by
	\begin{equation}
		\widehat{\mbox{Var}}^*[p_{W_k,X_{m+1}}|\mathcal{X}] =  \dfrac{1}{B-1}\sum_{b=1}^{B} \left[ p_{W_k,X_{m+1}}(\widetilde{\pmb{\theta}}^{(b)}) - \bar{p}_{W_k,X_{m+1}} \right]^2.
		\label{eq.critical_var_h} 
	\end{equation}
	
\end{sloppypar}


\subsection{Sensitivity Study for Model Uncertainty}
\label{subsec:SA}


\begin{sloppypar}
	
	Since there is often limited process data in biomanufacturing, model uncertainty tends to be large. We propose the BN-SV-MU based sensitivity analysis studying the effect of estimation uncertainty of each model coefficient, which can guide the process monitoring and ``most informative" data collection.
	We provide the CPPs/CQAs criticality estimation uncertainty quantification and BN-SV-MU based sensitivity analysis in Algorithm~\ref{alg:procedure3}. 
	Specifically, Steps~(1)--(3) evaluate ${\mbox{Var}}^*[\mbox{Sh}_{W_k,X_{m+1}}|\mathcal{X}]$  quantifying the overall estimation uncertainty of $\mbox{Sh}_{W_k,X_{m+1}}$. Steps~(4)--(13) further study the impact from each model coefficient estimation uncertainty.
\end{sloppypar}

Here we use $\mbox{Var}^*[\mbox{Sh}_{W_k,X_{m+1}}|\mathcal{X}]$ for illustration and the similar procedure can be applied to CPPs/CQAs criticality
assessment $\mbox{Var}^*[p_{W_k,X_{m+1}}|\mathcal{X}]$.
Let $\pmb{\theta}(W_k,X_{m+1})\subset \pmb{\theta}$ represent the subset of model coefficients that impacts on $\mbox{Sh}(W_k|\pmb{\theta})$ estimation. 
Notice that $\pmb{\mu}$ has no impact on $\mbox{Sh}_{W_k,X_{m+1}}(\pmb{\theta})$.
Since SV can account for the probabilistic dependence of model coefficient estimation uncertainty, characterized by the joint posterior distribution $p(\pmb{\theta}|\mathcal{X})$, and bioprocess structural interactions, we can measure the contribution from any parameter $\theta_\ell \in \pmb{\theta}(W_k,X_{m+1})$ through the posterior variance decomposition,
\begin{equation}
	\mbox{Var}^*[\mbox{Sh}_{W_k,X_{m+1}}|\mathcal{X}] = \sum_{ {\theta_\ell \in \pmb{\theta}(W_k,X_{m+1})}}
		\mbox{Sh}^*_{\theta_\ell}\left[
		\left. \mbox{Sh}_{W_k,X_{m+1}}\left(\widetilde{\pmb{\theta}} \right) \right| \mathcal{X} \right]=
		\sum_{\theta_\ell \in \pmb{\theta}(W_k,X_{m+1})} 
		\mbox{Sh}^*_{\theta_\ell}\left[ \left. \mbox{Sh}_{W_k,X_{m+1}} \right| \mathcal{X}\right].
	\nonumber
\end{equation}
The proposed BN-SV-MU sensitivity analysis can provide the comprehensive and interpretable understanding on how model uncertainty 
impacts on the process risk analysis and identify those parameters $\theta_\ell$ contributing the most on the estimation uncertainty of $\mbox{Sh}_{W_k,X_{m+1}}(\pmb{\theta})$. 

\begin{algorithm}[hbt!]
	\small
	\caption{Procedure for the BN-SV-MU Based UQ and SA} \label{alg:procedure3}
	\KwIn{BN structure $G(\mathbf{N}|\pmb \theta)$, 
		data $\mathcal{X}$, number of samples $N_{\pi}$, $B$, $B_O$ and $B_I$, index subset $\mathcal{L}_k$.}
	\KwOut{Return $\widehat{\widehat{\mbox{Sh}}}^*_{\theta_\ell}\left[ \left. \mbox{Sh}_{W_k,X_{m+1}} \right| \mathcal{X}\right]$ and $\widehat{\mbox{Sh}}_{\theta_{\ell}}^*\left[ \left. p_{W_k,X_{m+1}} \right| \mathcal{X}\right]$ for any $W_k \in \pmb W =\{ \mathbf{X}^p\cup \mathbf{X}^a\cup \mathbf{e}\}$.}
	
	(1) Call Algorithm~\ref{alg:procedure2} in Appendix~\ref{subsubsec:GibbsSample} to obtain the posterior samples $\widetilde{\pmb{\theta}}^{(b)} = (\widetilde{\pmb \mu}^{(b)}, \widetilde{\pmb v}^{(b)2}, \widetilde{\pmb \beta}^{(b)})$ with $b=1,2,\ldots,B$ for UQ and  $\widetilde{\pmb{\theta}}^{(b_O)} = (\widetilde{\pmb \mu}^{(b_O)}, \widetilde{\pmb v}^{(b_O)2}, \widetilde{\pmb \beta}^{(b_O)})$ with $b_O=1,2,\ldots,B_O$ for SA\;

	(2)  Call Algorithm~\ref{alg:procedure1} 
	to compute  $\mbox{Sh}_{W_k,X_{m+1}}(\widetilde{\pmb{\theta}}^{(b)})$ and criticality $p_{W_k,X_{m+1}}(\widetilde{\pmb{\theta}}^{(b)})$ for $b=1,2,\ldots,B$\;
	
	
	(3) Calculate the overall estimation uncertainty by using $\widehat{\mbox{Var}}^*[\mbox{Sh}_{W_k,X_{m+1}}|\mathcal{X}]$ and  $\widehat{\mbox{Var}}^*[p_{W_k,X_{m+1}}|\mathcal{X}]$ in Equations~\eqref{eq.shapley_var_h}  and \eqref{eq.critical_var_h}\; 
	
	(4) Randomly generate $N_{\pi}$ permutations, $\pi_n \sim  \Pi(\mathcal{L}_k)$ with $n=1,\ldots,N_{\pi}$\;
	
	\For{Each $\pi_n$}{
		(5) Set $\widehat{c}(P_{\pi_n(1)}(\pi_n)) = 0$\;
		
		\For{$\ell=1,\ldots,L_k$}{
			\uIf{$\ell < L_k$}{
				
				\For{$b_O = 1,\ldots, B_O$}{
					(7) Set initial value ${\pmb{\theta}}_{\mathcal{J}}^{(b_O,0)} = \widetilde{\pmb{\theta}}_{\mathcal{J}}^{(b_O)}$ with ${\mathcal{J}} = P_{\pi_n(\ell+1)}(\pi_n)$\;

					\For{$t=1,\ldots,T$}{
						(8) For each $\theta_{\mathcal{J}(\ell)} \in \pmb \theta_{\mathcal{J}}$,  generate $\theta_{\mathcal{J}(\ell)}^{(b_O,t)} \sim p(\theta_{\mathcal{J}(\ell)}|\mathcal{X}, \widetilde{\pmb{\theta}}_{
							\mathcal{L}_k-\mathcal{J}}^{(b_O)},  \theta_{\mathcal{J}(1)}^{(b_O,t)}, \ldots, \theta_{\mathcal{J}(\ell-1)}^{(b_O,t)}, \theta_{\mathcal{J}(\ell+1)}^{(b_O,t-1)},$ $\ldots, \theta_{\mathcal{J}(J)}^{(b_O,t-1)})$ by applying Equations~\eqref{eq.post_b1}/\eqref{eq.post_v1}/\eqref{eq.post_mu1} for the case with complete data or Equations~\eqref{eq.post_b_2}/\eqref{eq.post_v_2}/\eqref{eq.post_mu_2} for cases with mixing data (see Appendix~\ref{sec:derGibbs}). Obtain the new sample ${\pmb{\theta}}_{\mathcal{J}}^{(b_O,t)}$\;
					}
					
					(9) Set ${\widetilde{\pmb{\theta}}}_{\mathcal{J}}^{(b_O,b_I)} = {\pmb{\theta}}_{\mathcal{J}}^{(b_O, (b_I-1)h + 1)}$ with some constant integer $h$ to reduce the correlation between consecutive samples\;
				}
				
				(10) Compute $\widehat{c}(P_{\pi_n(\ell+1)}(\pi_n)) 
				$ by Equations~\eqref{eq.costfun2_h} and \eqref{eq.costfun3_h}\;
				
			}\uElse{
				(11) Set $\widehat{c}(P_{\pi_n(\ell+1)}(\pi_n)) = \widehat{\mbox{Var}}^*[\mbox{Sh}_{W_k,X_{m+1}}|\mathcal{X}]$ and $\widehat{\mbox{Var}}^*[p_{W_k,X_{m+1}}|\mathcal{X}]$\;
			}
			
			(12) Compute $\Delta_{\pi_n(\ell)} c(\pi_n) = \widehat{c}(P_{\pi_n(\ell+1)}(\pi_n)) - \widehat{c}(P_{\pi_n(\ell)}(\pi_n))$\;
		}
	}
	
	(13) Estimate  $\widehat{\widehat{\mbox{Sh}}}^*_{\theta_\ell}\left[ \left. \mbox{Sh}_{W_k,X_{m+1}} \right| \mathcal{X}\right]$ and $\widehat{\mbox{Sh}}_{\theta_{\ell}}^*\left[ \left. p_{W_k,X_{m+1}} \right| \mathcal{X}\right]$ by using Equations~\eqref{eq.sensitivity_h2} and \eqref{eq.sensitivity_h3}.

\end{algorithm}

\begin{sloppypar}
	Then, we derive SV measuring the estimation uncertainty contribution from each $\theta_\ell$,
	\[
	\mbox{Sh}^*_{\theta_\ell}\left[ \left. \mbox{Sh}_{W_k,X_{m+1}} \right| \mathcal{X}\right] = \sum_{\mathcal{J}\subset \mathcal{L}_k/\{\ell\}}\dfrac{(L_k -|\mathcal{J}|-1)! |\mathcal{J}|!} {L_k!} \\ \left[ c(\mathcal{J} \cup\{\ell\}) - c(\mathcal{J}) \right].
	\] 
	Denote the size of relevant parameters by $L_k=|\pmb \theta(W_k,X_{m+1})|$ and denote the index set by $\mathcal{L}_k$, $\pmb{\theta}(W_k,X_{m+1}) = \pmb{\theta}_{\mathcal{L}_k}$.
	We further denote any subset by $\pmb{\theta}_\mathcal{J}\subset \pmb{\theta}(W_k,X_{m+1})$ with size $J=|\pmb \theta_\mathcal{J}|$ and the corresponding index set $\mathcal{J} = \{\mathcal{J}(1), \mathcal{J}(2), \ldots, \mathcal{J}(J)\} \subset \mathcal{L}_k$.
	For any $\mathcal{J} \subset \mathcal{L}_k$, the cost function is given as,
	\begin{equation}
		c(\mathcal{J}) = \mbox{E}^*_{p(\pmb{\theta}_{\mathcal{L}_k-\mathcal{J}}|\mathcal{X})}[\mbox{Var}^*_{p(\pmb{\theta}_{\mathcal{J}}|{\pmb \theta}_{\mathcal{L}_k-\mathcal{J}}, \mathcal{X})}[\mbox{Sh}_{W_k,X_{m+1}}|\widetilde{\pmb \theta}_{\mathcal{L}_k-\mathcal{J}}]], 
		\label{eq.costfun2}
	\end{equation}
	where ${\pmb \theta}_{\mathcal{L}_k-\mathcal{J}} = \pmb{\theta}_{\mathcal{L}_k\setminus \mathcal{J}}$. Denote a permutation of $\mathcal{L}_k$ as $\pi$ and define the set $P_\ell(\pi)$ as the index set preceding $\ell$ in $\pi$. The SV can be rewritten as,
	\begin{equation}
		\mbox{Sh}^*_{\theta_\ell}\left[ \left. \mbox{Sh}_{W_k,X_{m+1}} \right| \mathcal{X}\right] = \sum_{\pi \in \Pi(\mathcal{L}_k)} \dfrac{1}{L_k!} \left[ c(P_\ell(\pi)\cup\{\ell\}) - c(P_\ell(\pi)) \right], \label{eq.sensitivity2}
	\end{equation}
	where $\Pi(\mathcal{L}_k)$ denotes the set of all $L_k!$ permutations of $\mathcal{L}_k$. 
\end{sloppypar}

\vspace{-0.03in}

The number of all possible subsets $\mathcal{J}$ could grow exponentially as $L_k$ increase. 
To address this computational issue, we use the Monte Carlo sampling approximation, ApproShapley, suggested by \cite{song2016shapley,castro2009polynomial}, which estimates the Shapley value in \eqref{eq.sensitivity2} by
\begin{equation}
	\widehat{\mbox{Sh}}^*_{\theta_\ell}\left[ \left. \mbox{Sh}_{W_k,X_{m+1}} \right| \mathcal{X}\right] = \dfrac{1}{N_{\pi}} \sum_{n=1}^{N_{\pi}} \left[ c(P_\ell(\pi_n)\cup\{\ell\}) - c(P_\ell(\pi_n)) \right] \triangleq \dfrac{1}{N_{\pi}} \sum_{n=1}^{N_{\pi}} \Delta_\ell c(\pi_n), \label{eq.sensitivity_h}
\end{equation}
where $N_{\pi}$ denotes the number of permutations $\pi_1, \ldots, \pi_{N_{\pi}}$ randomly generated from $\Pi(\mathcal{L}_k)$ and
$\Delta_\ell c(\pi_n) = c(P_\ell(\pi_n)\cup\{\ell\}) - c(P_\ell(\pi_n))$ is the incremental posterior variance
$\mbox{Var}^*[\mbox{Sh}_{W_k,X_{m+1}}|\mathcal{X}]$ induced by including the $\ell$-th model parameter in $P_\ell(\pi_n)$.

\begin{sloppypar}
	As $c(\mathcal{J})$ in \eqref{eq.costfun2} is analytically intractable, we develop Monte Carlo sampling estimation. 
	However, since the posterior samples
	obtained from the Gibbs sampler in Appendix \ref{subsubsec:GibbsSample}
	cannot be directly used to estimate  $\mbox{E}^*_{p(\pmb{\theta}_{\mathcal{L}_k-\mathcal{J}}|\mathcal{X})}[\mbox{Var}^*_{p(\pmb{\theta}_{\mathcal{J}}|{\pmb \theta}_{\mathcal{L}_k-\mathcal{J}}, \mathcal{X})}[\mbox{Sh}_{W_k,X_{m+1}}|\widetilde{\pmb \theta}_{\mathcal{L}_k-\mathcal{J}}]]$, we introduce a \textit{nested Gibbs sampling approach}. 
	For the ``outer" samples used to estimate $\mbox{E}^*_{p(\pmb{\theta}_{\mathcal{L}_k-\mathcal{J}}|\mathcal{X})}[\cdot]$,
	the posterior samples $\widetilde{\pmb{\theta}}_{\mathcal{L}_k-\mathcal{J}}^{(b_O)}$ with $b_O = 1, \ldots, B_O$
	can be directly obtained by applying the Gibbs sampling in Appendix \ref{subsubsec:GibbsSample}. We generate $\widetilde{\pmb{\theta}}^{(b_O)} \sim p(\pmb{\theta}|\mathcal{X})$ and keep components with index $\mathcal{L}_k-\mathcal{J}$.
	Then, at each $\widetilde{\pmb{\theta}}_{\mathcal{L}_k-\mathcal{J}}^{(b_O)}$, a conditional sampling is further developed to 
	generate samples from 
	$p(\pmb{\theta}_{\mathcal{J}}|{\widetilde{\pmb \theta}}_{\mathcal{L}_k-\mathcal{J}}^{(b_O)}, \mathcal{X})$.  
	More specifically, we set the initial value ${\pmb{\theta}}_{\mathcal{J}}^{(b_O,0)} = \widetilde{\pmb{\theta}}_{\mathcal{J}}^{(b_O)}$. In each $t$-th MCMC iteration, given the previous sample ${\pmb{\theta}}_{\mathcal{J}}^{(b_O,t-1)}$, we apply the Gibbs sampling to sequentially generate one sample from the conditional posterior for each $\theta_{\mathcal{J}(\ell)} \in \pmb{\theta}_{\mathcal{J}}$ with $\ell=1,\ldots,|\mathcal{J}|$, 
	\begin{equation}
		\theta_{\mathcal{J}(\ell)}^{(b_O,t)} \sim p\left(\theta_{\mathcal{J}(\ell)}
		\left|\mathcal{X}, 
		\widetilde{\pmb{\theta}}_{\mathcal{L}_k-\mathcal{J}}^{(b_O)},
		\theta_{\mathcal{J}(1)}^{(b_O,t)}, \ldots,
		\theta_{\mathcal{J}(\ell-1)}^{(b_O,t)},
		\theta_{\mathcal{J}(\ell+1)}^{(b_O,t-1)}, \ldots,
		\theta_{\mathcal{J}(|\mathcal{J}|)}^{(b_O,t-1)}  \right. \right).
		\nonumber 
	\end{equation}
	By repeating this procedure, we can get samples ${\pmb{\theta}}_{\mathcal{J}}^{(b_O,t)}$ with $t=0, \ldots, T$. We keep one for every $h$ samples to reduce the correlations between consecutive samples. Consequently, we obtain ``inner" samples $\widetilde{\pmb{\theta}}_{\mathcal{J}}^{(b_O,b_I)}$ with $b_I = 1, \ldots, B_I$ to estimate
	$\mbox{Var}^*_{p(\pmb{\theta}_{\mathcal{J}}|{\pmb \theta}_{\mathcal{L}_k-\mathcal{J}}, \mathcal{X})}[\mbox{Sh}_{W_k,X_{m+1}}|\widetilde{\pmb \theta}_{\mathcal{L}_k-\mathcal{J}}]$.

	Thus, this nested Gibbs sampling can generate  $B_O\cdot B_I$ samples $\{(\widetilde{\pmb{\theta}}_{\mathcal{J}}^{(b_O,b_I)}, \widetilde{\pmb{\theta}}_{\mathcal{L}_k-\mathcal{J}}^{(b_O)}):b_O = 1,\ldots,B_O \mbox{ and }b_I = 1, \ldots, B_I\}$ to estimate $c(\mathcal{J})$ in \eqref{eq.costfun2}.
	For any $\mathcal{J} \subset \mathcal{L}_k$, the cost function can be estimated as,
	\begin{equation}
		\widehat{c}(\mathcal{J}) = \dfrac{1}{B_O}\sum_{b_O=1}^{B_O}\left\{ \dfrac{1}{B_I-1}\sum_{b_I=1}^{B_I} \left[ \mbox{Sh}_{W_k,X_{m+1}}(
		\widetilde{\pmb \theta}_{\mathcal{J}}^{(b_O, b_I)}, \widetilde{\pmb \theta}_{\mathcal{L}_k-\mathcal{J}}^{(b_O)}) - \bar{\mbox{Sh}}_{W_k,X_{m+1}}( \widetilde{\pmb \theta}_{\mathcal{L}_k-\mathcal{J}}^{(b_O)}) \right]^2 \right\},
		\label{eq.costfun2_h}  
	\end{equation}
	where $\bar{\mbox{Sh}}_{W_k,X_{m+1}}( \widetilde{\pmb \theta}_{\mathcal{L}_k-\mathcal{J}}^{(b_O)}) = 
	\sum_{b_I=1}^{B_I}
	{\mbox{Sh}}_{W_k,X_{m+1}}(
	\widetilde{\pmb \theta}_{\mathcal{J}}^{(b_O, b_I)}, \widetilde{\pmb \theta}_{\mathcal{L}_k-\mathcal{J}}^{(b_O)})/B_I$. By plugging $\widehat{c}(\mathcal{J})$ into Equation~\eqref{eq.sensitivity_h}, we can quantify the estimation uncertainty contribution from each model coefficient $\theta_\ell \in \pmb \theta_{\mathcal{L}_k}$, 
	\begin{equation}
		\widehat{\widehat{\mbox{Sh}}}^*_{\theta_\ell}\left[ \left. \mbox{Sh}_{W_k,X_{m+1}} \right| \mathcal{X}\right] = \dfrac{1}{N_{\pi}} \sum_{n=1}^{N_{\pi}} \Delta_\ell \widehat{c}(\pi_n),  \label{eq.sensitivity_h2}
	\end{equation}
	where $\Delta_\ell \widehat{c}(\pi_n) =  \widehat{c}(P_\ell(\pi_n)\cup\{\ell\}) - \widehat{c}(P_\ell(\pi_n))$ 
	for all $\ell=1,\ldots,L_k$. 
	Similarly, for CPP/CQA criticality assessment, we can estimate the cost function,
	\begin{equation}
		\widehat{c}^\prime(\mathcal{J}) = \dfrac{1}{B_O}\sum_{b_O=1}^{B_O}\left\{ \dfrac{1}{B_I-1}\sum_{b_I=1}^{B_I} \left[ p_{W_k,X_{m+1}}(
		\widetilde{\pmb \theta}_{\mathcal{J}}^{(b_O, b_I)}, \widetilde{\pmb \theta}_{\mathcal{L}_k-\mathcal{J}}^{(b_O)}) - \bar{p}_{W_k,X_{m+1}} (\widetilde{\pmb \theta}_{\mathcal{L}_k-\mathcal{J}}^{(b_O)}) \right]^2 \right\},
		\label{eq.costfun3_h}  
	\end{equation}
	where $\bar{p}_{W_k,X_{m+1}}( \widetilde{\pmb \theta}_{\mathcal{L}_k-\mathcal{J}}^{(b_O)}) = 
	\sum_{b_I=1}^{B_I}
	p_{W_k,X_{m+1}}(
	\widetilde{\pmb \theta}_{\mathcal{J}}^{(b_O, b_I)}, \widetilde{\pmb \theta}_{\mathcal{L}_k-\mathcal{J}}^{(b_O)})/B_I$. Then, we estimate the estimation uncertainty contribution from $\theta_\ell$ on the criticality assessement,
	\begin{equation}
		\widehat{\mbox{Sh}}_{\theta_{\ell}}^*\left[ \left. p_{W_k,X_{m+1}} \right| \mathcal{X}\right] = \dfrac{1}{N_{\pi}} \sum_{n=1}^{N_{\pi}} \Delta_\ell \widehat{c}^\prime(\pi_n).  \label{eq.sensitivity_h3}
	\end{equation}

	More real-world data can reduce the impact of process model coefficient estimation uncertainty and improve the criticality estimation accuracy of input factors, say $W_k$, that contribute the most to the variance of output $X_{m+1}$. This study can guide the ``most informative" data collection. Basically, we can focus on the dominant criticality measurements with high estimation uncertainty induced by model uncertainty, assessed by variance $\widehat{\mbox{Var}}^*[\mbox{Sh}_{W_k,X_{m+1}}|\mathcal{X}]$ in Equation~\eqref{eq.shapley_var_h}, or $\widehat{\mbox{Var}}^*[p_{W_k,X_{m+1}}|\mathcal{X}]$ in Equation~\eqref{eq.critical_var_h}. Given the real-world data $\mathcal{X}$, the proportion of estimation uncertainty contributed from each coefficient $\theta_{\ell} \in \pmb{\theta}(W_k,X_{m+1})$ can be estimated by,
	\begin{equation}
		\widehat{p}^*_{\theta_\ell}\left[ \left. \mbox{Sh}_{W_k,X_{m+1}} \right| \mathcal{X}\right] = \dfrac{\widehat{\widehat{\mbox{Sh}}}^*_{\theta_\ell}\left[ \left. \mbox{Sh}_{W_k,X_{m+1}} \right| \mathcal{X}\right]}{\widehat{\mbox{Var}}^*[\mbox{Sh}_{W_k,X_{m+1}}|\mathcal{X}]}, 
		~\mbox{or}~~
		\widehat{p}_{\theta_{\ell}}^*\left[ \left. p_{W_k,X_{m+1}} \right| \mathcal{X}\right] = \dfrac{\widehat{\mbox{Sh}}_{\theta_{\ell}}^*\left[ \left. p_{W_k,X_{m+1}} \right| \mathcal{X}\right]}{\widehat{\mbox{Var}}^*[p_{W_k,X_{m+1}}|\mathcal{X}]}.
		\nonumber 
	\end{equation}
By ranking the proportional contribution, we can find the coefficient $\theta_\ell$ with the highest contribution, which can guide the collection of the most informative data to control the impact of model estimation uncertainty and support production process risk analysis. We will study the impact of additional data collection and provide a systematic and rigorous approach to guide efficient data collection in the future research.

\end{sloppypar}

\section{Empirical Study}
\label{sec:empiricalStudy}

To assess the performance of proposed risk and sensitivity analyses, we first consider an integrated biomanufacturing process with simulated data in Section~\ref{subsec:simulationEmpiricalStudy}. Then, we study the performance by utilizing the real-world process data collected from a cell culture process with multiple stages in Section~\ref{subsec:caseStudy}.
Even though there are many factors impacting on the biomanufacturing outputs, the amount of real-world bioprocess observations is often very limited. Therefore, it is important to explore the causal relationships of {the} biopharmaceutical production process, which can reduce the model uncertainty, increase the interpretability for process sensitivity analysis, and guide the decision making to improve the production stability.

\subsection{Study the Performance of Proposed Framework with Simulation Data}
\label{subsec:simulationEmpiricalStudy}



We revisit the example described at the end of Section~\ref{subsec:summaryProposedFramework} in Fig.~\ref{fig:MR_example}. 
In total, the process graph model includes 20 nodes, consisting of 10 CPPs ($\mathbf{X}^p$) and 8 CQAs ($\mathbf{X}^a$) for intermediate product and 2 CQAs ($\mathbf{Y}$) for {the} final drug substance. 
The size of coefficients $\pmb \theta$ is 84, including 20 $\mu_i$'s, 20 $v_i$'s, and 44 $\beta_{ij}$'s coefficients.
To study the performance of {the} proposed framework, we generate the simulated production process data $\mathcal{X}$, which mimics the real-world data collection. The BN-based probabilistic knowledge graph with parameters $\pmb{\theta}^c= (\pmb{\mu}^c, (\pmb{v}^2)^c, \pmb{\beta}^c)$, characterizing the underlying bioprocess risk behaviors and CPPs/CQAs interdependencies, is used for data generation, which is built on the biomanufacturing domain knowledge; see the detailed setting in Appendix \ref{sec:inputDataSimulation}. 
To assess the performance of {the} proposed framework, we assume that the true parameter values are unknown.
We empirically study the convergence of process model parameter inference in Appendix \ref{subsec:emp_convergence}. In Sections~\ref{subsec:empiricalCriticality} and \ref{subsec:emp_sensitivity}, we show the capabilities of the proposed process risk and sensitivity analyses by studying both process inherent stochastic uncertainty and model uncertainty.

\subsubsection{Bioprocess Sensitivity Analysis and CPPs/CQAs Criticality Assessment}
\label{subsec:empiricalCriticality}

\begin{sloppypar}
	We generate the data $\mathcal{X}$ with the number of batch $R=30$ to study the performance of {the} proposed risk and sensitivity analyses. For any intermediate and final product CQA output $X_i$ of interest, at each posterior sample $\widetilde{\pmb \theta}$, we follow Algorithm~\ref{alg:procedure1} to assess the criticality of any input factor $W_k$ (i.e., CPPs/CQAs, residual factors).
	Specifically, in the $h$-th macro-replication of simulation, we first generate the ``real-world" batch data $\mathcal{X}^{(h)}$ with $h=1,2,\ldots,H$, which is used to mimic the process data collection.
	Considering the criticality of input $W_k$ to the output variance $p_{W_k,X_i}(\widetilde{\pmb{\theta}})= {\mbox{Sh}_{W_k,X_{i}}(\widetilde{\pmb{\theta}})}/{\mbox{Var}(X_{i}|\widetilde{\pmb{\theta}})}$, 
	we estimate the expected value $\mbox{E}[p_{W_k,X_i}] = \iint p_{W_k,X_i}(\pmb{\theta})dP(\pmb{\theta}|\mathcal{X})dP(\mathcal{X}|\pmb{\theta}^c) \times 100\%$ by using $\widehat{\mbox{E}}[p_{W_k,X_i}] = \frac{1}{HB} \sum_{h=1}^H\sum_{b=1}^B 
	p_{W_k,X_i}(\widetilde{\pmb{\theta}}^{(h,b)}) \times 100\%$ with $\widetilde{\pmb{\theta}}^{(h,b)} \sim p(\pmb{\theta}|\mathcal{X}^{(h)})$ for $h=1,\ldots,H$ and $b=1,\ldots,B$, with $H=20$ and $B=1000$, and then record the results in terms of percentage (\%)
	in Tables~\ref{table:cpp2cqa_mean} and \ref{table:cqa2cqa_mean}. Each row records the criticality for each CPP, CQA, or residual input factor $W_k$, and each column corresponds to an intermediate or final product CQA output $X_i$.
	
	\vspace{-0.0in}
	
	\begin{table}[hbt!]
		\centering
		\caption{The estimated criticality level $\widehat{\mbox{E}}[p_{W_k,X_i}]$ and standard deviation $\widehat{\mbox{SD}}[p_{W_k,X_i}] $ (in \%) of any input CPP or other factor $W_k$ impacting on the variance of intermediate or final product CQA $X_i$.}
		\label{table:cpp2cqa_mean}
		\resizebox{\textwidth}{!}{%
			\begin{tabular}{|c|ccc|cc|ccc|cc|}
				\hline
				$\widehat{\mbox{E}}[p_{W_k,X_i}]$ & $X_i=X_5$           & $X_6$           & $X_7$           & $X_{10}$        & $X_{11}$       & $X_{14}$        & $X_{15}$        & $X_{16}$       & $X_{19}$        & $X_{20}$                \\ \hline
				$W_k = X_1$                       & 8.91(3.09)  & 8.93(3.11)  & 9.42(3.59)  & 8.51(2.87)  & 8.51(2.87) & 5.87(1.88)  & 5.87(1.88)  & 5.87(1.88) & 5.52(1.74)  & 5.52(1.74)    \\
				$X_2$                             & 0.82(0.38)  & 0.76(0.35)  & 0.96(0.75)  & 0.75(0.32)  & 0.75(0.32) & 0.52(0.21)  & 0.52(0.21)  & 0.52(0.2)  & 0.49(0.19)  & 0.49(0.19)    \\
				$X_3$                             & 4.28(1.6)   & 4.33(1.61)  & 4.22(1.97)  & 4.05(1.46)  & 4(1.44)    & 2.75(0.93)  & 2.75(0.93)  & 2.75(0.93) & 2.59(0.86)  & 2.59(0.86)    \\
				$X_4$                             & 85.75(4.02) & 85.73(4.03) & 83.29(4.84) & 81.52(4.55) & 81.6(4.53) & 58.2(7.32)  & 58.22(7.32) & 58.2(7.32) & 55.09(7.21) & \textbf{55.09(7.21)}  \\
				$e_5$                             & 0.23(0.1)   &                 &                 & 0.05(0.04)  & 0.04(0.04) & 0.03(0.02)  & 0.03(0.02)  & 0.03(0.02) & 0.03(0.02)  & 0.03(0.02)    \\
				$e_6$                             &                 & 0.24(0.1)   &                 & 0.04(0.03)  & 0.05(0.04) & 0.03(0.02)  & 0.03(0.02)  & 0.03(0.02) & 0.03(0.02)  & 0.03(0.02)    \\
				$e_7$                             &                 &                 & 2.11(0.86)  & 0.05(0.04)  & 0.06(0.04) & 0.03(0.02)  & 0.03(0.02)  & 0.03(0.02) & 0.03(0.02)  & 0.03(0.02)    \\ \hline
				$X_8$                             &                 &                 &                 & 1.02(0.41)  & 1.01(0.41) & 0.68(0.24)  & 0.68(0.25)  & 0.69(0.25) & 0.64(0.23)  & 0.64(0.23)    \\
				$X_9$                             &                 &                 &                 & 3.91(1.42)  & 3.86(1.41) & 2.66(0.9)   & 2.68(0.91)  & 2.67(0.9)  & 2.51(0.84)  & 2.51(0.84)    \\
				$e_{10}$                          &                 &                 &                 & 0.1(0.04)   &                & 0.02(0.01)  & 0.02(0.01)  & 0.02(0.01) & 0.02(0.01)  & 0.02(0.01)    \\
				$e_{11}$                          &                 &                 &                 &                 & 0.12(0.05) & 0.02(0.01)  & 0.02(0.02)  & 0.02(0.01) & 0.02(0.01)  & 0.02(0.01)   \\ \hline
				$X_{12}$                          &                 &                 &                 &                 &                & 1.86(0.63)  & 1.9(0.66)   & 1.86(0.63) & 1.76(0.59)  & 1.76(0.59)    \\
				$X_{13}$                          &                 &                 &                 &                 &                & 27.31(6.46) & 27.18(6.45) & 27.3(6.46) & 25.72(6.09) & \textbf{25.73(6.09)}  \\
				$e_{14}$                          &                 &                 &                 &                 &                & 0.02(0.01)  &                 &                & $<$0.01($<$0.01)        & $<$0.01($<$0.01)               \\
				$e_{15}$                          &                 &                 &                 &                 &                &                 & 0.06(0.02)  &                & $<$0.01($<$0.01)        & $<$0.01($<$0.01)                \\
				$e_{16}$                          &                 &                 &                 &                 &                &                 &                 & 0.02(0.01) & $<$0.01($<$0.01)        & $<$0.01($<$0.01)               \\ \hline
				$X_{17}$                          &                 &                 &                 &                 &                &                 &                 &                & 1.27(0.45)  & 1.27(0.43)    \\
				$X_{18}$                          &                 &                 &                 &                 &                &                 &                 &                & 4.23(1.39)  & 4.26(1.39)     \\
				$e_{19}$                          &                 &                 &                 &                 &                &                 &                 &                & 0.04(0.01)  &                                \\
				$e_{20}$                          &                 &                 &                 &                 &                &                 &                 &                &                 & 0.01($<$0.01)                     \\
				\hline
			\end{tabular}%
		}
	\end{table}

	\begin{table}[hbt!]
		\centering
		\caption{The estimated criticality level $\widehat{\mbox{E}}[p_{W_k,X_i}] $ and standard deviation $\widehat{\mbox{SD}}[p_{W_k,X_i}] $ (in \%) of any input CQA $W_k$ on the variance of intermediate or final product CQA $X_i$.}
		\label{table:cqa2cqa_mean}
		\resizebox{\textwidth}{!}{%
			\begin{tabular}{|c|cc|ccc|cc|}
				\hline
				$\widehat{\mbox{E}}[p_{W_k,X_i}]$ & $X_i=X_{10}$         & $X_{11}$         & $X_{14}$        & $X_{15}$        & $X_{16}$        & $X_{19}$         & $X_{20}$              \\ \hline
				$W_k = X_5$                       & 43.1(11.18)  & 38.05(11.44) & 28.68(6.48) & 28.46(6.68) & 28.44(6.48) & 27.05(6.14)  & 26.99(6.11)  \\
				$X_6$                             & 37.97(10.79) & 42.44(11.27) & 28.64(6.32) & 28.8(6.5)   & 28.88(6.28) & 27.13(5.98)  & 27.18(5.95) \\
				$X_7$                             & 13.91(4.42)  & 14.52(4.92)  & 10.11(2.62) & 10.2(2.73)  & 10.11(2.67) & 9.59(2.49)   & 9.6(2.49)     \\ \hline
				$X_{10}$                          &                  &                  & 37.17(6.55) & 33.62(9.62) & 34.59(6.7)  & 33.52(5.69)  & 33.01(5.16)  \\
				$X_{11}$                          &                  &                  & 33.64(6.42) & 37.24(9.74) & 36.23(6.72) & 33.44(5.69)  & 33.95(5.2)   \\ \hline
				$X_{14}$                          &                  &                  &                 &                 &                 & 32.65(12.69) & 31.91(7.8)   \\
				$X_{15}$                          &                  &                  &                 &                 &                 & 21.49(12.22) & 25.1(6.9)    \\
				$X_{16}$                          &                  &                  &                 &                 &                 & 40.31(14.51) & 37.45(7.64)  \\ \hline
			\end{tabular}%
		}
	\end{table}

	
	\textit{The process model uncertainty is characterized by the posterior $p(\pmb{\theta}|\mathcal{X})$ and the overall impact on the CPPs/CQAs criticality assessment can be quantified by the posterior standard deviation (SD), $\mbox{SD}^*[p_{W_k,X_i}(\widetilde{\pmb{\theta}})|\mathcal{X}]$.}
	Based on the results from $H$ macro-replications, we compute the expected SD for criticality estimation, $\mbox{SD}[p_{W_k,X_i}]= \sqrt{\mbox{E}[\mbox{Var}^*(p_{W_k,X_i}(\widetilde{\pmb{\theta}})|\mathcal{X})]}\times 100\%$,  with the estimate,
	\begin{equation}
		\widehat{\mbox{SD}}[p_{W_k,X_i}]  = \sqrt{
			\dfrac{1}{H(B-1)}\sum_{h=1}^H\sum_{b=1}^{B} \left[ p_{W_k,X_i}\left(\widetilde{\pmb{\theta}}^{(h,b)}\right) - 
			\bar{p}_{W_k,X_i}^{(h)} \right]^2} \times 100\%
		\nonumber 
	\end{equation}
	where $\bar{p}_{W_k,X_i}^{(h)}=\frac{1}{B}\sum_{b=1}^B p_{W_k,X_i}(\widetilde{\pmb{\theta}}^{(h,b)})$. In Tables~\ref{table:cpp2cqa_mean} and \ref{table:cqa2cqa_mean}, 
	we record the results of SD in terms of percentage (\%) in the bracket.

	For each CQA output $X_i$, we record the criticality with the estimated mean $\widehat{\mbox{E}}[p_{W_k,X_i}]$ and standard deviation $\widehat{\mbox{SD}}[p_{W_k,X_i}]$ from any CPP or other factor $W_k$ 
	in Table~\ref{table:cpp2cqa_mean}. 
	Under the example setting, we can see that the variations in $X_4$ (dissolved oxygen in main fermentation) and $X_{13}$ (temperature in chromatography) have the dominant impact on both intermediate and final product CQAs' variance. Compared with main fermentation and chromatography, the other two operation units (i.e., centrifuge and filtration) have relatively small impact on the final product quality variation. 
	Based on the process risk and sensitivity analyses, we also provide the result visualization; see for example Fig.~\ref{fig:MR_example}. 
	
	By studying the \textit{subplots} of the bioprocess probabilistic knowledge graph illustrated in Fig.~\ref{fig:MR_example}, we can study the contributions from the dependent CQAs of intermediate products as inputs
	to the variance of final drug substance CQAs outputs, i.e., nodes $\{X_{19}, X_{20}\}$. 
	We consider the subplots: (1) starting from the end of main fermentation with $\{X_5, X_6,X_7\}$; (2) starting from the end of centrifuge with  $\{X_{10},X_{11}\}$; and (3) starting from the end of chromatography with $\{X_{14},X_{15},X_{16}\}$. The results of process sensitivity analysis are recorded in Table~\ref{table:cqa2cqa_mean}. 
	The CQAs after main fermentation, i.e., $\{X_5, X_6, X_7\}$, together account for about 50\% variance of final output $X_{19}$ or $X_{20}$; and CQAs after chromatography, i.e., $\{X_{14}, X_{15}, X_{16}\}$ together account for about 90\% of final output variation. 
	\textit{Thus, the CQAs of intermediate product close to the end of production process provides better explanation of the variation of final drug substance CQAs and we can predict more accurate on its productivity and quality.} This information can be used to guide the production process quality control and support the real-time release.
\end{sloppypar}


\subsubsection{Criticality Assessment Estimation Performance Comparison}
\label{subsubsec:comparison}

In this section, we use the same example studied in Section~\ref{subsec:empiricalCriticality} to 
compare the performance of criticality assessment obtained by the proposed BN-SV approach (denoted by $p_{W_k,X_{20}}^{BN-SV}$) with an existing approach, which uses multiple linear regression and Morris sensitivity analysis (represented by ML-M); see \cite{Hassan_2013,Zi_2011,helton1993uncertainty}. Basically, we first fit the multiple linear regression to the random inputs (i.e., $W_k=X_i$ listed in the first column of Table~\ref{table:cpp2cqa_mean}) and output $X_{20}$, and then use Morris sensitivity analysis to measure the criticality of each input $W_k$.  
Here, we use the same experiment setting with that used in Section~\ref{subsec:empiricalCriticality}. 
With the underlying parameters setting $\pmb{\theta}^c= (\pmb{\mu}^c, (\pmb{v}^2)^c, \pmb{\beta}^c)$ given in Appendix~\ref{sec:inputDataSimulation}, the true criticality of any input factor $W_k$ can be calculated with $p_{W_k,X_{20}}^c = {\mbox{Sh}_{W_k,X_{20}}(\pmb{\theta}^c)}/ {\mbox{Var}(X_{20}|\pmb{\theta}^c)}$, where ${\mbox{Sh}_{W_k,X_{20}}(\pmb{\theta}^c)}$ and ${\mbox{Var}(X_{20}|\pmb{\theta}^c)}$ are obtained by applying Equations \eqref{eq.shapley_3} and \eqref{eq.VarDecompCPP}. Then, suppose the underlying process model coefficients are unknown, and we can compare the criticality assessment performance of both approaches.
In Table~\ref{table:comparison}, we record the mean and SD of criticality estimates obtained from LM-M and proposed BN-SV approaches with $H=30$ macro-replications and $R=30$ batches. The mean absolute error (MAE) is calculated by, 
\begin{equation}
    MAE(p_{W_k,X_{20}}^\gamma) = 
    \frac{1}{HB} \sum_{h=1}^H\sum_{b=1}^B 
	\left|p_{W_k,X_{20}}^\gamma(\widetilde{\pmb{\theta}}^{(h,b)}) - p_{W_k,X_{20}}^c\right|
	\times 100\%
\end{equation}
where $\gamma$ is ML-M or BN-SV. The results in Table~\ref{table:comparison} show that the proposed BN-SV sensitivity analysis provides better criticality assessment of critical inputs.
\begin{table}[hbt!]
	\small
	\centering
	\caption{The CPPs criticality estimation results obtained by BN-SV sensitivity analysis and existing multiple regression based sensitivity analysis. 
	}
	\label{table:comparison}
		\begin{tabular}{|c|c|cc|cc|}
			\hline
			  Criticality (\%)     & True Value $p_{W_k,X_{20}}^c$ &  $p_{W_k,X_{20}}^{ML-M}$ & MAE & $p_{W_k,X_{20}}^{BN-SV}$ & MAE    \\ \hline
$W_k=X_4$ & 59.55 & 56.91 (14.94) & 11.14 & 55.09 (7.21) & 7.86 \\ \hline 
$X_{13}$ & 24.01 & 26.06 (9.47) & 6.41 & 25.73 (6.09) & 5.19
\\ \hline 
$X_1$ & 4.67 & 5.40 (1.63) & 1.24 & 5.52 (1.74) & 1.41
 \\ \hline 
$X_{18}$ & 3.66 & 4.25 (1.64) & 1.21 & 4.26 (1.39) & 1.13
 \\ \hline  
 $X_3$ & 2.38 & 2.55 (0.77) & 0.61 & 2.59 (0.86) & 0.47
  \\ \hline  
 $X_9$ & 2.16 & 2.36 (0.77) & 0.64 & 2.51 (0.84) & 0.68
     \\ \hline 
$X_{12}$ & 1.5 & 1.73 (0.52) & 0.45 & 1.76 (0.59) & 0.42
    \\ \hline 
$X_{17}$ & 1.04 & 1.25 (0.59) & 0.43 & 1.27 (0.43) & 0.35
    \\ \hline  
			\end{tabular}%
\end{table}

\subsubsection{Sensitivity Analysis for Model Uncertainty}
\label{subsec:emp_sensitivity}

Here we consider the product protein content $X_{20}$ in Fig.~\ref{fig:MR_example} as the output to study the performance of sensitivity analysis for model uncertainty. Based on the results in Table~\ref{table:cpp2cqa_mean}, the CPPs $X_4$ and $X_{13}$ have the dominant contributions to the variance of output $X_{20}$, and the estimates of $p_{W_k,X_i}$ also have the high estimation uncertainty. Thus, we conduct the BN-SV-MU sensitivity analysis to study how the estimation uncertainty of each model coefficient impacts on the criticality assessment for $p_{X_{4},X_{20}}$ and $p_{X_{13},X_{20}}$.

\begin{sloppypar}
	Given the data $\mathcal{X}$, we provide the posterior variance decomposition studying the criticality estimation uncertainty induced by the MU,
	$
	\mbox{Var}^*_{p(\pmb{\theta}|\mathcal{X})}[p_{W_k,X_i}(\widetilde{\pmb{\theta}})|\mathcal{X}]
	= \sum_{\theta_\ell \in \pmb{\theta}(W_k,X_i)} \mbox{Sh}_{\theta_{\ell}}^*\left[ \left. p_{W_k,X_i}(\widetilde{\pmb{\theta}}) \right| \mathcal{X}\right].$
	Then, we can estimate the expected relative contribution from each model coefficient $\theta_\ell \in \pmb{\theta}(W_k,X_i)$ with
	$
	\mbox{EP}_{\theta_\ell}(p_{W_k,X_i}) \equiv \mbox{E}\left[
	\frac{ {{\mbox{Sh}}}^*_{\theta_{\ell}}\left[ \left. p_{W_k,X_i}(\widetilde{\pmb{\theta}}) \right| \mathcal{X}\right]} {
		{\mbox{Var}^*_{p(\pmb{\theta}|\mathcal{X})}\left[ \left. p_{W_k,X_i}(\widetilde{\pmb{\theta}})
			\right|\mathcal{X}
			\right]}}\right].  $
	In the $h$-th macro-replication, given the data $\mathcal{X}^{(h)}$, we can estimate the contribution from each $\theta_\ell $ by using $ \widehat{\widehat{\mbox{Sh}}}^*_{\theta_{\ell}}\left[ \left. p_{W_k,X_i}(\widetilde{\pmb{\theta}}) \right| \mathcal{X}^{(h)}\right]$ and $\widehat{\mbox{Var}}^*_{p(\pmb{\theta}|\mathcal{X})}\left[ \left. p_{W_k,X_i}(\widetilde{\pmb{\theta}})
	\right|\mathcal{X}^{(h)}
	\right]$ following Equations~\eqref{eq.sensitivity_h2} and \eqref{eq.critical_var_h}, which is estimated by using $N_{\pi} = 500$, $B_O=5$ and $B_I=20$; see \cite{song2016shapley} for the selection of sampling parameter setting.
	Thus, we have the estimation uncertainty proportion $\widehat{\mbox{EP}}_{\theta_\ell}(p_{W_k,X_i}) \equiv \frac{1}{H} \sum_{h=1}^H 
	\frac{ \widehat{\widehat{\mbox{Sh}}}^*_{\theta_{\ell}}\left[ \left. p_{W_k,X_i}(\widetilde{\pmb{\theta}}) \right| \mathcal{X}^{(h)} \right]} 
	{
		{\widehat{\mbox{Var}}^*_{p(\pmb{\theta}|\mathcal{X})}\left[ \left. p_{W_k,X_i}(\widetilde{\pmb{\theta}})
			\right|\mathcal{X}^{(h)}
			\right]}}
	$ with $H=20$.
\end{sloppypar}

The coefficients contributing to the estimation of $\mbox{Sh}_{X_4,X_{20}}$ include $v_4^2$ and 18 linear coefficients $\pmb \beta$ on the paths from node $X_4$ to node $X_{20}$. The coefficients contributing to the estimation of $\mbox{Sh}_{X_{13},X_{20}}$ include $v_{13}^2$ and 6 linear coefficients $\pmb \beta$ located on the paths from $X_{13}$ to $X_{20}$. 
Due to the space limit, we only present the top five coefficients contributing most to the estimation uncertainty of criticality $p_{X_4,X_{20}}$ and $p_{X_{13},X_{20}}$, and aggregate the results for remaining coefficients. The sensitivity analysis results, $\widehat{\mbox{EP}}_{\theta_\ell}(p_{W_k,X_i})\pm \mbox{SE}\Big[ \widehat{\mbox{EP}}_{\theta_\ell}(p_{W_k,X_i})\Big]$, for $p_{X_4,X_{20}}$ and $p_{X_{13},X_{20}}$ are shown in Table~\ref{table:sensitivity1}, where SE stands for the standard error (SE). Notice that the coefficients that contribute the most to the estimation error of the criticality $p_{X_4,X_{20}}$ and $p_{X_{13},X_{20}}$ are the variance coefficients of CPPs ($v_4^2$ and $v_{13}^2$). The estimation uncertainty of linear coefficients have similar and relatively small contributions. Similar information can be presented through the sample visualization of the integrated bioprocess sensitivity analysis in Fig.~\ref{fig:MR_example}. The darkness of directed edges and circle boundaries indicates the seriousness of model uncertainty from corresponding model coefficients $\pmb{\beta}$ and $\pmb{v}$. \textit{Thus, this information can guide the process monitoring and data collection to efficiently reduce the impact of model uncertainty and facilitate learning and systematic risk control for integrated biomanufacturing system.}

\vspace{-0.in}
\begin{table}[hbt!]
	\small
	\centering
	\caption{The estimated relative contribution of each BN parameter estimation uncertainty (in terms of \%) on criticality assessment $\widehat{\mbox{EP}}_{\theta_\ell}(p_{W_k,X_i})\pm \mbox{SE}\Big[ \widehat{\mbox{EP}}_{\theta_\ell}(p_{W_k,X_i})\Big]$ for $p_{X_4,X_{20}}$ and $p_{X_{13},X_{20}}$. 
	}
	\label{table:sensitivity1}
		\begin{tabular}{|c|ccccc|c|}
			\hline
			$\theta_\ell\in \pmb{\theta}(X_4,X_{20})$        & $v_4^2$  & $\beta_{11, 15}$ & $\beta_{10, 14}$ & $\beta_{15, 20}$ & $\beta_{14, 20}$ & rest     \\ \hline
			$\widehat{\mbox{EP}}_{\theta_\ell}(p_{X_{4},X_{20}})$  & 73.75$\pm$1.96  & 1.59$\pm$0.13           & 1.58$\pm$0.19          & 1.57$\pm$0.21           & 1.56$\pm$0.13         & 19.95$\pm$4.70  \\ \hline 
			\hline
			$\theta_\ell\in \pmb{\theta}(X_{13},X_{20})$      & $v_{13}^2$ & $\beta_{13, 15}$ & $\beta_{16, 20}$ & $\beta_{15, 20}$ & $\beta_{13, 16}$ & rest     \\ \hline
			$\widehat{\mbox{EP}}_{\theta_\ell}(p_{X_{13},X_{20}})$ & 68.16$\pm$6.40    & 5.57$\pm$0.50           & 5.54$\pm$0.49           & 5.42$\pm$0.44           & 5.23$\pm$0.46           & 10.08$\pm$8.87  \\ \hline
		\end{tabular}%
\end{table}


\subsection{Real Case Study for Multiple Phase Cell Culture Process Risk and Sensitivity Analyses}
\label{subsec:caseStudy}

To study the performance of proposed bioprocess risk and sensitivity analyses, in this section, we consider the fed-batch fermentation process of \textit{Yarrowia lipolytica} yeast for citrate or citric acid (CA) production. This \textit{multiple-phase cell culture process} includes seed culture, cell growth and production processes. 
In the seed culture, the thawed seed vial solution of \textit{Y. lipolytica} strain is transferred to a shake flask containing seed culture medium, and then grown at \ang{30}C and 280 rpm until cell concentration reaches around 2--5 in OD600 (optical density measured at a wavelength of 600 nm), which usually takes 18--24h (hours). In the Fed-Batch Fermentation, the seed culture (50 mL) is first transferred to the bioreactor, which contains the initial fermentation medium (600 mL) and initial substrate (here we use 35g/L soybean oil). The feeding starts when the substrate concentration decreases below 20g/L, while the rate is adjusted to maintain the concentration of substrate about 20 g/L. During the fermentation, the dissolved oxygen level, denoted by p\ch{O2}, is set around 30\% of air saturation by cascade controls of agitation speed between 500 and 1,400 rpm, and the aeration rate is fixed at 0.3 L/min. The pH is controlled at 6.0 during 0--12h, then increased to 7.0 in 6 hours, and maintained at 7.0 in the remainder of run by feeding \ch{KOH} (i.e., feed of base). The temperature is maintained at \ang{30}C for the entire run. 
At several middle points of each run, the bioreactor state is estimated by using pH/p\ch{O2} probes and off-line sample measurement for residual substrate, which can guide the adjustment of operation decisions (i.e., feed rate).

\begin{figure}[h!]
	\centering
	\includegraphics[width=1.0\textwidth]{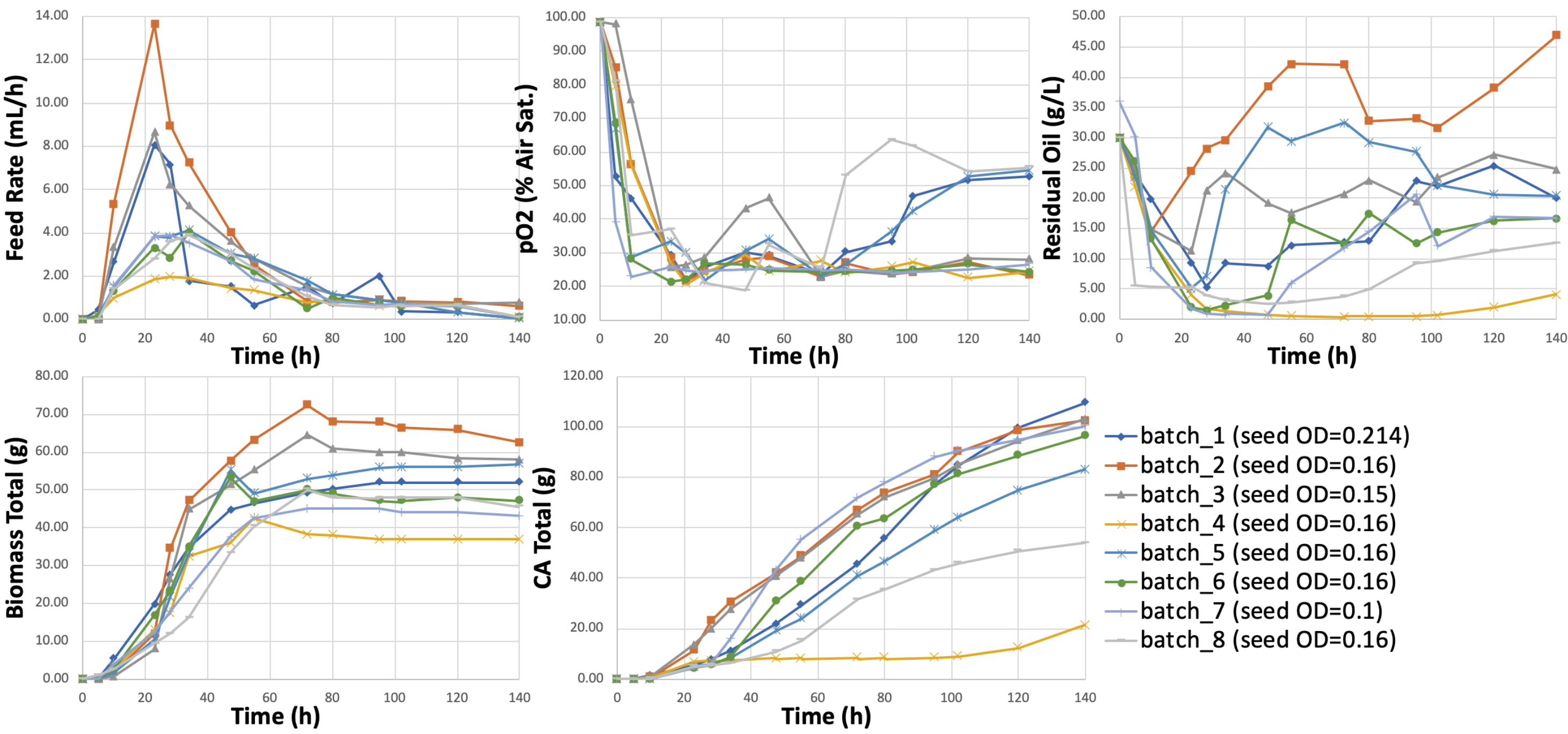}
	\vspace{-0.3in}
	\caption{Data of citric acid fed-batch fermentation case study.} \label{fig_caseData}
\end{figure}

\textit{In this real case study, we focus on the critical CPPs during the fed-batch fermentation, including cell concentration after seed culture process, feed rate, dissolved oxygen (p\ch{O2}), and residual oil. We consider main CQAs related to cell (i.e., total cell biomass) and productivity (i.e., total CA production).} Experiments are conducted in Dr. Dongming Xie's Lab to generate process data generate the data $\mathcal{X}$ with $R=8$ batches during 140 hours; see the data in Fig.~\ref{fig_caseData}. 
We want to study how the CPPs at different time contribute to the variation of intermediate and final CQAs outputs, while  evaluating the impact from model uncertainty. 

\begin{figure}[h!]
	\centering
	\includegraphics[width=0.9\textwidth]{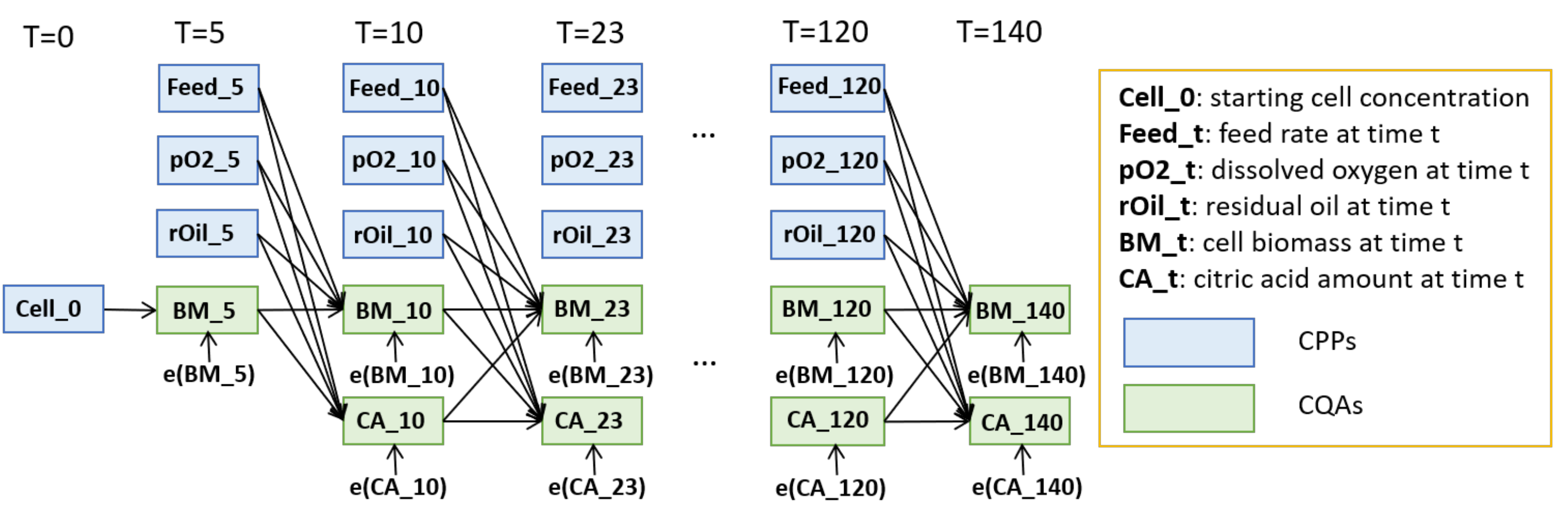}
	\vspace{-0.15in}
	\caption{BN model for citric acid fed-batch fermentation case study.} \label{fig_caseBN}
\end{figure}

\textit{Based on the interactions of CPPs/CQAs, we develop the BN-based bioprocess probabilistic model with 62 nodes; see the illustration in Fig.~\ref{fig_caseBN}.}
We first estimate the expected criticality $\mbox{E}[p_{W_k,X_i}]$ by using $B$ posterior samples of model coefficients, $\widehat{\mbox{E}}[p_{W_k,X_i}] = \frac{1}{B} \sum_{b=1}^B 
p_{W_k,X_i}(\widetilde{\pmb{\theta}}^{(b)}) \times 100\%$ with $\widetilde{\pmb{\theta}}^{(b)} \sim p(\pmb{\theta}|\mathcal{X})$ for $b=1,\ldots,B$, with $B=1000$. We record the results in terms of percentage (\%) in Tables~\ref{table:cpp2cqa_case_BM} and  \ref{table:cpp2cqa_case_CA} for cell biomass and CA production respectively with 
each row and column corresponding to random input $W_k$ and output $X_i$.
In addition, the overall impact of model uncertainty on the CPPs/CQAs criticality assessment can be quantified by the posterior standard deviation (SD), which can be estimated by,
$\widehat{\mbox{SD}}[p_{W_k,X_i}]  = \sqrt{
	\dfrac{1}{(B-1)}\sum_{b=1}^{B} \left[ p_{W_k,X_i}\left(\widetilde{\pmb{\theta}}^{(b)}\right) - 
	\bar{p}_{W_k,X_i} \right]^2} \times 100\%
$, where $\bar{p}_{W_k,X_i}=\frac{1}{B}\sum_{b=1}^B p_{W_k,X_i}(\widetilde{\pmb{\theta}}^{(b)})$. 
The results of SD are recorded in the bracket in Tables~\ref{table:cpp2cqa_case_BM} and  \ref{table:cpp2cqa_case_CA}. Due to the space limit, we only provide the dominant (high criticality level) part of time points. We also study the subplots and assess the impact from intermediate CQAs (biomass and CA amount) as inputs on the following output variation in Table~~\ref{table:cqa2cqa_case}.

\begin{table}[hbt!]
	\centering
	\caption{The estimated criticality level $\widehat{\mbox{E}}[p_{W_k,X_i}]$ and standard error $\widehat{\mbox{SD}}[p_{W_k,X_i}] $ (in \%) of any input CPP or other factor $W_k$ impacting on the variance of intermediate or final biomass $X_i$.}
	\label{table:cpp2cqa_case_BM}
	\resizebox{0.8\textwidth}{!}{%
		\begin{tabular}{|c|c|c|c|c|c|c|c|}
			\hline
			$\widehat{\mbox{E}}[p_{W_k,X_i}]$ & $X_i=\mbox{BM\_10}$       & BM\_23       & BM\_34       & BM\_55       & BM\_80       & BM\_102      & BM\_140              \\ \hline
			$W_k=\mbox{Cell\_0}$                          & 1.88(4.12)   & 2.92(5.57)   & 1.07(2.76)   & 0.76(2.35)   & 0.68(2.29)   & 0.65(2.22)   & 0.56(2.03)           \\
			Feed\_5                           & 15.62(15.79) & 4.29(7.15)   & 0.84(1.97)   & 0.66(1.81)   & 0.57(1.47)   & 0.52(1.43)   & 0.5(1.51)            \\
			pO2\_5                            & 34.23(22.29) & 14.77(16.21) & 2.83(5.82)   & 1.98(4.63)   & 1.69(4.52)   & 1.5(3.94)    & 1.41(3.84)           \\
			rOil\_5                           & 16.07(16.33) & 15.31(15.16) & 5.62(9.25)   & 4.02(7.3)    & 3.33(6.54)   & 3.07(6.41)   & 2.81(5.95)           \\
			e(BM\_5)                          & 13.6(15.38)  & 19.87(18.02) & 7.46(11.46)  & 5.35(9.5)    & 4.54(8.95)   & 4.23(8.65)   & 3.85(8.11)           \\ \hline
			Feed\_10                          &              & 3.09(5.72)   & 2(3.4)       & 1.17(2.23)   & 1.03(2.27)   & 0.98(2.16)   & 0.91(2.05)           \\
			pO2\_10                           &              & 8.4(10.58)   & 6.34(8.82)   & 4.11(6.13)   & 3.51(5.65)   & 3.29(5.44)   & 3.04(5.15)           \\
			rOil\_10                          &              & 9.02(11.96)  & 1.15(2.66)   & 0.71(1.86)   & 0.63(1.78)   & 0.58(1.68)   & 0.53(1.68)           \\
			e(BM\_10)                         & 18.6(13.88)  & 5.43(7.15)   & 0.74(1.52)   & 0.59(1.32)   & 0.48(1.15)   & 0.43(1.09)   & 0.41(1.05)           \\
			e(CA\_10)                         &              & 4.62(6.61)   & 1.55(3.04)   & 0.97(2.03)   & 0.82(1.8)    & 0.77(1.73)   & 0.72(1.64)           \\ \hline
			Feed\_23                          &              &              & 21.69(21.63) & 7.23(12.14)  & 6.19(11.11)  & 5.71(10.56)  & \textbf{5.17(9.8)}   \\
			pO2\_23                           &              &              & 4.41(6.16)   & 1.64(2.89)   & 1.4(2.63)    & 1.27(2.44)   & 1.15(2.31)           \\
			rOil\_23                          &              &              & 12.52(15.96) & 5.93(8.72)   & 5.22(8.27)   & 4.77(8)      & 4.31(7.31)           \\
			e(BM\_23)                         &              & 12.28(12.71) & 1.47(2.84)   & 0.8(1.62)    & 0.7(1.54)    & 0.63(1.36)   & 0.56(1.24)           \\
			e(CA\_23)                         &              &              & 0.63(1.39)   & 0.4(0.83)    & 0.35(1.01)   & 0.33(1.03)   & 0.31(0.93)           \\ \hline
			Feed\_28                          &              &              & 7.06(8.28)   & 2.59(4.54)   & 2.13(4.16)   & 1.97(3.76)   & 1.75(3.36)           \\
			pO2\_28                           &              &              & 1.25(2.27)   & 1.07(1.91)   & 0.85(1.47)   & 0.77(1.38)   & 0.71(1.35)           \\
			rOil\_28                          &              &              & 13.68(15.45) & 14.07(15.83) & 11.88(14.56) & 10.83(13.98) & \textbf{9.84(13.28)} \\
			e(BM\_28)                         &              &              & 5.56(5.64)   & 1.56(2.44)   & 1.34(2.29)   & 1.24(2.1)    & 1.15(1.99)           \\
			e(CA\_28)                         &              &              & 0.04(0.07)   & 0.05(0.1)    & 0.05(0.1)    & 0.04(0.09)   & 0.04(0.08)           \\ \hline
			Feed\_34                          &              &              &              & 4(6.98)      & 3.2(6.14)    & 2.67(5.43)   & 2.27(5.33)           \\
			pO2\_34                           &              &              &              & 3.02(5.28)   & 2.54(4.85)   & 2.18(4.27)   & 1.96(4.04)           \\
			rOil\_34                          &              &              &              & 3.47(6.19)   & 2.76(5.38)   & 2.55(5.2)    & 2.32(5.04)           \\
			e(BM\_34)                         &              &              & 2.09(4.34)   & 0.45(1.17)   & 0.38(1.08)   & 0.32(0.94)   & 0.29(0.71)           \\
			e(CA\_34)                         &              &              &              & 1.23(2.44)   & 1.01(2.3)    & 0.88(2.11)   & 0.77(1.87)           \\ \hline
			Feed\_55                          &              &              &              &              & 2.1(4.45)    & 1.68(3.61)   & 1.23(2.75)           \\
			pO2\_55                           &              &              &              &              & 3.99(7.64)   & 3.07(6.24)   & 2.16(4.77)           \\
			rOil\_55                          &              &              &              &              & 7.34(13.28)  & 5.86(10.8)   & 4.42(8.74)           \\
			e(BM\_55)                         &              &              &              & 9.88(10.83)  & 5.31(7.39)   & 4.4(6.26)    & 3.33(5.29)           \\
			e(CA\_55)                         &              &              &              &              & 0.05(0.15)   & 0.08(0.4)    & 0.08(0.54)           \\ \hline
			Feed\_80                          &              &              &              &              &              & 0.94(2.56)   & 0.67(2.05)           \\
			pO2\_80                           &              &              &              &              &              & 1.12(2.99)   & 0.78(2.57)           \\
			rOil\_80                          &              &              &              &              &              & 6.2(11.46)   & 4.09(7.93)           \\
			e(BM\_80)                         &              &              &              &              & 0.15(0.47)   & 0.1(0.27)    & 0.07(0.19)           \\
			e(CA\_80)                         &              &              &              &              &              & 0.02(0.06)   & 0.03(0.08)           \\ \hline
			Feed\_102                         &              &              &              &              &              &              & 0.47(1.33)           \\
			pO2\_102                          &              &              &              &              &              &              & 0.4(1.07)            \\
			rOil\_102                         &              &              &              &              &              &              & \textbf{7.77(14.9)}  \\
			e(BM\_102)                        &              &              &              &              &              & 0.62(1.76)   & 0.4(1.25)            \\
			e(CA\_102)                        &              &              &              &              &              &              & 0(0.01)              \\ \hline
			Feed\_120                         &              &              &              &              &              &              & 1.97(3.28)           \\
			pO2\_120                          &              &              &              &              &              &              & 1.19(2.15)           \\
			rOil\_120                         &              &              &              &              &              &              & 4.33(7.71)           \\ \hline
		\end{tabular}%
	}
\end{table}

\begin{table}[hbt!]
	\centering
	\caption{The estimated criticality level $\widehat{\mbox{E}}[p_{W_k,X_i}]$ and standard error $\widehat{\mbox{SD}}[p_{W_k,X_i}] $ (in \%) of any input CPP or other factor $W_k$ impacting on the variance of intermediate or final CA amount $X_i$.}
	\label{table:cpp2cqa_case_CA}
	\resizebox{0.8\textwidth}{!}{%
		\begin{tabular}{|c|c|c|c|c|c|c|c|}
			\hline
			$\widehat{\mbox{E}}[p_{W_k,X_i}]$ & $X_i=\mbox{CA\_10}$       & CA\_23       & CA\_34       & CA\_55       & CA\_80       & CA\_102     & CA\_140               \\ \hline
			$W_k=\mbox{Cell\_0}$                           & 5.22(7.79)   & 2.41(4.68)   & 1.22(3.11)   & 0.96(2.83)   & 0.89(2.69)   & 0.82(2.39)  & 0.79(2.46)            \\
			Feed\_5                           & 7.88(10.99)  & 3.27(5.04)   & 1.25(2.2)    & 0.82(1.81)   & 0.75(1.72)   & 0.68(1.62)  & 0.65(1.59)            \\
			pO2\_5                            & 5.5(8.81)    & 4.98(9.24)   & 3.05(6.24)   & 2.41(5.46)   & 2.21(5.09)   & 2.09(4.95)  & 2.01(4.74)            \\
			rOil\_5                           & 30.63(18.78) & 13.61(13.6)  & 6.37(9.37)   & 4.9(8.2)     & 4.63(8.18)   & 4.36(7.89)  & 4.24(7.87)            \\
			e(BM\_5)                          & 38.42(18.88) & 16.6(15.47)  & 8.09(11.72)  & 6.33(10.22)  & 6(10.14)     & 5.59(9.63)  & 5.39(9.51)            \\ \hline
			Feed\_10                          &              & 12.64(12.02) & 2.49(3.9)    & 1.61(2.76)   & 1.45(2.53)   & 1.35(2.45)  & 1.3(2.46)             \\
			pO2\_10                           &              & 34.18(21.2)  & 7.97(9.38)   & 5.23(7.39)   & 4.79(6.94)   & 4.42(6.62)  & 4.2(6.39)             \\
			rOil\_10                          &              & 1.52(3.62)   & 1.2(2.53)    & 0.97(2.58)   & 0.9(2.5)     & 0.84(2.34)  & 0.77(2.16)            \\
			e(BM\_10)                         &              & 1.6(3.54)    & 1.06(2.14)   & 0.71(1.46)   & 0.66(1.45)   & 0.62(1.41)  & 0.6(1.38)             \\
			e(CA\_10)                         & 12.36(13.95) & 3.69(5.19)   & 1.52(2.98)   & 1.18(2.24)   & 1.12(2.2)    & 1.03(2.08)  & 0.99(2.07)            \\ \hline
			Feed\_23                          &              &              & 8.07(10.91)  & 9.44(13.3)   & 8.23(12.12)  & 7.7(11.7)   & \textbf{7.26(11.28)}  \\
			pO2\_23                           &              &              & 2.71(3.52)   & 2.17(3.5)    & 1.91(3.08)   & 1.82(3.03)  & 1.74(2.96)            \\
			rOil\_23                          &              &              & 11.22(11.85) & 7.8(10.56)   & 6.99(9.71)   & 6.63(9.55)  & \textbf{6.32(9.31)}   \\
			e(BM\_23)                         &              &              & 1.15(1.87)   & 1.04(1.95)   & 0.92(1.63)   & 0.84(1.52)  & 0.81(1.5)             \\
			e(CA\_23)                         &              & 5.51(7.08)   & 0.91(1.58)   & 0.56(1.31)   & 0.51(1.24)   & 0.47(1.18)  & 0.45(1.12)            \\ \hline
			Feed\_28                          &              &              & 3.5(7.99)    & 3.91(6.66)   & 3.6(6.27)    & 3.38(5.89)  & 3.19(5.48)            \\
			pO2\_28                           &              &              & 2.43(3.6)    & 1.35(2.29)   & 1.25(2.19)   & 1.17(2.09)  & 1.12(2.09)            \\
			rOil\_28                          &              &              & 29.32(20.85) & 17.37(17.85) & 16.13(17.22) & 14.9(16.39) & \textbf{14.35(16.17)} \\
			e(BM\_28)                         &              &              & 1.17(2.25)   & 2.28(3.33)   & 2(2.97)      & 1.88(2.83)  & 1.77(2.73)            \\
			e(CA\_28)                         &              &              & 0.15(0.24)   & 0.07(0.12)   & 0.06(0.11)   & 0.06(0.11)  & 0.05(0.1)             \\ \hline
			Feed\_34                          &              &              &              & 6.47(11.6)   & 4.79(8.82)   & 4.22(7.96)  & 3.92(7.42)            \\
			pO2\_34                           &              &              &              & 6.22(9.95)   & 4.72(8.05)   & 4.18(7.17)  & 3.84(6.77)            \\
			rOil\_34                          &              &              &              & 4.85(7.94)   & 3.97(7.04)   & 3.63(6.56)  & 3.36(6.3)             \\
			e(BM\_34)                         &              &              &              & 0.71(1.79)   & 0.63(1.58)   & 0.56(1.41)  & 0.52(1.31)            \\
			e(CA\_34)                         &              &              & 5.16(8.58)   & 1.74(3.99)   & 1.52(3.31)   & 1.42(3.25)  & 1.32(3.08)            \\ \hline
			Feed\_55                          &              &              &              &              & 0.5(1.55)    & 0.54(1.59)  & 0.57(1.57)            \\
			pO2\_55                           &              &              &              &              & 0.62(1.86)   & 0.7(2.02)   & 0.7(1.99)             \\
			rOil\_55                          &              &              &              &              & 2.97(5.6)    & 2.76(5.43)  & 2.82(6.04)            \\
			e(BM\_55)                         &              &              &              &              & 1.58(3.7)    & 1.72(3.62)  & 1.78(3.58)            \\
			e(CA\_55)                         &              &              &              & 0.5(1.34)    & 0.46(1.68)   & 0.38(1.23)  & 0.35(1.21)            \\ \hline
			Feed\_80                          &              &              &              &              &              & 0.72(3.18)  & 0.66(2.92)            \\
			pO2\_80                           &              &              &              &              &              & 0.7(2.98)   & 0.67(2.9)             \\
			rOil\_80                          &              &              &              &              &              & 2.75(6.66)  & 2.58(6.22)            \\
			e(BM\_80)                         &              &              &              &              &              & 0(0.02)     & 0.01(0.03)            \\
			e(CA\_80)                         &              &              &              &              & 0.34(1.12)   & 0.29(0.96)  & 0.26(0.84)            \\ \hline
			Feed\_102                         &              &              &              &              &              &             & 0.16(0.49)            \\
			pO2\_102                          &              &              &              &              &              &             & 0.04(0.18)            \\
			rOil\_102                         &              &              &              &              &              &             & 1.77(4.04)            \\
			e(BM\_102)                        &              &              &              &              &              &             & 0.02(0.16)            \\
			e(CA\_102)                        &              &              &              &              &              & 0.06(0.22)  & 0.05(0.21)            \\ \hline
			Feed\_120                         &              &              &              &              &              &             & 0.42(1.97)            \\
			pO2\_120                          &              &              &              &              &              &             & 0.32(1.48)            \\
			rOil\_120                         &              &              &              &              &              &             & 1.03(3.77)            \\ \hline
		\end{tabular}%
	}
\end{table}

\begin{table}[hbt!]
	\centering
	\caption{The estimated criticality level $\widehat{\mbox{E}}[p_{W_k,X_i}]$ and standard error $\widehat{\mbox{SD}}[p_{W_k,X_i}] $ (in \%) of any input CQA $W_k$ impacting on the variance of intermediate or final biomass or CA amount $X_i$.}
	\label{table:cqa2cqa_case}
	\resizebox{\textwidth}{!}{%
		\begin{tabular}{|c|cc|cc|cc|cc|}
			\hline
			$\widehat{\mbox{E}}[p_{W_k,X_i}]$ & $X_i=\mbox{BM\_55}$       & CA\_55       & BM\_80       & CA\_80       & BM\_102      & CA\_102      & BM\_140               & CA\_140               \\ \hline
			$W_k=\mbox{BM\_10}$                            & 3.59(7.62)   & 4.29(8.09)   & 2.99(6.7)    & 3.95(7.58)   & 2.67(6.23)   & 3.74(7.34)   & 2.45(5.83)            & 3.61(7.15)            \\
			CA\_10                            & 11.09(16.78) & 13.45(18.7)  & 9.41(15.62)  & 12.72(18.59) & 8.74(14.99)  & 11.82(17.68) & 8.04(14.08)           & 11.44(17.56)          \\ \hline
			BM\_23                            & 11.29(18.41) & 13.97(20.33) & 9.59(16.82)  & 12.93(19.42) & 8.71(15.77)  & 12.01(18.61) & 7.99(14.9)            & 11.53(18.18)          \\
			CA\_23                            & 11.02(15.02) & 13.72(17.42) & 9.43(14.17)  & 12.81(16.92) & 8.88(13.8)   & 11.94(16.22) & 8.16(13.01)           & 11.46(16.07)          \\ \hline
			BM\_28                            & 20.66(25.26) & 27.56(27.27) & 17.94(23.75) & 24.68(26.25) & 16.92(22.87) & 23.42(25.67) & 15.56(21.97)          & 22.26(25.2)           \\
			CA\_28                            & 17.94(19.6)  & 21.7(20.84)  & 15.18(18.26) & 20.08(20.67) & 13.63(17.48) & 18.47(19.75) & 12.34(16.23)          & 17.74(19.64)          \\ \hline
			BM\_34                            & 25.11(29.47) & 35.52(33.18) & 21.67(26.95) & 31.32(31.48) & 20.52(26.31) & 29.67(30.59) & 18.67(25.32)          & \textbf{27.91(29.73)} \\
			CA\_34                            & 35.03(33.43) & 40.71(35.87) & 29.32(31.31) & 38.45(35.08) & 26.44(29.72) & 35.59(33.99) & 24.04(28.29)          & 34.43(33.78)          \\ \hline
			BM\_47.5                          & 37.75(35.28) & 5.79(13.43)  & 25.9(30.21)  & 12.71(20.88) & 22.94(28.07) & 12.96(20.94) & 19.66(26.42)          & 13.03(20.92)          \\
			CA\_47.5                          & 36.49(34.21) & 93.23(14.45) & 36.68(32.59) & 75.43(28.48) & 34.07(31.48) & 68.86(30.46) & \textbf{32.04(31.25)} & 64.62(31.61)          \\ \hline
			BM\_55                            &              &              & 58.85(34.64) & 12.7(22.51)  & 49.22(35.71) & 14.46(24.33) & 39.99(34.47)          & 15.33(25.24)          \\
			CA\_55                            &              &              & 19.49(25.99) & 82.28(25.87) & 22.22(28.09) & 75.43(29.57) & 23.04(28.06)          & 70.83(31.2)           \\ \hline
			BM\_72                            &              &              & 95.2(9.58)   & 3.25(9.46)   & 73.79(29.1)  & 7.78(15.83)  & 57.99(33.23)          & 9.38(18.1)            \\
			CA\_72                            &              &              & 3.03(8.26)   & 95.77(10.77) & 13.71(22.8)  & 86.34(20.9)  & 16.95(24.58)          & 81.4(24.37)           \\ \hline
			BM\_80                            &              &              &              &              & 76.49(27.65) & 6.85(14.48)  & 59.42(32.6)           & 9.15(17.69)           \\
			CA\_80                            &              &              &              &              & 12.56(21.31) & 88.21(19.51) & 16.54(23.74)          & 82.39(23.6)           \\ \hline
			BM\_95                            &              &              &              &              & 94.44(10.84) & 1.48(3.17)   & \textbf{70.6(28.14)}  & 5.15(11.38)           \\
			CA\_95                            &              &              &              &              & 4.39(9.63)   & 98.43(3.24)  & 12.25(18.9)           & \textbf{91.22(15.09)} \\ \hline
			BM\_102                           &              &              &              &              &              &              & 72.79(27.28)          & 5.28(11.54)           \\
			CA\_102                           &              &              &              &              &              &              & 10.93(17.17)          & 91.31(15.3)           \\ \hline
			BM\_120                           &              &              &              &              &              &              & 84.91(18.97)          & 3.16(8.28)            \\
			CA\_120                           &              &              &              &              &              &              & 7.81(14.19)           & 95.19(11.36)          \\ \hline
		\end{tabular}%
	}
\end{table}

Differing with the simulation study in Section~\ref{subsec:simulationEmpiricalStudy}, 
there is no macro-replication in the real case study.
\textit{Notice that the posterior standard deviation (SD) can measure the overall model uncertainty, i.e., the variation of criticality estimates cross different posterior samples characterizing the model coefficient estimation uncertainty.}
Based on the sample average of $B$ posterior samples $\bar{p}_{W_k,X_i} = \frac{1}{B} \sum_{b=1}^B 
p_{W_k,X_i}(\widetilde{\pmb{\theta}}^{(b)}) \times 100\%$, the estimation accuracy of criticality $\bar{p}_{W_k,X_i}$ is measured by the standard error (SE) with $\mbox{SE}(\bar{p}_{W_k,X_i}) = SD(\bar{p}_{W_k,X_i})/\sqrt{B}$.

The results in Tables~\ref{table:cpp2cqa_case_BM} and  \ref{table:cpp2cqa_case_CA} show that the variations of residual oil and feed rate in the cell growth phase (about from time $t=23h$ to $28h$) have dominant impact on both intermediate and final cell biomass and CA productivity. 
As fermentation time further increases, the criticality level of input factors $W_k$ on the output $X_i$, CA production, tends to decrease. It matches well with the data in Fig.~\ref{fig_caseData}, the cell growth and production both become slower and more stable. This observation suggests that controlling the CPPs (i.e., feed rate and residual oil) to ensure the good cell growth stage is more important in order to improve the process stability. For the cell total biomass output in Table~\ref{table:cpp2cqa_case_BM}, since the residual oil generates the scattering particles impacting on OD600 and cell biomass measurement accuracy, this effect becomes larger as the residual oil increases, which explains the high contribution of residual oil at the end of process (i.e., rOil\_102) to the final cell biomass measurement variation. 

By studying the subplots, we study 
the impact of middle step CQAs (i.e., cell biomass and CA amount in the cell growth and production phases) on the final output variation. We record the results in Table~\ref{table:cqa2cqa_case}. 
As the fermentation time $t$ increases, the explained variations of final biomass and CA by current values  increase. They reach to around 70\% for biomass and 90\% for CA at time $t=95h$. This observation is consistent with the data in Fig.~\ref{fig_caseData} and the growth of biomass/CA is relative slow in the periods after it. However, compared with CA, biomass has relatively larger prediction variation even in the later stages of production, which can be explained by the measurement errors of Cell OD induced by large amount of residual oil. 
In terms of biomass impacting on final CA (or CA impacting on biomass), the most critical part is biomass at time $t=34h$ (or CA amount at $t=47.5h$).
Since cell growth needs nitrogen, the production phase usually starts when nitrogen concentration becomes small.
During those periods (around 30h to 50h), nitrogen from the initial medium is consumed and both intracellular lipid (which becomes part of biomass) accumulation and extracellular CA production are induced by nitrogen limitation. It can be also observed in Fig.~\ref{fig_caseData}, where the sudden increase of CA total slopes happens around 30h to 50h, whose variations have critical contribution to final CA output uncertainty. 


\begin{sloppypar}
	We also conduct the sensitivity analysis studying the impact of model uncertainty on the CPPs/CQAs criticality assessment. Here we focus on criticality assessment estimation of $p_{rOil\_{28},CA\_{140}}$ and $p_{rOil\_{102},BM\_{140}}$, which have high criticality and overall model uncertainty; see Tables~\ref{table:cpp2cqa_case_BM} and  \ref{table:cpp2cqa_case_CA}.
	The model coefficients contributing to the estimation of $p_{rOil\_{28},CA\_{140}}$ include $v_{rOil\_{28}}^2$ and 32 linear coefficients $\pmb \beta$ on the paths from node $rOil\_{28}$ to node $CA\_{140}$, whereas coefficients contributing to the estimation of $p_{Feed\_{23},CA\_{140}}$ include $v_{Feed\_{23}}^2$ and 36 linear coefficients $\pmb \beta$. We present the top five coefficients contributors to the estimation uncertainty of criticality $p_{rOil\_{28},CA\_{140}}$ and $p_{Feed\_{23},CA\_{140}}$, and aggregate the results for remaining coefficients in Table~\ref{table:sensitivity_case}. 
	From the results, the estimation uncertainty of variance coefficients of CPPs ($v^2_{rOil\_28}$ and $v^2_{Feed\_23}$) have the largest contribution to the estimation uncertainty of the criticality $p_{rOil\_{28},CA\_{140}}$ and $p_{Feed\_{23},CA\_{140}}$. The estimation uncertainty of coefficients $\pmb{\beta}$ in both sets
	$\pmb{\theta}(rOil\_{28},CA\_{140})$ and $ \pmb{\theta}(Feed\_{23},CA\_{140})$ 
	have similar and relative lower contributions. 
\end{sloppypar}

\begin{table}[hbt!]
	\centering
	\caption{The estimated relative contribution of each BN parameter estimation uncertainty (in terms of \%) on criticality assessment $\widehat{\mbox{EP}}_{\theta_\ell}(p_{W_k,X_i})$ for $p_{Feed\_{20},CA\_{140}}$. 
	}
	\label{table:sensitivity_case}
		\begin{tabular}{|c|ccccc|c|}
			\hline
			$\theta_\ell\in \pmb{\theta}(rOil\_{28},CA\_{140})$        & $v^2_{rOil\_28}$  & $\beta_{CA\_72, BM\_80}$ & $\beta_{CA\_55, CA\_72}$ & $\beta_{rOil\_28, CA\_34}$ & $\beta_{CA\_120, CA\_140}$ & rest     \\ \hline
			$\widehat{\mbox{EP}}_{\theta_\ell}(p_{rOil\_{28},CA\_{140}})$  & 23.38  & 4.00           & 4.00          & 3.86           & 3.55        & 61.21  \\ \hline 
			\hline 		$\theta_\ell\in \pmb{\theta}(Feed\_{23},CA\_{140})$      & $v^2_{Feed\_23}$ & $\beta_{BM\_72, BM\_80}$ & $\beta_{CA\_72, BM\_80}$ & $\beta_{BM\_95, CA\_102}$ & $\beta_{BM\_72, CA\_80}$ & rest     \\ \hline
			$\widehat{\mbox{EP}}_{\theta_\ell}(p_{Feed\_{23},CA\_{140}})$ & 19.52    & 5.10           & 4.10           & 3.97           & 3.95           & 63.36  \\ \hline
		\end{tabular}%
\end{table}

\vspace{-0.2in}

\section{Conclusions}
\label{sec:conclusion}

Driven by the critical challenges in biomanufacturing, we create an integrated bioprocess knowledge graph and propose interpretable risk and sensitivity analyses, which can provide the production process risk- and science-based understanding, guide the CPPs/CQAs specifications and production stability control, and facilitate the process development. Since hundreds of factors can impact on the product quality and productivity, and also the amount of process observations is often very limited, we explore the process interactions and causal relationships, and then develop a Bayesian network (BN) based probabilistic knowledge graph characterizing the causal interdependencies of production process CPPs/CQAs. Building on the knowledge graph, we propose the BN-SV based sensitivity analysis to assess the criticality of each random input factor on the variance of intermediate/final product quality attributes by using the Shapley value (SV), which can correctly account for input interdependencies and process structural interactions. 
We further introduce the BN-SV-MU sensitivity analysis, which can provide the comprehensive understanding on how the estimation uncertainty of each part of process model coefficients impacts on the production risk analysis and CPPs/CQAs criticality assessment. It can guide bioprocess sensor monitoring and ``most informative" data collection to facilitate bioprocess learning and model uncertainty reduction.
Both simulation and real case studies are used to demonstrate the promising performance of proposed bioprocess risk and sensitivity analyses.



\section*{ACKNOWLEDGEMENTS}
The authors are grateful for constructive comments from Dr. Barry Nelson (Northeastern University), Peter Baker (Green Mountain Quality Assurance, LLC), help from Hua Zheng (NEU) on the development of bioprocess knowledge graph visualization, and help from Na Liu (UMass Lowell) conducting lab experiments.

\printendnotes

\bibliographystyle{apalike}
\bibliography{BN_ref,proj_ref,sensitivity_ref}

\appendix

\section{Ontology based Data and Process Integration}
\label{subsec:relationalGraph}

\begin{figure}[h!]
	\centering
	\includegraphics[width=1\textwidth]{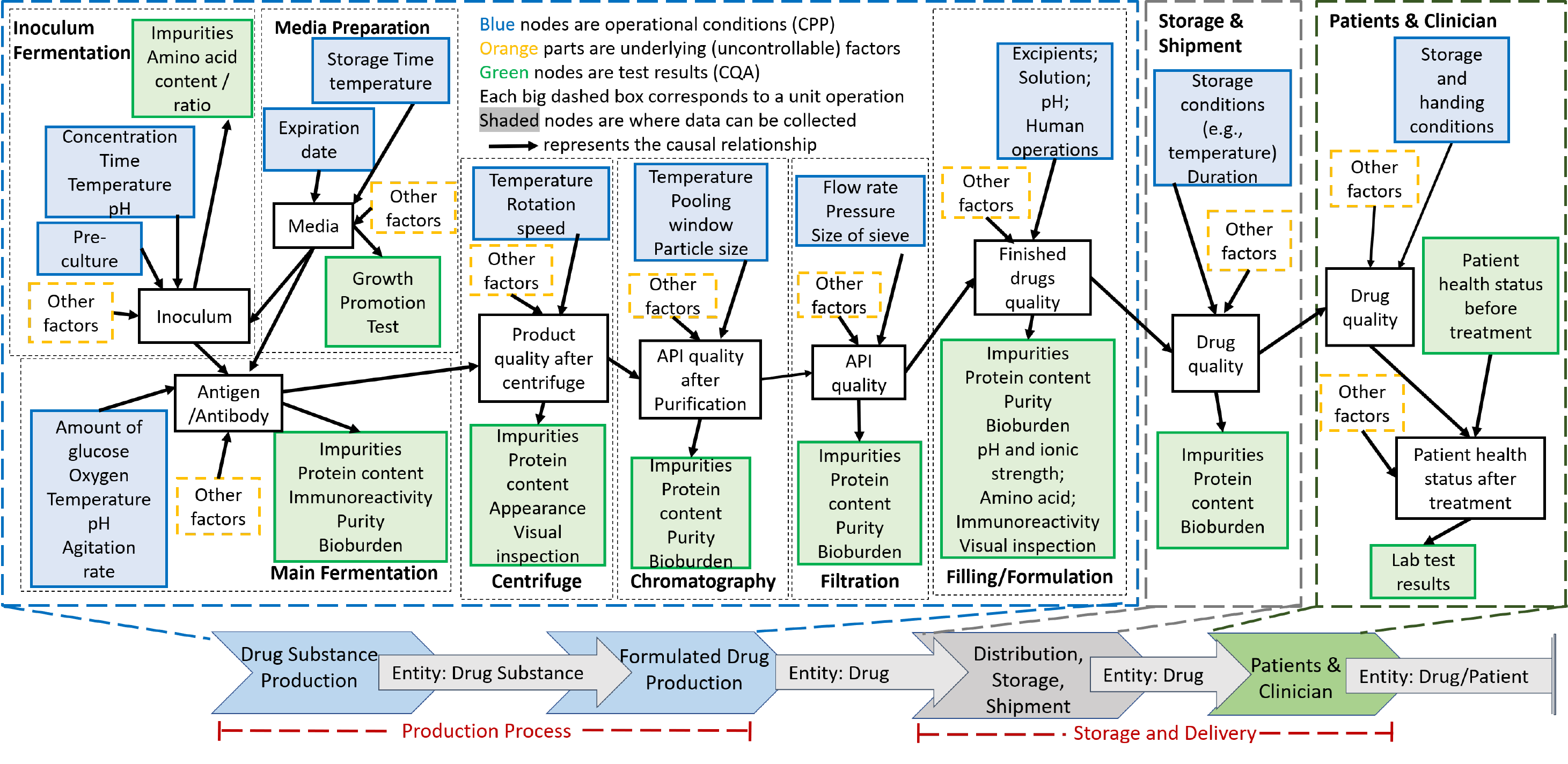}
	\vspace{-0.2in}
	\caption{Biopharmaceutical production process ontology based causal relationships. 
	}\label{fig_1}
\end{figure}

By exploring the causal relationships and interactions in the production processes, we introduce \textit{bioprocess ontology-based data integration}, which can connect all distributed and heterogeneous data collected from bioprocess.
This relational graph can enable the connectivity of  end-to-end process from drug development to patient response;
see Fig.~\ref{fig_1} for a simplified illustration of integrated biopharmaceutical manufacturing supply chain. 
Nodes represent factors (i.e., CPPs/CQAs, media feed, bioreactor operating conditions, other uncontrolled factors) impacting the process outputs, and the directed edges model the causal relationships. Each dashed block could represent a \textit{module}, which can be each phase or each unit operation. 
In this relational graph, the shaded nodes represent the variables with real-world observations, including the testing and sensor monitoring data of CPPs/CQAs for raw materials, operation conditions, and intermediate/final drug products. The unshaded and dashed nodes represent variables without observations and residuals, including the complete quality status of intermediate and final drug products, and other uncontrollable factors (e.g., contamination) introduced during the process unit operations.

\section{Detailed Derivation of Equation~(\ref{eq.represent})}\label{sec:appendix1}

In order to show Equation~\eqref{eq.represent}, we consider more general results as following, 
\begin{equation}
	X_{n} = \mu_{n} + \sum_{k=1}^{m^p}\gamma_{k,n}(X_k - \mu_k) + \sum_{k=m^p+1}^{n}\gamma_{k,n} e_k, \label{eq.represent_g}
\end{equation}
for $n=m^p+1, \ldots, m+1$, where $\gamma_{k,n}$ is given as Equations~\eqref{eq.weightY} and \eqref{eq.weightY2}. Notice according to linear Gaussian model \eqref{eq.CQA}, we can write
$X_{m^p+1} = \mu_{m^p+1} + \sum_{k=1}^{m^p}\beta_{k,m^p+1}(X_k - \mu_k) + e_{m^p+1}$,
where $\beta_{k,m^p+1} = 0$ for $k \notin Pa(X_{m^p+1})$. Suppose  Equation~\eqref{eq.represent_g} holds for all $n=m^p+1, \ldots, n_0$. For $n=n_0+1$,
by applying linear Gaussian model, we have
\begin{align}
	&X_{n_0+1} = \mu_{n_0+1} + \sum_{k=1}^{n_0}\beta_{k,n_0+1}(X_k - \mu_k) + e_{n_0+1}, \nonumber \\
	& =
	\mu_{n_0+1} + \sum_{k=1}^{m^p}\beta_{k,n_0+1}(X_k - \mu_k) + \sum_{\ell=m^p+1}^{n_0}\beta_{\ell,n_0+1}\left[ \sum_{k=1}^{m^p}\gamma_{k,\ell}(X_k - \mu_k) + \sum_{k=m^p+1}^{\ell}\gamma_{k,\ell} e_k \right]  + e_{n_0+1}, \label{mid101} \\
	& = \mu_{n_0+1} + \sum_{k=1}^{m^p}\left[ \beta_{k,n_0+1} + \sum_{\ell=m^p+1}^{n_0} \gamma_{k,\ell}\beta_{\ell,n_0+1} \right] (X_k - \mu_k) +  \sum_{k=m^p+1}^{n_0} \left[ \sum_{\ell=k}^{n_0}\gamma_{k,\ell} \beta_{\ell,n_0+1} \right]e_k + e_{n_0+1} \nonumber \\
	& =  \mu_{n_0+1} + \sum_{k=1}^{m^p}\gamma_{k,n_0+1}(X_k - \mu_k) + \sum_{k=m^p+1}^{n_0+1}\gamma_{k,n_0+1} e_k. \label{mid102}
\end{align}
Step (\ref{mid101}) follows by applying (\ref{eq.represent_g}). Step~(\ref{mid102}) follows by applying Equations~\eqref{eq.weightY} and \eqref{eq.weightY2}.
By mathematical induction, we can conclude that Equation~\eqref{eq.represent_g} holds for all $n=m^p+1, \ldots, m+1$.

\section{Detailed Derivation of Equation~(\ref{eq.shapley_3})}\label{sec:appendix2}

We consider $W_k$ and $\mathcal{J} \subset \mathcal{K}/\{k\}$. For $\mathcal{J}=\emptyset$, we have
\[
\dfrac{(m -|\mathcal{J}|)! |\mathcal{J}|!} {(m+1)!} [c(\mathcal{J}\cup\{k\}) - c(\mathcal{J})] = \dfrac{1}{m+1} \gamma_{k, m+1}^2\mbox{Var}(W_k).
\]
For $|\mathcal{J}|=m^\prime$ with $m^\prime = 1, \ldots, m$, we have
\begin{align}
	&\sum_{\left\{\mathcal{J}: |\mathcal{J}|=m^\prime\right\} }  \dfrac{(m -|\mathcal{J}|)! |\mathcal{J}|!} {(m+1)!} [c(\mathcal{J}\cup\{k\}) - c(\mathcal{J})] \nonumber \\
	&= \sum_{\left\{\mathcal{J}: |\mathcal{J}|=m^\prime\right\} } \dfrac{(m -m^\prime)! m^\prime!} {(m+1)!}\left[ \gamma_{k, m+1}^2\mbox{Var}(W_k) + 2\sum_{\ell \in \mathcal{J}}\gamma_{k, m+1}\gamma_{\ell, m+1}\mbox{Cov}(W_k, W_\ell) \right] \nonumber \\
	&= \dfrac{(m -m^\prime)! m^\prime!}{(m+1)!} \Bigg\{ {{m} \choose {m^\prime}}\gamma_{k, m+1}^2\mbox{Var}(W_k) \label{mid103} \\
	& 
	~~~~+ 2
	\sum_{\ell \in \mathcal{K}/\{k\}} \left[
	\sum_{\left\{\mathcal{J}: |\mathcal{J}|=m^\prime \mbox{ and } \ell \in \mathcal{J}\right\}} \gamma_{k, m+1}\gamma_{\ell, m+1}\mbox{Cov}(W_k, W_\ell) \right] \Bigg\} \label{mid105}  \\
	&= \dfrac{1}{m+1}\gamma_{k, m+1}^2\mbox{Var}(W_k) + 2
	\dfrac{(m -m^\prime)! m^\prime!}{(m+1)!}{{m-1} \choose {m^\prime-1}}\sum_{\ell \in \mathcal{K}/\{k\}}\gamma_{k, m+1}\gamma_{\ell, m+1}\mbox{Cov}(W_k, W_\ell)  \label{mid104} \\
	& = \dfrac{1}{m+1}\gamma_{k i}^2\mbox{Var}(W_k) + \dfrac{2m^\prime}{m(m+1)}\sum_{\ell \in \mathcal{K}/\{k\}}\gamma_{k, m+1}\gamma_{\ell, m+1}\mbox{Cov}(W_k, W_\ell).
	\nonumber 
\end{align}
Step \eqref{mid103} holds because the number of all subsets $\mathcal{J}$ with size $m^\prime$ is ${{m} \choose {m^\prime}}$. In Step~(\ref{mid105}), we shift the order of sums over $\mathcal{J}$ and $\ell$. Then, Step~(\ref{mid104}) holds because given $W_\ell$, the number of subset $\left\{\mathcal{J}: |\mathcal{J}|=m^\prime \mbox{ and } \ell \in \mathcal{J}\right\}$ is ${{m-1} \choose {m^\prime-1}}$. 
So, we get the Shapley value,
\begin{eqnarray}
	\lefteqn{	\mbox{Sh}_{W_k,X_{m+1}}(\pmb{\theta})  = \sum_{\mathcal{J}\subset \mathcal{K}/\{k\}}\dfrac{(m -|\mathcal{J}|)! |\mathcal{J}|!} {(m+1)!} [c(\mathcal{J}\cup\{k\}) - c(\mathcal{J})] } \nonumber \\
	&= & \sum_{m^\prime=0}^{m}\left[ 
	\dfrac{1}{m+1}\gamma_{k i}^2\mbox{Var}(W_k) + \dfrac{2m^\prime}{m(m+1)}\sum_{\ell \in \mathcal{K}/\{k\}}\gamma_{k, m+1}\gamma_{\ell, m+1}\mbox{Cov}(W_k, W_\ell)
	\right] \nonumber \\
	&= &  \gamma_{k, m+1}^2\mbox{Var}(W_k) + \sum_{\ell \in \mathcal{K}/\{k\}}\gamma_{k, m+1}\gamma_{\ell, m+1}\mbox{Cov}(W_k, W_\ell). \nonumber 
\end{eqnarray}



\section{Derivation and Procedure for BN Learning and Gibbs Sampler}
\label{sec:derGibbs}


We derive the posterior distribution of BN model parameters $p(\pmb{\theta}|\mathcal{X})$ and introduce a Gibbs sampling approach to generate the posterior samples, $\widetilde{\pmb{\theta}}^{(b)}\sim p(\pmb{\theta}|\mathcal{X})$ with $b=1,2,\ldots,B$ quantifying the model uncertainty. In Section~\ref{subsubsec:completeData}, we first 
provide the derivation for conditional posterior distribution with complete production process data described in Section~\ref{subsec:BayesianLearning}.
Considering the situations where we could have some additional incomplete batch data (e.g., batches in the middle of production or thrown away at certain production step based on the quality control strategy), we further extend the Bayesian learning approach to cases with mixing data in Section~\ref{subsubsec:completeIncompleteData}.
Then, we provide the Gibbs sampling procedure to generate the posterior samples $\widetilde{\pmb{\theta}}^{(b)}$ with $b=1,2,\ldots,B$ in Section~\ref{subsubsec:GibbsSample}.

\subsection{Knowledge Learning for Cases with Complete Production Process Data}
\label{subsubsec:completeData}

Following Section~\ref{subsec:BayesianLearning}, we first derive the conditional posterior distribution for the weight coefficient 
$\beta_{ij}$,
\begin{align}
	&p(\beta_{ij}|\mathcal{X}, \pmb \mu, \pmb v^2, \pmb \beta_{-ij})\propto \left[\prod_{r=1}^{R}p(x_j^{(r)}|x_{Pa(X_j)}^{(r)})\right] p(\beta_{ij}),
	\nonumber \\
	&\propto \exp\left\{-\sum_{r=1}^{R}\dfrac{1}{2v_j^2}\left[(x_j^{(r)}-\mu_j) - \beta_{ij}(x_i^{(r)}-\mu_i) - \sum_{k\in Pa(j)/\{i\}}\beta_{kj}(x_k^{(r)} - \mu_k) \right]^2 \right. \nonumber\\
	&~~~~ \left.
	-\dfrac{1}{2\tau_{ij}^{(0)2}}\left(\beta_{ij}-\theta_{ij}^{(0)} \right)^2 \right\}, \nonumber \\
	&\propto \exp\left\{-\dfrac{1}{2v_j^2}\sum_{r=1}^{R} \left(\alpha_i^{(r)}\beta_{ij} - m_{ij}^{(r)} \right)^2 -\dfrac{1}{2\tau_{ij}^{(0)2}}\left(\beta_{ij}-\theta_{ij}^{(0)} \right)^2 \right\}, 
	\nonumber \\
	&\propto 
	\exp\left\{-\dfrac{\beta_{ij}^2}{2}\left( \sum_{r=1}^{R}\dfrac{\alpha_i^{(r)2}}{v_j^2}+ \dfrac{1}{\tau_{ij}^{(0)2}}\right)+ \beta_{ij}\left( \sum_{r=1}^{R} \dfrac{\alpha_{i}^{(r)}m_{ij}^{(r)}}{v_j^2}+\dfrac{\theta_{ij}^{(0)}}{\tau_{ij}^{(0)2}}\right) \right\}
	= \mathcal{N}(\theta_{ij}^{(R)}, \tau_{ij}^{(R)2}),
	\nonumber 
\end{align}
where
$\theta_{ij}^{(R)} = \dfrac{\tau_{ij}^{(0)2}\sum_{r=1}^{R} \alpha_i^{(r)} m_{ij}^{(r)} + v_j^2\theta_{ij}^{(0)}}{\tau_{ij}^{(0)2}\sum_{r=1}^{R} \alpha_i^{(r)2} + v_j^2}$ and 
$\tau_{ij}^{(R)2} = \dfrac{\tau_{ij}^{(0)2}v_j^2} {\tau_{ij}^{(0)2}\sum_{r=1}^{R} \alpha_i^{(r)2} + v_j^2}$
with 
$
\alpha_{i}^{(r)} = x_i^{(r)} - \mu_i, 
\quad \textrm{and} \quad
m_{ij}^{(r)} = (x_j^{(r)} - \mu_j) - \sum_{X_k\in Pa(X_j)/\{X_i\}}\beta_{kj}(x_k^{(r)} - \mu_k).$

Second, we derive the conditional posterior distribution for the variance coefficient $v_i^2=\mbox{Var}[X_i|Pa(X_i)]$ with $i=1,2,\ldots,m+1$,
\begin{eqnarray}
	\lefteqn{ p(v_i^2|\mathcal{X}, \pmb \mu, \pmb v_{-i}^2, \pmb \beta) \propto  \left[\prod_{r=1}^{R}p(x_i^{(r)}|x_{Pa(X_i)}^{(r)})\right] p(v_i^2)  }\nonumber \\
	&\propto& (v_i^2)^{-R/2 - \kappa_i^{(0)}/2 -1}\exp\left\{ -\dfrac{1}{2v_j^2}\sum_{r=1}^{R}\left[ (x_i^{(r)} - \mu_i) - \sum_{X_k\in Pa(X_i)}\beta_{ki}(x_k^{(r)} - \mu_k) \right]^2 \right\}  \nonumber \\
	&\propto& (v_i^2)^{-R/2 - \kappa_i^{(0)}/2 -1}\exp\left\{ -\dfrac{1}{2v_j^2}\sum_{r=1}^{R}u_i^{(r)2} - \dfrac{\lambda_i^{(0)}}{2v_j^2} \right\} 
	= \mbox{Inv-}\Gamma\left(\dfrac{\kappa_i^{(R)}}{2}, \dfrac{\lambda_i^{(R)}}{2}\right), 
	\nonumber 
\end{eqnarray}
where $\kappa_i^{(R)} = \kappa_i^{(0)} + R$, 
$\lambda_i^{(R)} = \lambda_i^{(0)} + \sum_{r=1}^{R}u_i^{(r)2}$ and 
$u_i^{(r)} = (x_i^{(r)} - \mu_i) - \sum_{X_k\in Pa(X_i)}\beta_{ki}(x_k^{(r)} - \mu_k).
$

Third, we derive the conditional posterior distribution of mean coefficient $\mu_i$ with $i=1,2,\ldots,m+1$ for any CPP and CQA,
\begin{align}
	&p(\mu_i|\mathcal{X}, \pmb \mu_{-i}, \pmb v^2, \pmb \beta) \propto p(\mu_i) \prod_{r=1}^{R}\left[p(x_i^{(r)}|x_{Pa(X_i)}^{(r)}) \prod_{j\in \mathcal{S}(X_i)} p(x_j^{(r)}|x_{Pa(X_j)}^{(r)}) \right] 
	\nonumber \\
	&\propto \exp\Bigg\{-\dfrac{1}{2v_i^2}\sum_{r=1}^{R} \left[ (x_i^{(r)} - \mu_i) - \sum_{X_k\in Pa(X_i)}\beta_{ki}(x_k^{(r)} - \mu_k) \right]^2  \nonumber \\
	&~~~~ - \sum_{r=1}^{R}\sum_{X_j\in 
		\mathcal{S}(X_i)} \dfrac{1}{2v_j^2} \left[ (x_j^{(r)} - \mu_j) - \sum_{X_k\in Pa(X_j)}\beta_{kj}(x_k^{(r)} - \mu_k) \right]^2
	-\dfrac{1}{2\sigma_{i}^{(0)2}}\left(\mu_i-\mu_i^{(0)} \right)^2 \Bigg\}, 
	\nonumber \\
	&\propto \exp\left\{-\dfrac{1}{2v_i^2}\sum_{r=1}^{R} \left(\mu_i - a_i^{(r)} \right)^2 - \sum_{r=1}^{R}\sum_{X_j\in 
		\mathcal{S}(X_i)} -\dfrac{1}{2v_j^2}\left(\beta_{ij}\mu_i - c_{ij}^{(r)} \right)^2 \right. \nonumber\\
	&~~~~ \left.
	-\dfrac{1}{2\sigma_{i}^{(0)2}}\left(\mu_i-\mu_i^{(0)} \right)^2 \right\}, 
	\nonumber \\
	&\propto \exp\left\{-\dfrac{\mu_i^2}{2}\left( \dfrac{R}{v_i^2} + \sum_{X_j\in S(X_i)}\dfrac{R\beta_{ij}^2}{v_j^2} + \dfrac{1}{\sigma_{i}^{(0)2}} \right)
	+ \mu_i\left( \sum_{r=1}^{R}\dfrac{a_{i}^{(r)}}{v_i^2} + \sum_{r=1}^{R}\sum_{X_j\in \mathcal{S}(X_i)}\dfrac{\beta_{ij}c_{ij}^{(r)}}{v_j^2} \right. \right. \nonumber\\
	&~~~~ \left. \left.
	+ \dfrac{\mu_i^{(0)}}{\sigma_i^{(0)2}} \right) \right\} = \mathcal{N}(\mu_i^{(R)}, \sigma_i^{(R)2}), 
	\nonumber 
\end{align}
where
$\mu_i^{(R)} = \sigma_i^{(R)2} \left[ \dfrac{\mu_i^{(0)}}{\sigma_i^{(0)2}} + \sum_{r=1}^{R} \dfrac{a_i^{(r)}}{v_i^2} + \sum_{r=1}^{R}\sum_{X_j\in S(X_i)} \dfrac{\beta_{ij}c_{ij}^{(r)}}{v_j^2} \right] $ and
$\dfrac{1}{\sigma_i^{(R)2}} = \dfrac{1}{\sigma_i^{(0)2}} + \dfrac{R}{v_i^2} + \sum_{X_j\in S(X_i)}\dfrac{R\beta_{ij}^2}{v_j^2},
$
with
$
a_{i}^{(r)} = x_i^{(r)} - \sum_{X_k\in Pa(X_i)}\beta_{kj}(x_k^{(r)} - \mu_k)$  and 
$c_{ij}^{(r)} = \beta_{ij}x_i^{(r)} - (x_j^{(r)} - \mu_j) + \sum_{X_k\in Pa(X_j)/\{X_i\}}\beta_{kj}(x_k^{(r)} - \mu_k).
$

\subsection{Knowledge Learning for Cases with Mixing Data}
\label{subsubsec:completeIncompleteData}

Except the case with complete production data discussed in Section~\ref{subsubsec:completeData}, 
we consider the cases with additional incomplete data corresponding to certain ``Top Sub-Graph", denoted by $G(\mathbf{N}^\prime|\pmb \theta(\mathbf{N}^\prime))$ with $\mathbf{N}^\prime \subseteq \mathbf{N}$, such that any CQA node $X_j \in \mathbf{N}^\prime$ has $Pa(X_j)\subset \mathbf{N}^\prime$.
Since batch data collected from biopharmaceutical production process are usually limited, we want to fully utilize both complete and incomplete data to estimate the BN model coefficients and improve our knowledge of production process. 

Without loss of generality, we consider the real-world data including two data sets $\mathcal{X}=\{\mathcal{X}_1,\mathcal{X}_2\}$ with the complete data $\mathcal{X}_1 = \{(x_1^{(r_1)}, x_2^{(r_1)}, \ldots, x_{m+1}^{(r_1)})$ for $r_1=1,2,\ldots,R_1\}$ and the incomplete data $\mathcal{X}_2 = \{ (x_i^{(r_2)}: X_i\in \mathbf{N}^\prime)$ for $ r_2=R_1+1,R_1+2,\ldots,R\}$, where $R=R_1+R_2$. 
Our approach can be easily extended to cases with multiple incomplete data sets.
We use the same prior distribution $p(\pmb \mu, \pmb v^2, \pmb \beta)$ as shown in Equation~(\ref{eq.prior}). Given the mixing data  $\mathcal{X} = \{\mathcal{X}_1, \mathcal{X}_2\}$, we can derive the posterior distribution of $\pmb{\theta}$,
\begin{equation}
	p(\pmb \mu, \pmb v^2, \pmb \beta|\mathcal{X}) \propto \prod_{r_1=1}^{R_1}\left[\prod_{i=1}^{m+1}p(x_i^{(r_1)} | x_{Pa(X_i)}^{(r_1)})\right] \prod_{r_2=R_1+1}^{R}\left[\prod_{X_i\in \mathbf{N}^\prime}p(x_i^{(r_2)} | x_{Pa(X_i)}^{(r_2)})\right] p(\pmb \mu, \pmb v^2, \pmb \beta). \label{eq.posterior_2} \nonumber 
\end{equation}
For $\beta_{ij}$ with $X_j\notin \mathbf{N}^\prime$ or $v_i^2$ and $\mu_i$ with node $X_i\notin \mathbf{N}^\prime$,
the conditional posterior is the same as complete data case and we can utilize Equations~(\ref{eq.post_b1}), (\ref{eq.post_v1}) and (\ref{eq.post_mu1}) by replacing $\mathcal{X}$ with $\mathcal{X}_1$.

Thus, to derive the full Gibbs sampler, we only need to provide the updated conditional posterior accounting for those nodes included in the incomplete data set $\mathcal{X}_2$. 
We first derive the conditional posterior distribution for weight coefficient $\beta_{ij}$ with $X_j\in \mathbf{N}^\prime$.
\begin{eqnarray}
	\lefteqn{ p(\beta_{ij}|\mathcal{X}, \pmb \mu, \pmb v^2, \pmb \beta_{-ij})\propto \left[\prod_{r=1}^{R_1+R_2}p(x_j^{(r)}|x_{Pa(X_j)}^{(r)})\right] p(\beta_{ij}), }
	\nonumber \\
	&\propto& \exp\left\{-\sum_{r=1}^{R_1+R_2}\dfrac{1}{2v_j^2}\left[(x_j^{(r)}-\mu_j) - \beta_{ij}(x_i^{(r)}-\mu_i) - \sum_{k\in Pa(j)/\{i\}}\beta_{kj}(x_k^{(r)} - \mu_k) \right]^2 \right. \nonumber \\
	& & \left. -\dfrac{1}{2\tau_{ij}^{(0)2}}\left(\beta_{ij}-\theta_{ij}^{(0)} \right)^2 \right\}, \nonumber \\
	&\propto & \exp\left\{-\dfrac{1}{2v_j^2}\sum_{r=1}^{R_1+R_2} \left(\alpha_i^{(r)}\beta_{ij} - m_{ij}^{(r)} \right)^2 
	-\dfrac{1}{2\tau_{ij}^{(0)2}}\left(\beta_{ij}-\theta_{ij}^{(0)} \right)^2 \right\}, 
	\nonumber \\
	&\propto & \exp\left\{-\dfrac{\beta_{ij}^2}{2}\left( \sum_{r=1}^{R_1+R_2}\dfrac{\alpha_i^{(r)2}}{v_j^2}+ \dfrac{1}{\tau_{ij}^{(0)2}}\right)+ \beta_{ij}\left( \sum_{r=1}^{R_1+R_2} \dfrac{\alpha_{i}^{(r)}m_{ij}^{(r)}}{v_j^2}+\dfrac{\theta_{ij}^{(0)}}{\tau_{ij}^{(0)2}}\right) \right\} 
	\nonumber \\
	&=& \mathcal{N}(\theta_{ij}^{(R_1+R_2)}, \tau_{ij}^{(R_1+R_2)2}), 
	\label{eq.post_b_2}
\end{eqnarray}
where
$
\theta_{ij}^{(R_1+R_2)} = \dfrac{\tau_{ij}^{(0)2}\sum_{r=1}^{R_1+R_2} \alpha_i^{(r)} m_{ij}^{(r)} + v_j^2\theta_{ij}^{(0)}}{\tau_{ij}^{(0)2}\sum_{r=1}^{R_1+R_2} \alpha_i^{(r)2} + v_j^2}$ and $
\tau_{ij}^{(R_1+R_2)2} = \dfrac{\tau_{ij}^{(0)2}v_j^2} {\tau_{ij}^{(0)2}\sum_{r=1}^{R_1+R_2} \alpha_i^{(r)2} + v_j^2} 
\nonumber 
$
with 
$
\alpha_{i}^{(r)} = x_i^{(r)} - \mu_i$
and 
$m_{ij}^{(r)} = (x_j^{(r)} - \mu_j) - \sum_{X_k\in Pa(X_j)/\{X_i\}}\beta_{kj}(x_k^{(r)} - \mu_k)$ for $r = 1,2,\ldots,R$.

\begin{sloppypar}
	Then, we derive the conditional posterior distribution for $v_i^2$ with $X_i\in \mathbf{N}^\prime$,
	\begin{eqnarray}
		\lefteqn{ p(v_i^2 |\mathcal{X}, \pmb \mu, \pmb v_{-i}^2, \pmb \beta) \propto \left[\prod_{r=1}^{R_1+R_2}p(x_i^{(r)}|x_{Pa(X_i)}^{(r)})\right] p(v_i^2), }
		\nonumber \\
		&\propto& (v_i^2)^{-(R_1+R_2)/2 - \kappa_i^{(0)}/2 -1}\exp\left\{ -\dfrac{1}{2v_j^2}\sum_{r=1}^{R_1+R_2}\left[ (x_i^{(r)} - \mu_i) - \sum_{X_k\in Pa(X_i)}\beta_{ki}(x_k^{(r)} - \mu_k) \right]^2 \right\},  \nonumber \\
		&\propto & (v_i^2)^{-(R_1+R_2)/2 - \kappa_i^{(0)}/2 -1}\exp\left\{ -\dfrac{1}{2v_j^2}\sum_{r=1}^{R_1+R_2}u_i^{(r)2} - \dfrac{\lambda_i^{(0)}}{2v_j^2} \right\} \nonumber \\
		&=&  \mbox{Inv-}\Gamma\left(\dfrac{\kappa_i^{(R_1+R_2)}}{2}, \dfrac{\lambda_i^{(R_1+R_2)}}{2}\right),  \label{eq.post_v_2}
	\end{eqnarray}
	where $
	\kappa_i^{(R_1+R_2)} = \kappa_i^{(0)} + R$ and $
	\lambda_i^{(R_1+R_2)} = \lambda_i^{(0)} + \sum_{r=1}^{R}u_i^{(r)2}$ 
	with
	$
	u_i^{(r)} = (x_i^{(r)} - \mu_i) - \sum_{X_k\in Pa(X_i)}\beta_{ki}(x_k^{(r)} - \mu_k)$ for $r = 1,2,\ldots,R$.
\end{sloppypar}

\begin{sloppypar}
	After that, we derive the conditional posterior for mean coefficient $\mu_i$ with $X_i\in \mathbf{N}^\prime$,
	\begin{align}
		&p(\mu_i|\mathcal{X},  \pmb \mu_{-i}, \pmb v^2, \pmb \beta)
		\propto p(\mu_i) 
		\prod_{r=1}^{R_1+R_2}p(x_i^{(r)}|x_{Pa(X_i)}^{(r)}) 
		\prod_{r_1=1}^{R_1}\prod_{X_j\in \mathcal{S}(X_i)}p(x_j^{(r_1)}|x_{Pa(X_j)}^{(r_1)}) \nonumber\\
		&~~~~~~~~~~~~~~~~~~~~~~~~~~~~~~~~~~~~\cdot
		\prod_{r_2=R_1+1}^{R_1+R_2}\prod_{X_j\in S(X_i)\cap \mathbf{N}^\prime}p(x_j^{(r_2)}|x_{Pa(X_j)}^{(r_2)}), 
		\nonumber \\
		&\propto \exp\Bigg\{-\dfrac{1}{2v_i^2}\sum_{r=1}^{R_1+R_2} \bigg[ (x_i^{(r)} - \mu_i) - \sum_{X_k\in Pa(X_i)}\beta_{ki}(x_k^{(r)} - \mu_k) \bigg]^2 
		\nonumber \\  
		&~~~~ 
		- \sum_{r_1=1}^{R_1}\sum_{X_j\in 
			\mathcal{S}(X_i)} \dfrac{1}{2v_j^2} \bigg[ (x_j^{(r_1)} - \mu_j) -   \sum_{X_k\in Pa(X_j)}\beta_{kj}(x_k^{(r_1)} - \mu_k) \bigg]^2 
		\nonumber\\ 
		&~~~~ 
		- \sum_{r_2=R_1+1}^{R_1+R_2}\sum_{X_j\in S(X_i)\cap \mathbf{N}^\prime} \dfrac{1}{2v_j^2} \bigg[ (x_j^{(r_2)} - \mu_j) -  
		\sum_{X_k\in Pa(X_j)}\beta_{kj}(x_k^{(r_2)} - \mu_k) \bigg]^2  \nonumber\\
		&~~~~
		-\dfrac{1}{2\sigma_{i}^{(0)2}}\left(\mu_i-\mu_i^{(0)} \right)^2 \Bigg\}, 
		\nonumber \\
		&\propto  \exp\left\{-\dfrac{\mu_i^2}{2}\left( \dfrac{R_1+R_2}{v_i^2} + \sum_{X_j\in \mathcal{S}(X_i)}\dfrac{R_1\beta_{ij}^2}{v_j^2} + \sum_{X_j\in S(X_i)\cap \mathbf{N}^\prime}\dfrac{R_2\beta_{ij}^2}{v_j^2} \dfrac{1}{\sigma_{i}^{(0)2}} \right) \right.
		\nonumber \\
		& ~~~~+ \left.
		\mu_i\left( \sum_{r=1}^{R_1+R_2}\dfrac{a_{i}^{(r)}}{v_i^2} + \sum_{r_1=1}^{R_1}\sum_{X_j\in \mathcal{S}(X_i)}\dfrac{\beta_{ij}c_{ij}^{(r_1)}}{v_j^2} + \sum_{r_2=R_1+1}^{R}\sum_{X_j\in \mathcal{S}(X_i)\cap \mathbf{N}^\prime}\dfrac{\beta_{ij}c_{ij}^{(r_2)}}{v_j^2}
		+ \dfrac{\mu_i^{(0)}}{\sigma_i^{(0)2}} \right) \right\}, 
		\nonumber \\
		&= \mathcal{N}\left(\mu_i^{(R_1+R_2)}, \sigma_i^{(R_1+R_2)2}\right), 
		\label{eq.post_mu_2}
	\end{align}
	$\mu_i^{(R_1+R_2)} = 
	\sigma_i^{(R_1+R_2)2} \left[ \dfrac{\mu_i^{(0)}}{\sigma_i^{(0)2}} + \sum_{r=1}^{R_1+R_2} \dfrac{a_i^{(r)}}{v_i^2} + \sum_{r_1=1}^{R_1}\sum_{X_j\in \mathcal{S}(X_i)}\dfrac{\beta_{ij}c_{ij}^{(r_1)}}{v_j^2}  + \right. \\ \left. \sum_{r_2=R_1+1}^{R}\sum_{X_j\in 
		\mathcal{S}(X_i)\cap \mathbf{N}^\prime}\dfrac{\beta_{ij}c_{ij}^{(r_2)}}{v_j^2} \right]$, and 
	$\dfrac{1}{\sigma_i^{(R_1+R_2)2}} = \dfrac{1}{\sigma_i^{(0)2}} + \dfrac{R_1+R_2}{v_i^2} + \sum_{X_j\in \mathcal{S}(X_i)}\dfrac{R_1\beta_{ij}^2}{v_j^2} + \sum_{X_j\in 
		\mathcal{S}(X_i)\cap \mathbf{N}^\prime}\dfrac{R_2\beta_{ij}^2}{v_j^2}
	$
	with 
	$
	a_{i}^{(r)} = x_i^{(r)} - \sum_{X_k\in Pa(X_i)}\beta_{kj}(x_k^{(r)} - \mu_k)$ and $
	c_{ij}^{(r)} = \beta_{ij}x_i^{(r)} - (x_j^{(r)} - \mu_j) + \sum_{X_k\in Pa(X_j)/\{X_i\}}\beta_{kj}(x_k^{(r)} - \mu_k)$ for $r = 1,2,\ldots,R$.
	Here for illustration, we have only provided the conditional posteriors with two datasets $\mathcal{X}_1$ and $\mathcal{X}_2$. These derivations can be easily extended to similar cases with multiple datasets collected from complete graph and different top sub-graphs.
\end{sloppypar}


\subsection{Gibbs Sampling Procedure for BN Model Bayesian Inference}
\label{subsubsec:GibbsSample}

Based on the derived conditional posterior distributions in Sections~\ref{subsubsec:completeData} and \ref{subsubsec:completeIncompleteData}, we provide the Gibbs sampling procedure in Algorithm~\ref{alg:procedure2} to generate posterior samples  $\widetilde{\pmb{\theta}}^{(b)} \sim p(\pmb{\theta}|\mathcal{X})$ with $ \widetilde{\pmb{\theta}}^{(b)} = (\widetilde{\pmb \mu}^{(b)}, \widetilde{\pmb v}^{(b)2}, \widetilde{\pmb \beta}^{(b)})$ and $b=1,\ldots,B$. 
We first set the vague prior $p(\pmb \theta)=p(\pmb \mu, \pmb v^2, \pmb \beta)$ as Equation~\eqref{eq.prior}, and generate the initial point $\pmb \theta^{(0)} = (\pmb \mu^{(0)}, \pmb v^{(0)2}, \pmb \beta^{(0)})$ by sampling from the prior. Within each $t$-th iteration of Gibbs sampling, given the previous sample $\pmb \theta^{(t-1)} = (\pmb \mu^{(t-1)}, \pmb v^{(t-1)2}, \pmb \beta^{(t-1)})$, we sequentially compute and generate one sample from the conditional posterior distribution for each coefficient $\beta_{ij}$, $v_i^2$ and $\mu_i$.
By repeating this procedure, we can get samples $\pmb \theta^{(t)} = (\pmb \mu^{(t)}, \pmb v^{(t)2}, \pmb \beta^{(t)})$ with $t=1,\ldots,T$. To reduce the initial bias and correlations between consecutive samples, we remove the first $T_0$ samples and keep one for every $h$ samples. Consequently, we obtain the posterior samples $\widetilde{\pmb{\theta}}^{(b)}  \sim p(\pmb{\theta}|\mathcal{X})$ with $b=1,\ldots,B$.

\begin{algorithm}[hbt!]
	\small
	\caption{Gibbs Sampling Procedure for BN Model Uncertainty Quantification} \label{alg:procedure2}
	\KwIn{
		the prior $p(\pmb \theta)$ and real-world data $\mathcal{X}$.}
	\KwOut{Posterior samples $\widetilde{\pmb{\theta}}^{(b)} = (\widetilde{\pmb \mu}^{(b)}, \widetilde{\pmb v}^{(b)2}, \widetilde{\pmb \beta}^{(b)}) \sim p(\pmb{\theta}|\mathcal{X})$ with $b=1,\ldots,B$.}
	
	(1) Set the initial value $\pmb \theta^{(0)} = (\pmb \mu^{(0)}, \pmb v^{(0)2}, \pmb \beta^{(0)})$ by sampling from prior $p(\pmb \theta)$\;
	
	\For{$t=1,2,\ldots,T$}{
		(2) Given the previous sample $\pmb \theta^{(t-1)} = (\pmb \mu^{(t-1)}, \pmb v^{(t-1)2}, \pmb \beta^{(t-1)})$\;

		(3) For each $\beta_{ij}$, generate $\beta_{ij}^{(t)}\sim p(\beta_{ij}|\mathcal{X}, \beta_{12}^{(t)}, \ldots,$ $\beta_{i,j-1}^{(t)}, \beta_{i,j+1}^{(t-1)}, \ldots, \beta_{m,m+1}^{(t-1)}, \pmb \mu^{(t-1)}, \pmb v^{(t-1)2})$ through Equation~\eqref{eq.post_b1} for complete data or \eqref{eq.post_b_2} for mixing data\;

		(4) For each $v_i^2$, generate $v_i^{(t)2} \sim p(v_i^2|\mathcal{X}, \pmb \beta^{(t)}, v_1^{(t)2},$ $\ldots, v_{i-1}^{(t)2}, v_{i+1}^{(t-1)2}, \ldots, v_{m+1}^{(t-1)2}, \pmb \mu^{(t-1)}))$ through Equation~\eqref{eq.post_v1} for complete data or \eqref{eq.post_v_2} for mixing data\;

		(5) For each $\mu_i$, generate $\mu_i^{(t)} \sim p(\mu_i|\mathcal{X}, \pmb \beta^{(t)}, \pmb v^{(t),2},$ $\mu_1^{(t)}, \ldots, \mu_{i-1}^{(t)}, \mu_{i+1}^{(t-1)}, \ldots, \mu_{n}^{(t-1)})$ through Equation~\eqref{eq.post_mu1} for complete data or \eqref{eq.post_mu_2} for mixing data\;

		(6) Obtain a new posterior sample $\pmb \theta^{(t)} = (\pmb \mu^{(t)}, \pmb v^{(t)2}, \pmb \beta^{(t)})$\;
	}
	
	(7) Set $\widetilde{\pmb \theta}^{(b)} = \pmb \theta^{(T_0 + (b-1)h + 1)}$ with some constant integer $T_0$ and $h$, to reduce the initial bias and correlation between consecutive samples.
	
\end{algorithm}

\section{Simulated Biopharmaceutical Production Data}
\label{sec:inputDataSimulation}

To study the performance of proposed framework, we generate the simulated production process data $\mathcal{X}$, which mimics the ``real-world data collection." The BN with coefficients $\pmb{\theta}^c$ characterizing the underlying production process interdependence is used for data generation, which is built according to the biomanufacturing domain knowledge.
The ranges of CPPs/CQAs are listed Table~\ref{table:range}. 
For each CPP $X_j \in
\mathbf{X}^p$
with range $[x_j^{low}, x_j^{up}]$, we can specify the marginal distribution $X_j\sim \mathcal{N}(\mu_j^c,(v_j^c)^2)$ with mean $\mu^c_j = (x_j^{low} + x_j^{up})/2$ and standard deviation $v_j^c = (x_j^{up} - x_j^{low})/4$. 
For each CQA $X_i \in
\{\mathbf{X}^a\cup\mathbf{Y} \}$ with range $[x_i^{low}, x_i^{up}]$, we have mean $\mu_i^c = (x_i^{low} + x_i^{up})/2$ and marginal variance $\mbox{Var}(X_i) = [(x_i^{up} - x_i^{low})/4]^2$. Based on Equation~\eqref{eq.VarDecompCPP}, the corresponding coefficient $v^c_i$ can be computed through back-engineering.
For the complex interdependence, Table~\ref{table:corr} provides the relative associations with levels (i.e., high, median, low) between input CPPs/CQAs with output CQAs in each operation unit, which is built based on the ``cause-and-effect matrix" in \cite{mitchell2013determining}.
For the high, median and low association between $X_i$ to $X_j$, we set the coefficient $\beta^c_{ij}=0.9, 0.6, 0.3$ respectively. 
Thus, we can specify the underlying true coefficients $\pmb{\theta}^c = (\pmb{\mu}^c, (\pmb{v}^2)^c, \pmb{\beta}^c)$. To mimic the ``real-world" data collection, we generate the production batch data $\mathcal{X}$ using the BN model with $\pmb{\theta}^c$. Then, to assess the performance of proposed framework, we assume that the true coefficient values are unknown. 

\begin{table}[hbt!]
	\centering
	\caption{{Range of CPPs/CQAs in the production procedure.}}
	\label{table:range}
	\resizebox{0.9\textwidth}{!}{%
		\begin{tabular}{|c|cc|cc|}
			\hline
			Process Unit Operation             & CPP                        & Range                        & CQA                                         & Range                                 \\ \hline
			Main Fermentation & pH                         & 6.8-7.2                      & impurities             & 3-11 pl                               \\
			& temperature                & 20-30 C                      & protein content                             & 1-5 g/L                               \\
			& Oxygen                     & 2.5-7.5\%                    & bioburden                                   & 5-15 CFU/100mL  \\
			& agitation rate             & 1.1-2.5 m/s                  &                                             &                                       \\ \hline
			Centrifuge        & temperature                & 20 to 30 C                   & impurities             & 3-11 pl                               \\
			& rotation speed             & 3-5K RPM                     & protein content                             & 5-15 CFU/100mL  \\ \hline
			Chromatography    & pooling window             & 10-30 min  & impurities              & 3-11 pl                               \\
			& temperature             & 2-10 C                & protein content                             & 1-5 g/L                               \\
			&                 &                    & bioburden                                   & 5-15 CFU/100mL  \\ \hline
			Filtration        & size of sieve              & 0.1-0.5 um                   & impurities              & 3-11 pl                               \\
			& flow rate                  & 25-100 mL/min                & protein content                             & 1-5 g/L                               \\
			\hline
		\end{tabular}%
	}
\end{table}

\begin{table}[hbt!]
	\centering
	\caption{{Relative association between input CPPs/CQAs with output CQAs in each process unit operation.}}
	\label{table:corr}
	\resizebox{\textwidth}{!}{%
		\begin{tabular}{|c|c|ccc|}
			\hline
			Process Unit Operation             & Input CPPs/CQAs                        & \multicolumn{3}{c|}{Output CQAs}                                 \\ 
			&                                        &  impurities & protein content & bioburden \\ \hline
			Main Fermentation & pH                                         &    high        &      high           &      low     \\
			& temperature                                              &      high      &        high         &     low      \\
			& Oxygen                                                   &     high       &         high        &      low     \\
			& agitation rate                                          &     high       &         high        &      low     \\ \hline
			Centrifuge        & temperature                             &    medium        &         medium        & ---       \\
			& rotation speed                        &   medium         &       medium          & ---       \\
			& impurities (main fermentation)          &    medium        &      medium         & ---       \\
			& protein content (main fermentation)       &        medium     &           medium      & ---       \\
			& bioburden (main fermentation)                   &     medium       &        medium         & ---       \\ \hline
			Chromatography    & pooling window                                     &     high       &       medium          &     high      \\
			& temperature                                   &  high          &     medium            &     high      \\
			& impurities (centrifuge)                     &  high          &      medium           &     high      \\
			& protein content (centrifuge)                       &     high       &        medium         &    high       \\ \hline
			Filtration        & size of sieve                            &     low       &       medium          &     medium      \\
			& flow rate                                   &    low        &         medium        &     medium      \\
			& impurities (chromatography)                         &    low        &       medium          &     medium      \\
			& protein content (chromatography)                       &    low        &      medium           &  medium         \\
			& bioburden (chromatography)                            &    low        &      medium           &    medium       \\ \hline
		\end{tabular}%
	}
\end{table}

\section{Study the Bayesian Learning and Inference}
\label{subsec:emp_convergence}


\begin{sloppypar}
	To evaluate the accuracy and efficiency of proposed Bayesian learning,
	we empirically study the convergence of BN coefficient inference. In each $k$-th macro-replication, we first mimic the ``real-world" production batch data collection through generating $\mathcal{X}^{(k)}=\{\mathbf{X}^{(k)}_1,\ldots,\mathbf{X}^{(k)}_R\}$ with $\mathbf{X}_i^{(k)}\sim F(\mathbf{X}|\pmb{\theta}^c)$ for $i=1,\ldots,R$ and $k=1,\ldots,K$. Then, we generate $B$ posterior samples $\widetilde{\pmb{\theta}}^{(k,b)}\sim p(\pmb{\theta}|\mathcal{X}^{(k)})$ with $b=1, 2, \ldots, B$. 
	For the Gibbs sampler in Algorithm~\ref{alg:procedure2} provided in online Appendix~\ref{subsubsec:GibbsSample}, we set the initial warm-up length $T_0 = 500$ and step-size $h=10$. 	
	With different size of complete ``real-world" batch data $R=30, 100, 500$, we compute the mean squared error (MSE) for each coefficient $\theta_\ell \in \pmb \theta$: $
	\mbox{MSE}(\theta_\ell) = \iint \left({\theta_\ell} - \theta_\ell^c \right)^2 d P(\theta_\ell|\mathcal{X}) dP(\mathcal{X}|\pmb{\theta}^c).
	\label{eq.MSE}$
	Based on $K=20$ macro-replications and $B=1000$ posterior samples of BN coefficients, we estimate $\mbox{MSE}(\theta_\ell)$ with  $
	\widehat{\mbox{MSE}}(\theta_\ell) = \dfrac{1}{KB} \sum_{k=1}^K\sum_{b=1}^B\left(\widetilde{\theta}_\ell^{(k,b)} - \theta_\ell^c\right)^2.
	$
	Since the total number of coefficients is large, we further group coefficients by mean $\pmb \mu$, conditional variance $\pmb v^2$ and linear coefficients $\pmb \beta$, and take average of the sample MSE respectively: $\widehat{\mbox{MSE}}(\pmb{\mu}) = \frac{1}{|\pmb{\mu}|} \sum_{\theta_\ell \in \pmb{\mu}} \widehat{\mbox{MSE}}(\theta_\ell)$,
	$\widehat{\mbox{MSE}}(\pmb{v}^2) = \frac{1}{|\pmb{v}^2|} \sum_{\theta_\ell \in \pmb{v}^2} \widehat{\mbox{MSE}}(\theta_\ell)$, and $\widehat{\mbox{MSE}}(\pmb{\beta}) = \frac{1}{|\pmb{\beta}|} \sum_{\theta_\ell \in \pmb{\beta}} \widehat{\mbox{MSE}}(\theta_\ell)$. The corresponding results are reported in Table~\ref{table:mse}. As the size of real-world data $R$ increases, the average MSE decreases, which implies the posterior samples obtained by Gibbs sampling procedure can converge to the true coefficients. 
\end{sloppypar}

\begin{table}[hbt!]
	\small
	\centering
	\caption{The MSE of $\pmb \mu$, $\pmb v^2$ and $\pmb \beta$ esimated by using the Gibbs sampling.}
	\label{table:mse}
	\begin{tabular}{|c|ccc|}
		\hline
		Batch Data Size &   $\widehat{\mbox{MSE}}(\pmb{\mu})$  &  $\widehat{\mbox{MSE}}(\pmb{v}^2)$    &  $\widehat{\mbox{MSE}}(\pmb{\beta})$   \\ \hline
		$R=30$      & 0.122$\pm$0.032 & 0.276$\pm$0.029 & 0.0225$\pm$0.0013  \\
		$R=100$     & 0.075$\pm$0.023 & 0.061$\pm$0.006 & 0.0063$\pm$0.0004  \\
		$R = 500$   & 0.009$\pm$0.003 & 0.013$\pm$0.001 & 0.0011$\pm$0.00004 \\ \hline
	\end{tabular}
\end{table}

\end{document}